\documentclass[sigconf]{acmart}
\AtBeginDocument{%
  }

\setcopyright{acmlicensed}
\copyrightyear{2026}
\acmYear{2026}
\acmDOI{10.1145/3770855.3819057}
\acmConference[KDD '26]{Proceedings of the 32nd ACM SIGKDD Conference on Knowledge Discovery and Data Mining V.2}{August 9--13, 2026}{Jeju,Korea}
\acmISBN{978-1-4503-XXXX-X/2018/06}

\usepackage[colorinlistoftodos]{todonotes}
\usepackage{multirow}
\usepackage{supertabular}
\usepackage{booktabs,tabularx,array,ragged2e}
\usepackage{placeins}   
\usepackage{pifont}
\usepackage{makecell}
\usepackage{threeparttable}
\usepackage{titletoc}
\usepackage{siunitx}
\newcolumntype{C}[1]{>{\centering\arraybackslash}m{#1}} 
\usepackage{graphicx}   
\usepackage{ltxtable}
\usepackage{array}
\usepackage{hyperref}  
\usepackage[table]{xcolor} 

\newcommand{\std}[1]{\hspace{0.15em}{\fontsize{4}{5}\selectfont(#1)}}

\newcommand{\best}[1]{\cellcolor{blue!40}{#1}}
\newcommand{\second}[1]{\cellcolor{blue!30}{#1}}
\newcommand{\third}[1]{\cellcolor{blue!20}{#1}}
\newcommand{\fourth}[1]{\cellcolor{blue!10}{#1}}

\newcommand{\rankFirst}{\textcolor{blue!40}{first}}
\newcommand{\rankSecond}{\textcolor{blue!30}{second}}
\newcommand{\rankThird}{\textcolor{blue!20}{third}}
\newcommand{\rankFourth}{\textcolor{blue!10}{fourth}}

\newcommand{\vb}{ViroBench}

\newcommand{\grouphead}[2]{%
  \midrule
  \multicolumn{#2}{l}{\textit{#1}}\\
  \midrule
}

\usepackage[most]{tcolorbox}
\tcbset{
  promptbox/.style={
    colback=gray!3,
    colframe=gray!40,
    boxrule=0.4pt,
    arc=2pt,
    left=6pt,right=6pt,top=4pt,bottom=4pt,
    fontupper=\small,
  }
}

\begin{document}

\title{\vb: Benchmarking Nucleotide Foundation Models on Viral Genomics Tasks}


\author{Dongxin Ye}
\authornote{Equal contributions.}
\affiliation{%
  \institution{Shanghai Innovation Institute}
  \city{Shanghai}
  \country{China}}
\affiliation{%
  \institution{University of Electronic Science and Technology of China}
  \city{Chengdu}
  \country{China}}
\email{dongxinye@sii.edu.cn}

\author{Fang Hu}
\authornotemark[1]
\affiliation{%
  \institution{Shanghai Innovation Institute}
  \city{Shanghai}
  \country{China}}
\affiliation{%
  \institution{Fudan University}
  \city{Shanghai}
  \country{China}}
\email{fanghu@sii.edu.cn}

\author{Han Hu}
\authornotemark[1]
\affiliation{%
  \institution{Shanghai Artificial Intelligence Laboratory}
  \city{Shanghai}
  \country{China}}
\affiliation{%
  \institution{Fudan University}
  \city{Shanghai}
  \country{China}}
\email{huhan@pjlab.org.cn}

\author{Shu Hu}
\affiliation{%
  \department{Institute of Infection and Health}
  \institution{Fudan University}
  \city{Shanghai}
  \country{China}}
\affiliation{%
    \institution{Shanghai Sci-Tech Inno Center for Infection \& Immunity}
    \city{Shanghai}
  \country{China}}
\email{shu25@m.fudan.edu.cn}

\author{Yang Tan}
\affiliation{%
  \institution{Shanghai Innovation Institute}
  \city{Shanghai}
  \country{China}}
\affiliation{%
  \institution{Shanghai Jiao Tong University}
  \city{Shanghai}
  \country{China}}
\email{tanyang@sii.edu.cn}

\author{Wanli Ouyang}
\affiliation{%
  \institution{Shenzhen Loop Area Institute}
  \city{Shenzhen}
  \country{China}}
\affiliation{%
  \institution{Chinese University of Hong Kong}
  \city{Hong Kong}
  \country{China}}
\email{wanliouyang@slai.edu.cn}

\author{Stan Z. Li}
\affiliation{%
  \institution{Westlake University}
  \city{Hangzhou}
  \country{China}}
\email{stan.zq.li@westlake.edu.cn}

\author{Jie Cui}
\affiliation{%
  \department{Institute of Infection and Health}
  \institution{Fudan University}
  \city{Shanghai}
  \country{China}}
\affiliation{%
    \institution{Shanghai Sci-Tech Inno Center for Infection \& Immunity}
    \city{Shanghai}
  \country{China}}
\email{jiecui@fudan.edu.cn}

\author{Nanqing Dong}
\authornote{Corresponding author.}
\affiliation{%
  \institution{Shanghai Innovation Institute}
  \city{Shanghai}
  \country{China}}
\affiliation{%
  \institution{Shanghai Artificial Intelligence Laboratory}
  \city{Shanghai}
  \country{China}}
\email{nanqing.dong@sii.edu.cn}

\renewcommand{\shortauthors}{Ye et al.}

\begin{abstract}
Nucleotide sequences constitute the fundamental genetic basis of biological systems, rendering viral genomic analysis critical for biomedical advancement. Despite progress in biological foundation models, specifically nucleotide foundation models (NFMs), the field lacks a unified standard for viral genomics to facilitate community development and enforce biosecurity constraints. To address this, we introduce \vb, the first comprehensive and large-scale benchmark specifically designed for NFMs in viral settings. \vb~evaluates models across two critical dimensions: biological understanding and latent biosecurity risk, covering 18 diverse scenarios within 4 task types. Extensive evaluation of $66$ NFMs across diverse architectures yields three critical conclusions. Firstly, NFMs exhibit a performance degradation in biological understanding under phylogenetic and temporal shifts, indicating weak extrapolation capabilities. Secondly, generation tasks reveal a decoupling between statistical likelihood and biological functional validity, posing latent biosecurity risks. Thirdly, controlled ablation studies reveal that taxonomic diversity in pretraining data outweighs parameter scale. Specifically, a lightweight baseline trained on diverse data achieves a $67.5\%$ performance gain over its original model. Overall, \vb~provides interpretable, diagnostic evaluations and a reproducible measurement framework for future research on viral nucleotide foundation models. The datasets and code are publicly available at \url{https://github.com/QIANJINYDX/ViroBench}.

\end{abstract}






\maketitle

\section{Introduction}
\begin{figure*}[t]
  \centering
  \includegraphics[width=1\textwidth]{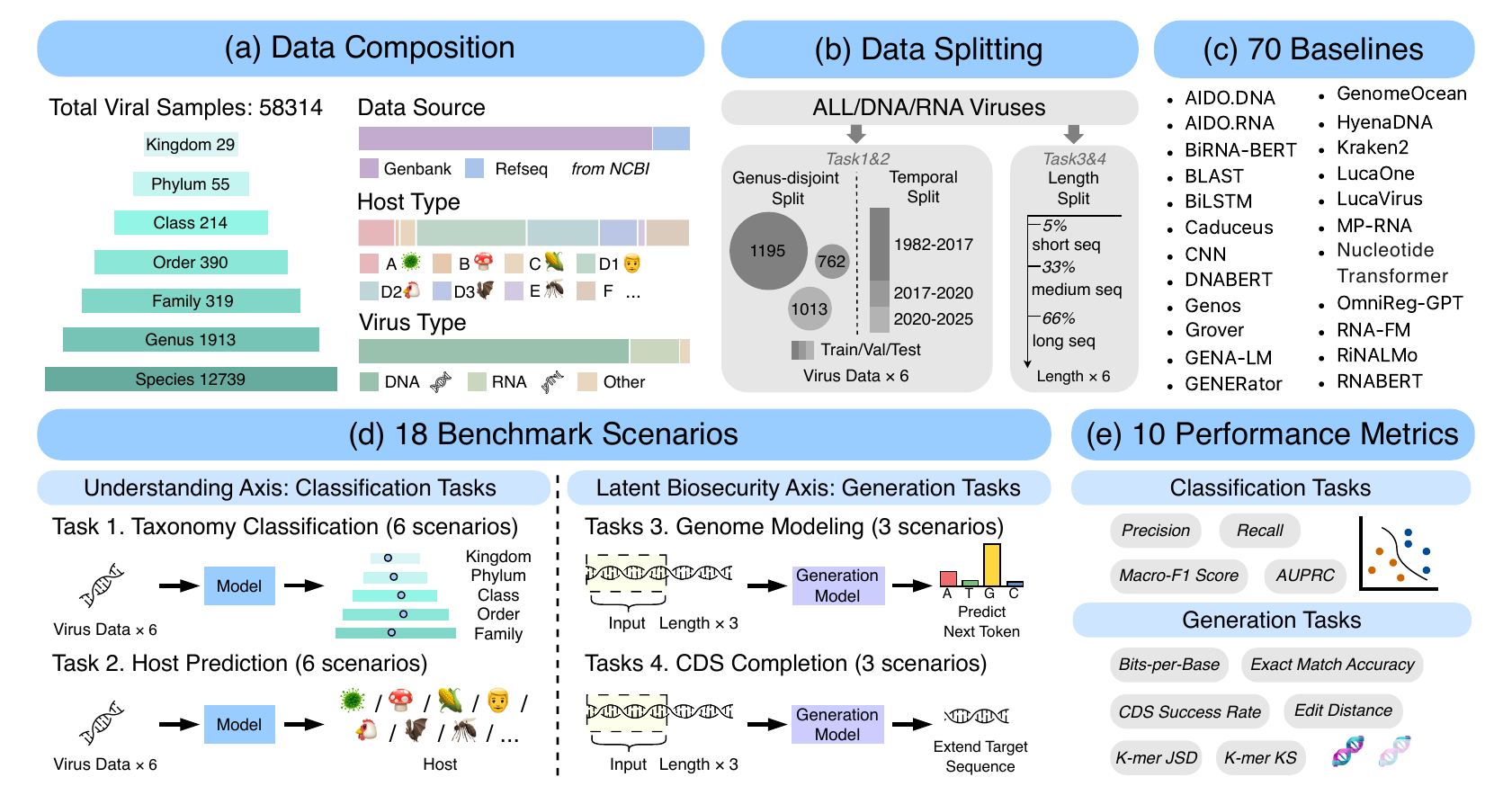}
  \caption{Overview of \vb. (a) \textbf{Data Composition}: \textbf{58,314} sequences featuring hierarchical taxonomy, host categories, and diverse nucleic acid types. (b) \textbf{Splitting}: Genus-disjoint and temporal axes for classification; length-based stratification for generation. (c) \textbf{Methods}: \textbf{70} baselines comprising 66 NFMs and 4 conventional baseline. (d) \textbf{Scenarios}: \textbf{18} scenarios spanning 4 task types. (e) \textbf{Metrics}: Evaluation metrics for discriminative and generative performance.}
  \label{fig:overview}
\end{figure*}

Nucleic acid sequences constitute the fundamental source code of life, underpinning biological structure and function across the biosphere~\cite{watson1953dna}. Within this genomic landscape, the investigation, surveillance, and risk assessment of viral sequences serve as a cornerstone of modern biomedical advancement. Unlike stable cellular genomes, viruses act as dynamic drivers of infectious disease and continuous evolution~\cite{sanjuan2010viral_mutation_rates}. Their capacity for rapid mutation and genetic recombination allows them to swiftly alter transmissibility, pathogenicity, and host range, creating persistent challenges for public health surveillance, vaccine development, and biosecurity governance~\cite{bowyer2025unveiling,holmes2013can}. Compared with general biological sequences, viral data exhibit extreme diversity, severe distribution shifts, and long-tailed structures. Substantial differences often arise across nucleic-acid types (DNA/RNA), phylogenetic levels, and temporal variants, making virus-related modeling both scientifically valuable and practically urgent~\cite{metsky2022designing,wheeler2025responsible,zerbini2023changes}.

Driven by advances in machine learning, computational virology has progressed rapidly. Early work used handcrafted features with classical classifiers (\emph{e.g.}, SVMs and random forests) for viral taxonomy and host prediction~\cite{purwono2024virus,perelygin2025effect}. Later, neural models (\emph{e.g.}, CNNs, RNNs, and Transformers) improved the modeling of sequence motifs and long-range dependencies~\cite{ren2020identifying,mock2021vidhop}. In recent years, general-purpose nucleotide foundation models (NFMs) have emerged and been widely adopted across diverse downstream tasks, offering possibilities for cross-species and cross-task transfer learning~\cite{nucleotidetransformer,evo2}. However, despite these advances, a standardized and reproducible benchmark tailored to viral scenarios remains notably lacking. Existing studies often evaluate on disparate datasets under splitting protocols (\emph{e.g.},~random splits that ignore phylogeny or temporal drift), which hinders fair comparisons and masks model failures in real-world generalization~\cite{li2025nabench}. The absence of a rigorous benchmark prevents the systematic measurement of both NFMs' comprehension of viral rules and their generation-related behaviors on viral sequences, including indicators that may be relevant to downstream risk assessment.

To bridge this gap, we introduce~\vb, the first comprehensive evaluation benchmark for NFMs in viral settings (Figure~\ref{fig:overview}). \vb~anchors evaluation to two key axes: (1) \textit{biological understanding} (classification), probing how models capture viral diversity, and (2) \textit{latent biosecurity risk} (generation), characterizing generation-related behaviors that may inform downstream biosecurity assessment. Built on a curated corpus of $58{,}314$ high-quality viral sequences, \vb~defines $4$ core types of tasks spanning 18 scenarios, including $12$ classification and $6$ generation tasks. 

For classification, we conduct a multi-scale evaluation starting from a comprehensive viral landscape to establish a performance baseline. To further investigate domain-specific nuances, we partition the data into DNA and RNA viral cohorts, analyzing the inherent distribution and separability differences across these major genomic groups. Furthermore, we propose two stringent protocols, \textit{Genus-disjoint Splits} for phylogenetic extrapolation and \textit{Temporal Splits} for robustness to evolutionary drift. 

For generation, we utilize genome modeling to test the consistency and stability of long-sequence completion, while CDS generation is employed to evaluate the models' capability to produce functional viral sequences with biological protein-coding potential. Beyond standard computational metrics on sequences, we further evaluate the biological significance of the generated results to ensure their functional relevance. Furthermore, by introducing length-bucketed evaluation, we explicitly characterize how modeling difficulty and potential risk signals evolve with sequence length, rendering the model’s capability boundaries and risk profiles more interpretable and diagnosable.

We benchmark $66$ NFMs, spanning diverse scales and pretraining paradigms. Our results reveal that NFMs exhibit a sharp performance degradation in biological understanding under phylogenetic and temporal shifts, indicating fragile extrapolation capabilities across the evolutionary landscape. For generation, we uncover a decoupling between statistical likelihood and functional validity; models often prioritize low perplexity over structural integrity, posing latent biosecurity risks. Furthermore, our results reveal that pretraining on diverse multi-species data is more effective for capturing viral genomic patterns than simply increasing model scale. Leveraging this insight, we developed a lightweight baseline that outperforms its much larger original version by 67.5\%, demonstrating that taxonomic diversity can outweigh parameter count.

Overall, our contributions are as follows:

\begin{itemize}
    \item We introduce~\vb, the first large-scale and comprehensive benchmark to explicitly unify discriminative understanding and latent biosecurity risk, providing a standardized environment to assess both the biological comprehension and the biosecurity risks of nucleotide models. 
    \item We conduct a large-scale evaluation of 66 NFMs, characterizing their behavioral traits across key biological dimensions.
    \item We provide ablation studies to validate our data composition insights, confirming that taxonomic diversity outweighs scale with a 67.5\% gain in our lightweight baseline.
\end{itemize}

\section{Related Works}

\subsection{Viral Sequence Analysis}
Viral sequence analysis has long revolved around two core questions: (i) assigning viruses to their taxonomic/phylogenetic ranks (\emph{e.g.}, order/family/genus) and (ii) inferring virus–host relationships such as host-range prediction and spillover risk assessment~\cite{raju2022virustaxo,mock2021vidhop}.Early pipelines typically relied on alignment/homology searches or handcrafted features (e.g., K-mer spectra, ORF statistics) combined with classical classifiers (e.g., SVM/Random Forest), which are often interpretable but can degrade when faced with novel, divergent, or database-sparse viruses~\cite{perelygin2025effect}.With the rise of deep learning, CNN/RNN/Transformer-style models have increasingly been used for end-to-end viral modeling across tasks such as taxonomy classification and host prediction~\cite{shang2021cheer}.However, evaluation is highly split-sensitive: random splits may place closely related (near-duplicate) sequences in both train and test, inflating generalization estimates, while the continual discovery of new viruses and their rapid evolution introduce temporal drift that challenges robustness to newly emerging variants~\cite{ferrer2024spanseq,perelygin2025effect,ren2017virfinder,ren2020identifying,sardanyes2024quasispecies}.These issues motivate biology-aware partitioning and more diagnostic evaluations beyond single-number rankings.

\subsection{Nucleotide Foundation Models}
Self-supervised pretrained NFMs have emerged as a dominant paradigm for genomic representation and generation~\cite{balakrishnan2025gene}. These models are typically pretrained on large-scale unlabeled sequences and transferred to diverse downstream tasks, including classification and base-level prediction ~\cite{nucleotidetransformer, ji2021dnabert, nguyen2023hyenadna}. Mainstream architectures include masked language models for discriminative tasks~\cite{ellington_accurate_2024, Zou2024.11.28.625345, Tahmid2024.07.02.601703,ji2021dnabert,zhou2023dnabert2,chen2022interpretable,penic2025rinalmo,akiyama2022informative}, as well as causal language models and sequence-to-sequence structures for generative modeling~\cite{nguyen2023hyenadna,wang2025omnireg,zhou2025genomeocean,10.1093/gigascience/giaf132,wu2025generator,nguyen2024sequence,merchant2025semantic,evo2}. Meanwhile, long-context mechanisms such as linear attention and state-space models have been widely explored to handle extensive genomic dependencies~\cite{zhou2023dnabert2, nguyen2023hyenadna, schiff2024caduceus}. In practice, tokenization strategies (\emph{e.g.}, K-mer, BPE, Single) and long-range modeling designs significantly dictate effective context length and cross-model comparability~\cite{lindsey2025impact, zhou2023dnabert2}. Given that viral genomes are fundamentally composed of nucleotide sequences, these powerful NFMs theoretically possess the potential to revolutionize viral research. However, there remains a conspicuous lack of systematic evaluation regarding their intrinsic capabilities in viral contexts. This evaluation gap leaves the models' generalization limits and associated biosecurity risks entirely uncharacterized.

\subsection{Benchmarks for Biological Sequence}
The biological modeling community has established diverse mature benchmarks for DNAs and proteins. Genomic Benchmarks~\cite{grevsova2023genomic} provides a foundational suite of classification tasks for consistent model comparison. BEND~\cite{marin2023bend} introduces a more specialized set of DNA functional annotations, while GenBench~\cite{liu2024genbench} offers a systematic diagnostic framework tailored for genomic foundation models. In the protein domain, specific benchmarks~\cite{dallago2021flip,NEURIPS2023_cac723e5,zhang2025venusmuthub} target large-scale mutation fitness prediction, whereas broader benchmarks~\cite{NEURIPS2023_d73078d4,xu2022peer} cover the spectrum from understanding to design. Despite their success, a unified evaluation ecosystem tailored to viruses remains underdeveloped, despite the importance of viral modeling for public health, surveillance, and responsible biotechnology. Existing virus-related efforts are predominantly fragmented and focus on specific tool-level applications, such as benchmarking metagenomic classifiers~\cite{glickman2021simulation} or taxonomic annotation pipelines~\cite{raju2022virustaxo}. These studies typically do not assess the underlying representation capabilities of foundation models, nor do they cover the critical challenges of phylogenetic generalization and temporal drift. This disparity underscores the necessity for specialized benchmarks to systematically evaluate the performance of NFMs in viral understanding and latent biosecurity risks.

\section{\vb~Construction}
\vb~centers on a unified viral corpus designed to evaluate models across two critical dimensions: biological understanding and latent biosecurity risk. This section outlines the data curation pipeline and the design principles governing our evaluation tasks.

\subsection{Data Curation}
We constructed the~\vb~corpus by systematically processing all known viral sequences to ensure biological grounding. The construction of \vb~began with the retrieval of $273,974$ virus-associated TaxIDs from NCBI (RefSeq~\cite{o2016reference} and GenBank~\cite{benson2012genbank}). For each entry, we integrated metadata across three key dimensions: (1) Taxonomy, by extracting hierarchical lineages (from \emph{Kingdom} to \emph{Genus}); (2) Chronology, by recording the earliest discovery dates; and (3) Host. To resolve the high entropy of raw host metadata, which originally contained over $8,000$ inconsistent strings, we used Qwen3-235B~\cite{yang2025qwen3} to standardize these labels into eight coarse-grained categories (\emph{e.g.}, \emph{Bacterial}, \emph{Plant}, \emph{Human}). To maintain data integrity, we applied a multi-stage filtering process: (i) removing entries with incomplete taxonomy or missing timestamps; (ii) retaining only verified, high-quality assemblies; and (iii) resolving species-level redundancies. The final \vb~corpus consists of 58,314 high-quality viral samples. A comprehensive breakdown of the curation pipeline, including the tie-breaking hierarchy and LLM prompting strategies, is provided in Appendix Section~\ref{app:Detailed Data Preprocessing}.

\subsection{Task Taxonomy and Instantiation}
\vb~establishes a multidimensional diagnostic framework derived from four core task types intersected with diverse Evaluation Regimes. This design systematically probes the boundaries of the performance of a model across two primary axes.

\subsubsection{Understanding Axis: Classification Tasks}
The Understanding Axis evaluates a model's capacity to internalize fundamental biological rules across $12$ diagnostic scenarios. To ground this evaluation in biological reality, we focused on two primary task types: \begin{itemize} \item \textbf{Taxonomy Classification}: Predicts five hierarchical labels (from \emph{Kingdom} to \emph{Family}). It evaluates the model's ability to recognize the conserved hierarchical structures that define viral evolution. \item \textbf{Host Prediction}: Categorizes sequences into standardized host classes to evaluate virus-host interaction patterns. This tests whether encoded representations capture functional ecological signals beyond internal genomics. \end{itemize} 

Tasks are structured across a global viral landscape (\textbf{ALL}) and two specific subsets (\textbf{DNA and RNA}), providing a multi-scale benchmark from universal understanding to specialized adaptation. By addressing the disparate mutation rates and evolutionary constraints of different nucleic-acid types, this setup evaluates how effectively pre-trained knowledge transfers from a broad viral context to specific replication strategies. To further explore model robustness, we evaluate performance across two rigorous data-splitting dimensions. A \textbf{Genus-disjoint Split} enforces strict taxonomical isolation to mandate phylogenetic extrapolation, ensuring performance reflects biological understanding rather than sequence memorization. In parallel, a \textbf{Temporal Split} partitions data chronologically to simulate real-world distribution drift, thereby challenging the model’s resilience against the rapid mutational drift and recombination characteristic of viral evolution. Detailed statistics for these partitioned datasets are provided in Table \ref{tab:dataset_stats_cls}, forming the shared basis for both classification tasks.


\subsubsection{Latent Biosecurity Axis: Generation Tasks}

The Axis evaluates potential safety risks through $6$ diagnostic scenarios. Specifically, we operationalize this assessment across two task types: \begin{itemize} \item \textbf{Genome Modeling}: Assesses sequence likelihood and stability of full-length genomic fragments to measure the model's capacity in capturing the global statistical landscape of viral genomes. This identifies risks associated with the assembly of plausible viral contigs. \item \textbf{CDS Generation}: Evaluates the capability to produce protein-coding sequences (CDS) given a partial prefix. It probes whether models can generate \emph{functional viral elements} that obey strict biological constraints, such as open reading frame (ORF) integrity and codon usage patterns. \end{itemize}

To further delineate performance boundaries, both tasks are stratified across three length regimes (Short, Medium, Long) defined by the 33rd and 66th percentiles of the sequence length distribution. We utilize all contigs for genome modeling to ensure comprehensive scale assessment. For CDS generation, we implement diversity-aware subsampling (limiting to 500 non-redundant sequences per host) to maintain a balanced representation of the viral landscape and prevent performance metrics from being skewed by over-represented species. This stratified approach serves as a diagnostic for error accumulation, revealing whether generative reliability remains robust or degrades as the model transitions from short biological motifs to long-range, functionally constrained genomic structures. Detailed statistics are provided in Table \ref{tab:dataset_stats_gen}.

\begin{table}[!t]
\centering
\caption{Dataset descriptions for classification tasks.}
\label{tab:dataset_stats_cls}
\renewcommand{\arraystretch}{1.1}
\resizebox{\columnwidth}{!}{
\begin{tabular}{lcc} 
\toprule
\textbf{Splitting Strategy} & \textbf{Information} & \textbf{Split Specifications} \\
\midrule
\rowcolor{gray!15} \multicolumn{3}{c}{\textbf{Panel A: ALL Viruses}} \\
\quad Genus-disjoint & 29 / 55 / 214 / 390 / 319 $^a$ & 8:1:1 Ratio (Train/Val/Test) \\
\quad Temporal Split   & 1982.06 $\rightarrow$ 2025.07 $^b$ & Cutoffs: 2017.10 / 2020.02 $^c$ \\
\midrule
\rowcolor{gray!15} \multicolumn{3}{c}{\textbf{Panel B: DNA Viruses}} \\
\quad Genus-disjoint & 9 / 14 / 20 / 51 / 160 $^a$  & 8:1:1 Ratio (Train/Val/Test) \\
\quad Temporal Split   & 1982.06 $\rightarrow$ 2024.09 $^b$ & Cutoffs: 2022.07 / 2023.08 $^c$ \\
\midrule
\rowcolor{gray!15} \multicolumn{3}{c}{\textbf{Panel C: RNA Viruses}} \\
\quad Genus-disjoint & 4 / 12 / 30 / 58 / 139 $^a$  & 8:1:1 Ratio (Train/Val/Test) \\
\quad Temporal Split   & 1982.06 $\rightarrow$ 2025.07 $^b$ & Cutoffs: 2017.03 / 2017.11 $^c$ \\
\bottomrule
\end{tabular}
} 
\vspace{2pt}
\begin{flushleft}
\footnotesize 
$^a$ \textbf{Hierarchy Count}: Num. labels at Kingdom/Phylum/Class/Order/Family.\\
$^b$ \textbf{Time Range}: Total span of collection. \quad $^c$ \textbf{Cutoffs}: Split points for Val/Test.
\end{flushleft}
\end{table}

\begin{table}[!t]
  \caption{Dataset descriptions for generation tasks.}
  \label{tab:dataset_stats_gen}
  \centering
  \small
  \setlength{\tabcolsep}{7pt}
  \renewcommand{\arraystretch}{1.02}

  \resizebox{\columnwidth}{!}{%
  \begin{tabular}{ccccc}
    \toprule
    \textbf{Source} & \textbf{Target} & \textbf{Strategy} & \textbf{Length range} & \textbf{Count} \\
    \midrule

    \multirow{3}{*}{Genome} & \multirow{3}{*}{\shortstack{Predict \\ Next Token}} & Short  & 855bp -- 1,440bp       & 43{,}040 \\
                            &                                              & Medium & 1,441bp -- 2,192bp     & 43{,}280 \\
                            &                                              & Long   & 2,193bp -- 1,385,869bp & 56{,}772 \\
    \midrule

    \multirow{3}{*}{CDS}    & \multirow{3}{*}{\shortstack{Sequence \\ Generation}}                      & Short  & 153bp -- 330bp         & 3{,}575 \\
                            &                                              & Medium & 333bp -- 765bp         & 5{,}495 \\
                            &                                              & Long   & 768bp -- 26,784bp      & 9{,}179 \\
    \bottomrule
  \end{tabular}%
  }
\end{table}

\section{Experiment}
\subsection{Experimental Setup}
\subsubsection{Baseline Models}
We extensively benchmarked 66 state-of-the-art NFMs, together with 4 conventional baseline, totaling 70 methods (Table~\ref{tab:short_model_summary}). Given that their pretraining corpora are highly heterogeneous and largely unoptimized for viral genomics, a simple aggregate ranking would fail to objectively reflect their true capability boundaries. To ensure a fairer diagnostic evaluation, we categorized models by their pretraining coverage lineage: \emph{Diverse Viral}, \emph{Phage-specific}, \emph{RNA-specific}, and \emph{Non-viral Coverage}. Architecturally, while all 66 NFMs serve as encoders for classification tasks, only those with autoregressive decoder architectures participated in generation evaluations. 

\begin{table}[!t]
  \centering
  \caption{Summary of $70$ methods in~\vb.}
  \label{tab:short_model_summary}
  \small 
  \setlength{\tabcolsep}{8pt} 
  \begin{threeparttable}
    \begin{tabular}{l c c c c}
      \toprule
      \textbf{Model Series} & \textbf{Lin.\tnote{*}} & \textbf{Cls.} & \textbf{Gen.} & \textbf{\# Models} \\
      \midrule
      AIDO.DNA~\cite{ellington_accurate_2024} & N & \ding{51} & \ding{55} & 2 \\
      AIDO.RNA~\cite{Zou2024.11.28.625345} & R & \ding{51} & \ding{51} & 4 \\
      BiRNA-BERT~\cite{Tahmid2024.07.02.601703} & R & \ding{51} & \ding{55} & 1 \\
      BLAST~\cite{camacho2009blast+} & - & \ding{51} & \ding{55} & 1 \\
      BiLSTM~\cite{schuster1997bidirectional} & D & \ding{51} & \ding{55} & 1 \\
      Caduceus~\cite{schiff2024caduceus} & N & \ding{51} & \ding{55} & 2 \\
      CNN~\cite{grevsova2023genomic} & D & \ding{51} & \ding{51} & 1 \\
      DNABERT(1~\cite{ji2021dnabert}/2~\cite{zhou2023dnabert2}) & N & \ding{51} & \ding{55} & 5 \\
      DNABERT-S~\cite{zhou2025dnabert} & D & \ding{51} & \ding{55} & 1 \\
      Evo (v1~\cite{nguyen2024sequence}/1.5~\cite{merchant2025semantic}/2~\cite{evo2}) & P & \ding{51} & \ding{51} & 8 \\
      GENA-LM~\cite{GENA_LM} & N & \ding{51} & \ding{55} & 3 \\
      GENERator v2~\cite{wu2025generator} & N & \ding{51} & \ding{51} & 4 \\
      Genos~\cite{10.1093/gigascience/giaf132} & N & \ding{51} & \ding{51} & 3 \\
      GenomeOcean~\cite{zhou2025genomeocean} & D & \ding{51} & \ding{51} & 3 \\
      Grover~\cite{sanabria2024dna} & N & \ding{51} & \ding{55} & 1 \\
      HyenaDNA~\cite{nguyen2023hyenadna} & N & \ding{51} & \ding{51} & 6 \\
      Kraken2~\cite{wood2019improved} & - & \ding{51} & \ding{55} & 1 \\
      LucaOne~\cite{he2025generalized} & D & \ding{51} & \ding{55} & 2 \\
      LucaVirus~\cite{pan2025predicting} & D & \ding{51} & \ding{55} & 2 \\
      MP-RNA~\cite{yang2024mp} & N & \ding{51} & \ding{55} & 1 \\
      NT (v1~\cite{nucleotidetransformer}/v2~\cite{nucleotidetransformer}) & N & \ding{51} & \ding{55} & 9 \\
      NT v3~\cite{boshar2025foundational} & P & \ding{51} & \ding{55} & 5 \\
      OmniReg-GPT~\cite{wang2025omnireg} & N & \ding{51} & \ding{51} & 1 \\
      RNA-FM~\cite{chen2022interpretable} & R & \ding{51} & \ding{55} & 1 \\
      RiNALMo~\cite{penic2025rinalmo} & R & \ding{51} & \ding{55} & 1 \\
      RNABERT~\cite{akiyama2022informative} & N & \ding{51} & \ding{55} & 1 \\
      \bottomrule
    \end{tabular}
    \begin{tablenotes}
      \footnotesize
      \item[*] \textbf{Pretraining Coverage Lineage}: (\textbf{D}) Diverse Viral Coverage; (\textbf{P}) Phage-specific Coverage; (\textbf{R}) RNA-specific Coverage; (\textbf{N}) Non-viral Coverage.
    \end{tablenotes}
  \end{threeparttable}
\end{table}

\subsubsection{Evaluation Protocol}
We establish a standardized evaluation protocol to ensure fair comparisons across models. Detailed formulations for all metrics are provided in Appendix Section \ref{subsec:Evaluation Metrics}.

\paragraph{\textbf{Taxonomy and Host Classification.}} 
To emulate real-world virus surveillance, we segment each viral genome into fixed-length, non-overlapping windows, including an extra window at the end to ensure tail coverage. 
This mimics the practical identification of viruses from localized genomic fragments. 
For training, we examine three distinct configurations: window sizes of 512, 1024, and 2048, paired with random sampling of 8, 4, and 2 windows per sequence, respectively. 
This strategy balances sample diversity with computational efficiency. 
During validation and testing, we evaluate all available windows and aggregate window-level predictions to obtain final sequence-level decisions. 
We extract embedding features from each model to train a standardized, lightweight classification head, minimizing biases from heterogeneous tokenizers and architectures. 
A CNN trained from scratch serves as a non-pretrained baseline under comparable settings. 
To address extreme class imbalances, we utilize a robust metric suite: Area Under the Precision-Recall Curve (AUPRC), Recall, Precision, and Macro-F1 score. 
We tune learning rates ($10^{-2}$ to $10^{-4}$) and evaluate performance across these distinct window configurations. 
Results are reported as the mean (standard deviation) across these hyperparameter settings to ensure a fair comparison.
\begin{table*}[t]
  \caption{Macro-F1 scores for viral taxonomy and host classification. Models are grouped by molecular modality and pretraining coverage lineage. Evaluation covers ALL, DNA, and RNA virus sets under Genus-disjoint (G-split) and Temporal (T-split) split strategies. Top-4 performers per column are highlighted in purple: \rankFirst, \rankSecond, \rankThird, and \rankFourth. Means (standard deviations) are reported. Extended results and the full suite of $70$ evaluated methods are provided in Appendix Section~\ref{app:Comprehensive Performance Metrics}.}
  \label{tab:classification_results}
  \centering
  \scriptsize
  \setlength{\tabcolsep}{3pt}
  \renewcommand{\arraystretch}{1}

  \resizebox{\textwidth}{!}{%
  \begin{tabular}{l*{12}{c}}
    \toprule
    \multirow{3}{*}{\textbf{Model Name}} &
    \multicolumn{4}{c}{\textbf{ALL Viruses}} &
    \multicolumn{4}{c}{\textbf{DNA Viruses}} &
    \multicolumn{4}{c}{\textbf{RNA Viruses}} \\
    \cmidrule(lr){2-5}\cmidrule(lr){6-9}\cmidrule(lr){10-13}
    &
    \multicolumn{2}{c}{\textbf{Taxonomy}} & \multicolumn{2}{c}{\textbf{Host}} &
    \multicolumn{2}{c}{\textbf{Taxonomy}} & \multicolumn{2}{c}{\textbf{Host}} &
    \multicolumn{2}{c}{\textbf{Taxonomy}} & \multicolumn{2}{c}{\textbf{Host}} \\
    \cmidrule(lr){2-3}\cmidrule(lr){4-5}\cmidrule(lr){6-7}\cmidrule(lr){8-9}\cmidrule(lr){10-11}\cmidrule(lr){12-13}
    &
    G-Split & T-Split & G-Split & T-Split &
    G-Split & T-Split & G-Split & T-Split &
    G-Split & T-Split & G-Split & T-Split \\

    \grouphead{Baseline}{13}
    BLAST & 47.67\std{0.00} & 41.22\std{0.00} & \best{92.50\std{0.00}} & \second{65.55\std{0.00}} & 75.68\std{0.00} & 39.91\std{0.00} & \best{75.42\std{0.00}} & 25.81\std{0.00} & 59.65\std{0.00} & \best{75.74\std{0.00}} & \best{93.01\std{0.00}} & \best{79.09\std{0.00}} \\
    Kraken2 & 26.78\std{0.00} & 34.93\std{0.00} & 61.70\std{0.00} & \best{69.41\std{0.00}} & 52.62\std{0.00} & 34.12\std{0.00} & \second{67.05\std{0.00}} & 35.71\std{0.00} & 39.36\std{0.00} & \third{71.46\std{0.00}} & 40.49\std{0.00} & \second{65.52\std{0.00}} \\
    BiLSTM & 66.05\std{1.89} & \fourth{54.67\std{2.27}} & \third{84.40\std{0.98}} & 44.69\std{1.31} & 69.67\std{3.25} & 57.79\std{2.76} & \third{62.90\std{7.82}} & \best{56.48\std{0.80}} & 73.96\std{3.79} & 57.43\std{1.77} & \second{81.56\std{0.57}} & \third{65.11\std{2.04}} \\
    CNN & 34.72\std{10.96} & 19.26\std{13.92} & 69.29\std{2.51} & 25.16\std{5.52} & 26.63\std{20.35} & 21.45\std{5.73} & 39.87\std{6.87} & 32.62\std{5.47} & 32.07\std{21.49} & 34.81\std{4.19} & 60.46\std{7.49} & 40.71\std{13.21} \\
    \grouphead{DNA Foundation Models (Diverse Viral Coverage)}{13}
    DNABERT-S & 65.96\std{2.52} & 47.57\std{2.87} & 80.17\std{0.74} & 47.41\std{0.88} & 75.95\std{1.87} & 57.70\std{3.50} & 57.97\std{8.96} & \fourth{45.67\std{7.83}} & 75.55\std{2.43} & 57.12\std{2.41} & \fourth{77.88\std{2.44}} & 52.50\std{12.38} \\
    GenomeOcean-4B & \second{71.53\std{3.08}} & 52.28\std{4.69} & 81.67\std{0.94} & 48.75\std{1.14} & \third{79.60\std{2.58}} & 58.55\std{3.93} & 56.73\std{1.74} & 44.84\std{1.28} & \fourth{80.72\std{2.31}} & 59.41\std{3.04} & 72.54\std{4.11} & 44.13\std{8.64} \\
    LucaOne-Default-Step36M & \fourth{69.79\std{3.57}} & \second{57.45\std{3.41}} & \fourth{81.97\std{0.66}} & 47.52\std{0.39} & \second{80.40\std{2.33}} & \second{68.84\std{3.63}} & 58.35\std{0.85} & \third{46.40\std{0.54}} & \second{83.79\std{1.79}} & \fourth{67.56\std{4.17}} & 65.55\std{5.67} & 49.85\std{3.99} \\
    LucaVirus-Default-Step3.8M & \best{75.88\std{2.76}} & \best{64.91\std{3.33}} & \second{84.56\std{1.28}} & \fourth{54.84\std{1.54}} & \best{82.20\std{3.00}} & \best{69.17\std{4.36}} & 58.62\std{2.33} & 43.93\std{1.39} & \best{85.83\std{1.54}} & \second{73.28\std{2.34}} & 74.28\std{1.53} & 50.91\std{6.96} \\
    \grouphead{DNA Foundation Models (Phage-specific Coverage)}{13}
    Evo1-131K & 39.97\std{2.47} & 28.02\std{3.61} & 71.87\std{0.46} & 35.48\std{2.51} & 52.38\std{3.09} & 49.78\std{2.91} & 56.00\std{0.68} & 43.27\std{1.79} & 43.76\std{2.17} & 31.93\std{1.10} & 67.44\std{2.86} & 51.01\std{1.45} \\
    Evo1.5-8K & 39.96\std{2.74} & 27.68\std{2.27} & 71.38\std{0.32} & 35.91\std{1.14} & 47.45\std{3.67} & 41.50\std{3.89} & 56.61\std{0.75} & 42.65\std{1.93} & 40.54\std{4.22} & 31.52\std{1.89} & 64.05\std{3.51} & 50.27\std{3.88} \\
    Evo2-40B & 58.48\std{1.94} & 51.33\std{2.67} & 81.27\std{0.58} & 45.71\std{1.19} & 63.83\std{2.11} & \fourth{59.62\std{5.73}} & 61.35\std{3.72} & \second{49.62\std{5.71}} & 66.26\std{4.26} & 54.09\std{2.31} & \third{79.76\std{1.14}} & \fourth{62.95\std{1.97}} \\
    NTv3-650M-Post & 57.26\std{6.35} & 37.77\std{6.89} & 77.12\std{2.13} & 36.72\std{2.26} & 66.22\std{3.78} & 46.01\std{2.94} & 55.39\std{5.19} & 39.36\std{0.73} & 68.32\std{2.22} & 47.12\std{3.77} & 63.06\std{10.41} & 35.70\std{3.04} \\
    \grouphead{DNA Foundation Models (Non-viral Coverage)}{13}
    AIDO.DNA-7B & \third{69.87\std{3.05}} & \third{55.27\std{6.59}} & 80.65\std{1.23} & 47.47\std{1.38} & \fourth{79.37\std{2.04}} & \third{63.52\std{4.08}} & 56.61\std{2.01} & 45.13\std{1.25} & \third{81.28\std{2.17}} & 64.64\std{4.66} & 62.90\std{0.69} & 40.72\std{6.88} \\
    Caduceus-PS & 33.56\std{6.61} & 18.04\std{3.17} & 54.57\std{5.57} & 19.54\std{2.97} & 36.78\std{3.11} & 16.90\std{8.42} & 44.42\std{2.67} & 16.11\std{0.00} & 46.68\std{5.25} & 31.39\std{2.20} & 37.34\std{3.58} & 12.45\std{14.80} \\
    DNABERT-2-117M & 35.58\std{5.46} & 16.24\std{4.41} & 49.48\std{1.56} & 9.02\std{8.07} & 40.06\std{6.51} & 24.97\std{6.53} & 43.30\std{3.24} & 26.16\std{9.15} & 49.97\std{4.13} & 31.02\std{5.24} & 34.87\std{1.90} & 3.91\std{0.00} \\
    DNABERT-6 & 37.28\std{2.47} & 19.87\std{2.82} & 61.97\std{1.87} & 24.60\std{2.21} & 39.16\std{4.30} & 26.77\std{4.12} & 45.32\std{9.49} & 35.11\std{2.11} & 40.71\std{3.70} & 30.20\std{1.33} & 49.40\std{3.81} & 29.56\std{1.87} \\
    Genos-10B & 18.06\std{15.67} & 10.67\std{10.03} & 56.49\std{1.28} & 5.54\std{9.15} & 18.87\std{15.36} & 8.75\std{0.15} & 32.28\std{7.72} & 16.11\std{0.00} & 39.66\std{10.05} & 12.68\std{3.00} & 40.10\std{10.76} & 3.91\std{0.00} \\
    GENA-LM-bert-large-t2t & 59.62\std{4.61} & 38.65\std{6.90} & 77.55\std{2.81} & 39.83\std{4.86} & 69.48\std{3.13} & 51.08\std{2.99} & 52.39\std{0.82} & 37.98\std{1.25} & 70.24\std{2.01} & 52.99\std{3.97} & 57.70\std{6.85} & 35.65\std{4.77} \\
    GROVER & 44.35\std{6.49} & 22.21\std{3.37} & 66.98\std{3.55} & 25.55\std{0.78} & 50.14\std{1.47} & 31.41\std{6.01} & 46.95\std{3.14} & 34.93\std{2.18} & 58.63\std{2.91} & 38.49\std{2.97} & 42.38\std{2.19} & 26.56\std{1.42} \\
    GENERator-v2-prokaryote-3B & 9.18\std{5.59} & 4.66\std{4.77} & 25.89\std{12.32} & 0.39\std{0.24} & 6.23\std{1.37} & 9.32\std{0.60} & 14.55\std{4.01} & 16.11\std{0.00} & 11.92\std{1.69} & 10.84\std{1.68} & 7.02\std{0.00} & 3.91\std{0.00} \\
    HyenaDNA-large-1M & 18.12\std{13.22} & 12.45\std{5.99} & 53.64\std{7.84} & 5.65\std{8.99} & 28.35\std{7.82} & 12.36\std{2.23} & 35.62\std{6.77} & 16.11\std{0.00} & 34.09\std{7.21} & 19.91\std{2.38} & 41.55\std{4.36} & 3.91\std{0.00} \\
    NT-2.5B-ms & 24.10\std{11.17} & 13.49\std{6.25} & 52.09\std{14.56} & 21.18\std{3.47} & 31.37\std{17.06} & 22.83\std{3.81} & 41.13\std{2.31} & 40.47\std{8.77} & 29.09\std{19.42} & 24.76\std{8.71} & 31.35\std{2.16} & 3.91\std{0.00} \\
    NTv2-500M-ms & 38.27\std{16.04} & 26.16\std{15.57} & 60.35\std{21.56} & 24.17\std{20.84} & 40.60\std{23.51} & 30.77\std{19.40} & 49.50\std{2.10} & 35.73\std{17.16} & 38.47\std{21.15} & 33.02\std{15.62} & 45.37\std{35.42} & 35.67\std{27.53} \\
    OmniReg-GPT & 22.82\std{9.86} & 13.43\std{5.46} & 60.53\std{6.44} & 18.15\std{4.24} & 27.21\std{12.64} & 19.50\std{6.31} & 36.76\std{1.49} & 28.97\std{5.23} & 23.57\std{8.55} & 21.24\std{4.33} & 43.53\std{7.83} & 17.69\std{23.88} \\
    \grouphead{RNA Foundation Models (RNA-specific Coverage)}{13}
    AIDO.RNA-1.6B-CDS & 60.84\std{6.35} & 43.18\std{5.64} & 77.74\std{1.30} & 38.19\std{1.97} & 69.71\std{3.74} & 51.35\std{4.84} & 53.31\std{2.64} & 44.74\std{0.83} & 74.10\std{2.83} & 51.40\std{3.10} & 68.17\std{3.83} & 29.98\std{2.18} \\
    BiRNA-BERT & 33.34\std{6.78} & 17.19\std{3.99} & 64.71\std{1.59} & 21.00\std{3.82} & 42.55\std{4.87} & 23.75\std{2.79} & 42.53\std{0.63} & 35.62\std{1.91} & 47.17\std{4.53} & 24.21\std{2.41} & 40.93\std{0.60} & 9.94\std{10.44} \\
    RNA-FM & 48.67\std{14.24} & 19.88\std{9.81} & 67.87\std{9.89} & 25.59\std{4.37} & 58.15\std{5.67} & 27.49\std{13.48} & 46.27\std{0.44} & 30.19\std{12.40} & 59.41\std{7.18} & 35.70\std{14.03} & 45.39\std{7.70} & 12.32\std{14.57} \\
    RiNALMo & 46.70\std{11.13} & 28.15\std{11.13} & 61.64\std{1.32} & 23.84\std{8.19} & 53.35\std{3.03} & 31.37\std{6.57} & 49.71\std{0.71} & 37.68\std{5.29} & 52.75\std{7.15} & 38.16\std{6.45} & 47.67\std{0.72} & 19.63\std{9.44} \\
    \grouphead{RNA Foundation Models (Non-viral Coverage)}{13}
    MP-RNA & 54.00\std{6.01} & 35.24\std{7.31} & 77.14\std{1.50} & 36.61\std{0.11} & 63.63\std{3.88} & 46.72\std{3.96} & 49.78\std{0.25} & 44.32\std{1.33} & 69.73\std{4.29} & 48.32\std{3.94} & 57.44\std{4.31} & 34.68\std{3.74} \\
    RNABERT & 9.83\std{1.32} & 6.38\std{0.70} & 44.35\std{1.37} & 15.98\std{1.59} & 14.84\std{2.23} & 10.80\std{1.87} & 36.31\std{2.12} & 20.97\std{1.18} & 17.81\std{1.13} & 15.89\std{0.53} & 36.83\std{2.88} & 24.76\std{1.73} \\
    \grouphead{In-house Models}{13}
    ViroHyena-1M & 36.16\std{3.48} & 20.19\std{3.66} & 60.88\std{4.09} & 21.04\std{1.49} & 39.55\std{3.91} & 27.66\std{3.31} & 48.15\std{0.93} & 36.03\std{2.05} & 48.33\std{2.71} & 30.84\std{3.99} & 46.07\std{2.83} & 26.39\std{1.64} \\
    ViroHyena-253M & 51.03\std{3.36} & 33.97\std{4.24} & 65.33\std{4.07} & 30.70\std{2.62} & 54.78\std{4.25} & 35.95\std{3.44} & 44.43\std{0.89} & 40.42\std{1.82} & 63.29\std{4.30} & 44.24\std{3.97} & 40.69\std{1.88} & 31.19\std{0.65} \\
    ViroDNABERT2 & 53.72\std{2.85} & 32.43\std{2.22} & 77.57\std{0.74} & \third{56.73\std{0.80}} & 59.02\std{6.55} & 30.03\std{2.03} & \fourth{62.35\std{3.92}} & 38.95\std{3.95} & 73.79\std{2.50} & 41.25\std{3.33} & 70.21\std{7.00} & 44.90\std{1.71} \\
    ViroCaduceus & 58.43\std{2.42} & 41.75\std{5.05} & 70.90\std{0.50} & 50.13\std{1.42} & 58.37\std{1.37} & 31.95\std{2.68} & 47.70\std{0.48} & 39.47\std{1.07} & 73.79\std{2.74} & 39.49\std{2.87} & 63.09\std{9.67} & 38.44\std{2.34} \\

    \bottomrule
  \end{tabular}%
  }
\end{table*}

\begin{figure*}[t]
  \centering
  \includegraphics[width=1\textwidth]{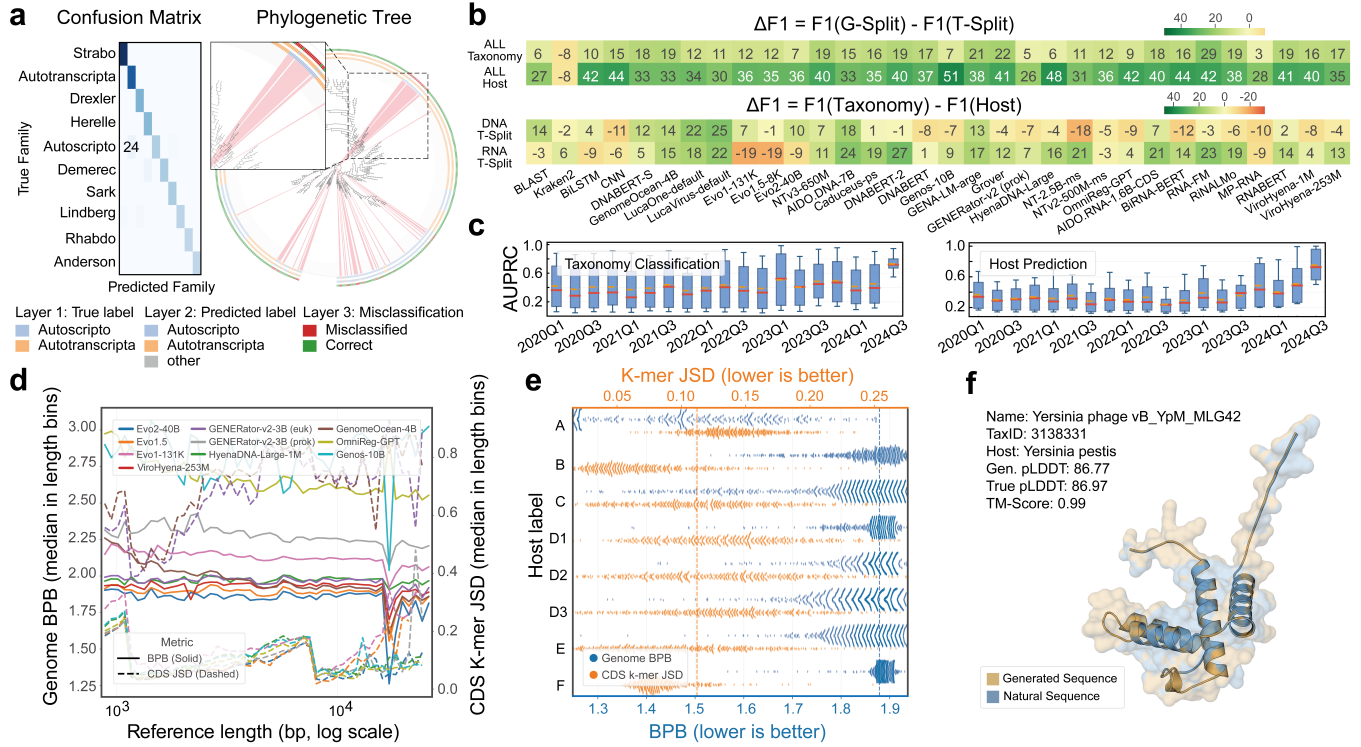}
  \caption{Experimental results. (a) Family-level confusion matrix (left) and phylogenetic tree (right) for AIDO.DNA-7B, visualizing misclassifications within \emph{Autoscriptoviridae} and \emph{Autotranscriptaviridae} lineages. (b) Comparative performance analysis ($\Delta$F1). (Top) Generalization gap between G-Split and T-Split for taxonomy and host classification tasks. (Bottom) Task-wise performance delta between taxonomy and host classification under DNA/RNA T-Splits. (c) Mean AUPRC trends for taxonomy and host classification across all models. (d) Comparison of BPB and K-mer JSD across identical sequence lengths. (e) Honeycomb density plot for Evo2-40B, superimposing genome BPB (blue, bottom axis) and K-mer JSD (orange, top axis). (f) AlphaFold3 (AF3) superimposition of the generated CDS structure (orange) and its natural counterpart (blue) for an example sequence from YpM\_MLG42.}
  \label{fig:experiment_result}
\end{figure*}

\paragraph{\textbf{Genome Modeling.}} We evaluate the model's ability to assign probabilities to the true sequence distribution at the ``next-token'' level. Using a fixed 128-bp prompt as model input, we compute the average negative log-likelihood (NLL) and convert perplexity to bits-per-base (BPB). BPB quantifies the average information (in bits) required to generate a single base, accounting for the specific number of tokens and bases utilized. Consequently, a lower BPB signifies more accurate next-step prediction and a closer alignment between the model's generative distribution and real viral sequences.

\paragraph{\textbf{CDS Generation.}} We evaluate the model's ability to complete protein-coding regions by providing a 129-bp prompt (aligning with triplet codons) as a prefix. Our evaluation framework distinguishes between sequence-level replication and biological function. We first quantify literal fidelity to the ground-truth using Exact Match Accuracy and Edit Distance. However, since verbatim mimicry does not imply functional integrity, we introduce the CDS Success Rate to verify frame consistency and the absence of internal stop codons. Finally, we employ K-mer distribution analysis (JSD and KS statistics) to ensure the generation respects the global statistical properties and codon usage bias of natural viral genomes. This tripartite approach effectively separates surface-level similarity from structural and biological authenticity.

\subsection{Main Results}
\subsubsection{Classification Tasks}
As shown in Table \ref{tab:classification_results}, classification performance is primarily driven by pretraining data coverage rather than raw parameter scaling. NFMs with diverse viral exposure consistently outperform much larger non-viral models. However, Evo2-40B demonstrates that massive scaling enables general-purpose models to internalize deep evolutionary patterns, achieving Macro-F1 scores that rival or even surpass those of models specifically optimized for viral data. Furthermore, AIDO.DNA-7B remains highly competitive despite the total absence of viral sequences in its training set. This suggests that large-scale metagenomic pretraining allows the model to capture implicit viral signals by learning from endogenous viral elements embedded within host genomes, which provide a latent template of viral architecture. 

We also observe that non-foundation baselines can be competitive in several settings. BLAST outperforms some NFMs on certain tasks, likely because similarity-based methods can directly exploit close sequence matches in the reference database. BiLSTM also surpasses some NFMs in specific cases, suggesting that simpler supervised sequence models may remain robust when pretrained NFMs suffer from limited viral coverage or domain mismatch. These results highlight that NFMs and conventional baselines rely on different inductive biases, and that larger pretrained models do not automatically guarantee better performance in viral classification.

To diagnose the failure modes, we projected the family-level confusion matrix of AIDO.DNA-7B onto a circular phylogenetic tree (Figure~\ref{fig:experiment_result}a). Evidence indicates that misclassifications are not stochastic; instead, they are heavily clustered within phylogenetically neighboring clades, such as the Autoscripto and Autotranscripta lineages. This pattern suggests current models capture coarse-grained evolutionary signals but lack the resolution needed to distinguish pathogens with fine-grained divergence, leading to a systemic collapse when forced to extrapolate beyond their training horizon. Additional phylogenetic analyses are available in Appendix~\ref{app:phylo_trees}.

Furthermore, we conducted a high-resolution visualization analysis across 12 diagnostic variants (full results are provided in Appendix~\ref{app:hotmap}). Our analysis reveals two critical insights regarding model robustness. \textbf{First}, models encounter a substantial performance gap when faced with realistic viral evolution. As shown in Figure \ref{fig:experiment_result}b top, nearly all models exhibit a pronounced decline in performance moving from the Genus-disjoint (G-split) to the Temporal (T-split) setting. In host classification, Macro-F1 scores frequently drop by over 50\% under temporal drift. For instance, Genos-10B achieves a competitive 56.49 for host prediction on the ALL-virus set under the genus-disjoint split but collapses to 5.54 under the temporal split. This precipitous decline highlights a fundamental vulnerability to mutational drift. This temporal decay is further nuanced by the longitudinal trends observed in Figure~\ref{fig:experiment_result}c, where the mean AUPRC across all models exhibits a clear upward trajectory, with models demonstrating significantly higher predictive accuracy on viral sequences discovered closer to the present day. \textbf{Second}, we observed a fundamental divergence in task difficulty between DNA and RNA viruses (Figure \ref{fig:experiment_result}b bottom). While RNA viruses exhibit a strong phylogenetic signal that favors Taxonomy Classification, DNA viruses present a more complex landscape where Host Prediction occasionally surpasses taxonomy in robustness, particularly under temporal shifts. This reflects a profound asymmetry in viral architecture, suggesting that biological understanding follows distinct logic for DNA and RNA entities.

\begin{table}[!t]
  \caption{BPB results on Genome Modeling across different length buckets (lower is better).}
  \label{tab:bpb_genome_buckets}
  \centering
  \small
  \setlength{\tabcolsep}{2pt}
  \renewcommand{\arraystretch}{1}

  \begin{tabular}{@{}>{\raggedright\arraybackslash}p{0.30\columnwidth}ccc@{}}
    \toprule
    \textbf{Model Name} & \textbf{Genome-Short} & \textbf{Genome-Medium} & \textbf{Genome-Long} \\
    \midrule
    Evo1-131K & 2.1739 & 2.1890 & 2.1341 \\
    Evo1.5 & 1.9230 & 1.9035 & 1.8772 \\
    Evo2-40B & 1.9010 & 1.8651 & 1.8660 \\
    HyenaDNA-Large-1M & 1.9693 & 1.9694 & 1.9625 \\
    Genos-10B & 5.3644 & 5.6987 & 5.4351 \\
    GenomeOcean-4B & 2.2308 & 2.0854 & 1.9649 \\
    GENERator-v2-3B\textsuperscript{*} & 2.3108 & 2.3832 & 2.3647 \\
    OmniReg-GPT & 2.9462 & 2.7808 & 2.6508 \\
    ViroHyena-1M & 1.9546 & 1.9480 & 1.9458 \\
    ViroHyena-253M & 1.9346 & 1.9483 & 1.9137 \\
    \bottomrule
    \multicolumn{4}{@{}l@{}}{\footnotesize \textsuperscript{*}Abbreviated name for GENERator-v2-Prokaryote-3B.} \\
    
  \end{tabular}
\end{table}
\newcommand{\up}{\,$\uparrow$}
\newcommand{\down}{\,$\downarrow$}

\begin{table*}[t]
  \caption{Results on CDS generation across length buckets. Exact Match Accuracy and CDS Success Rate are reported in \% ; Edit Distance, K-mer JSD, and K-mer KS are unitless. Top-1/2/3/4 per column are highlighted (dark-to-light purple). Lower is better for Edit Distance/K-mer JSD/K-mer KS; higher is better for Exact Match/CDS Success.}
  \label{tab:cds_gen_results}
  \centering
  \small
  \setlength{\tabcolsep}{2pt}
  \renewcommand{\arraystretch}{1}

  \begin{tabular*}{\textwidth}{@{\extracolsep{\fill}}l*{15}{c}@{}}
    \toprule
    \textbf{Model Name} &
    \multicolumn{5}{c}{\textbf{CDS-Short}} &
    \multicolumn{5}{c}{\textbf{CDS-Medium}} &
    \multicolumn{5}{c}{\textbf{CDS-Long}} \\
    \cmidrule(lr){2-6}\cmidrule(lr){7-11}\cmidrule(lr){12-16}

    & \textbf{Edit $\downarrow$}
    & \textbf{Match $\uparrow$}
    & \textbf{JSD $\downarrow$}
    & \textbf{KS $\downarrow$}
    & \textbf{Succ. $\uparrow$}
    & \textbf{Edit $\downarrow$}
    & \textbf{Match $\uparrow$}
    & \textbf{JSD $\downarrow$}
    & \textbf{KS $\downarrow$}
    & \textbf{Succ. $\uparrow$}
    & \textbf{Edit $\downarrow$}
    & \textbf{Match $\uparrow$}
    & \textbf{JSD $\downarrow$}
    & \textbf{KS $\downarrow$}
    & \textbf{Succ. $\uparrow$} \\
    \midrule
    Evo1-131K
      & 0.5784 & \fourth{26.29} & 0.2155 & 0.2280 & 0.7273
      & 0.5593 & 25.88 & 0.1986 & 0.2266 & \second{0.3822}
      & 0.5577 & 25.15 & 0.1675 & 0.2101 & \second{0.0436} \\
    Evo1.5
      & \third{0.5521} & \second{26.82} & \third{0.1563} & \third{0.1331} & 0.5315
      & \third{0.5326} & \second{26.42} & \third{0.1247} & \best{0.1139} & \third{0.2548}
      & \second{0.5235} & \best{26.15} & \best{0.1049} & \best{0.1021} & 0.0109 \\
    Evo2-40B
      & \best{0.5469} & \best{27.30} & \second{0.1525} & \second{0.1310} & \best{1.4270}
      & \second{0.5293} & \best{26.69} & \second{0.1243} & \third{0.1151} & \best{0.6005}
      & \best{0.5218} & \second{26.15} & \second{0.1076} & \second{0.1063} & \best{0.0545} \\
    HyenaDNA-large-1M
      & 0.5578 & 26.14 & 0.1649 & 0.1425 & \fourth{0.8392}
      & 0.5403 & 25.83 & 0.1404 & 0.1245 & 0.0182
      & \fourth{0.5317} & \fourth{25.72} & 0.1295 & 0.1228 & 0.0000 \\
    Genos-10B
      & 0.5607 & 26.06 & 0.1719 & 0.1510 & 0.6434
      & 0.5430 & 25.74 & 0.1470 & 0.1324 & 0.0546
      & 0.5364 & 25.58 & 0.1342 & 0.1313 & 0.0109 \\
    GenomeOcean-4B
      & 0.5718 & 25.94 & 0.3328 & 0.3191 & 0.3077
      & 0.5600 & 25.81 & 0.4446 & 0.4371 & 0.0728
      & 0.5652 & \third{25.84} & 0.5964 & 0.6348 & \fourth{0.0218} \\
    GENERator-v2-3B\textsuperscript{*}
      & \second{0.5481} & \third{26.59} & \best{0.1508} & \best{0.1261} & \third{0.8951}
      & \best{0.5291} & \third{26.18} & \best{0.1237} & \fourth{0.1183} & 0.0182
      & \third{0.5244} & 25.52 & \fourth{0.1191} & \fourth{0.1218} & \third{0.0327} \\
    OmniReg-GPT
      & 0.5685 & 25.43 & 0.1604 & 0.1372 & 0.8112
      & 0.5451 & 25.41 & \fourth{0.1289} & \second{0.1149} & 0.0546
      & 0.5335 & 25.39 & \third{0.1151} & \third{0.1093} & 0.0000 \\
    ViroHyena-1M
      & \fourth{0.5564} & 25.83 & \fourth{0.1588} & \fourth{0.1369} & 0.8112
      & \fourth{0.5380} & 25.66 & 0.1326 & 0.1236 & 0.0364
      & \textemdash & \textemdash & \textemdash & \textemdash & \textemdash \\
    ViroHyena-253M
      & 0.5571 & 26.15 & 0.1596 & 0.1394 & \second{1.0070}
      & 0.5385 & \fourth{26.00} & 0.1369 & 0.1253 & \fourth{0.0910}
      & \textemdash & \textemdash & \textemdash & \textemdash & \textemdash \\
    \midrule
    \multicolumn{16}{@{}l@{}}{\footnotesize \textsuperscript{*}Abbreviated name for GENERator-v2-Prokaryote-3B.} \\
  \end{tabular*}
\end{table*}

\subsubsection{Generation Tasks}
To understand the model's capability in generation tasks, we first examine whether generation difficulty is sensitive to input length to assess potential structural drift in generative difficulty. Then we stratify by host to examine whether capabilities are concentrated in specific niches or host categories, thereby identifying potential risk-related subgroups.

Within the overlapping length range of the two tasks (Figure~\ref{fig:experiment_result}d), BPB varies only mildly with length for most models. This suggests that per-base predictability of genomic fragments is not driven by length alone, but is more likely determined by heterogeneity in lineage composition, fragment provenance, and assembly fragmentation. In contrast, JSD shows clearer length sensitivity in some models, indicating that compositional constraints in coding sequences---such as codon preference, amino-acid composition, and functional motifs---are harder to maintain stably when generating longer sequences. Crucially, strong BPB does not necessarily imply low JSD. Evo2-40B and Evo1.5 achieve great performance on both BPB and JSD. However, GenomeOcean-4B and GENERator-v2-3B (euk) exhibit substantially elevated JSD despite non-worst BPB, demonstrating a typical decoupling in which likelihood-level fit remains acceptable while local compositional distributions deviate markedly. This pattern suggests that some models capture coarse-grained genomic statistics (e.g., overall nucleotide composition and low-order repeat patterns) but fail to preserve finer, functionally relevant $K$-mer constraints at the coding-fragment level.

To relate host types to the generatability reflected by the two metrics, we perform a host-stratified analysis (using Evo2-40B, the best overall performer, as a representative model). Specifically, we first aggregated multiple samples of the same virus at the taxid level by averaging the three buckets, thereby reducing the amplification of host bias caused by duplicate counting of the same virus. Then, we calculated the median and interquartile range within each host (Figure~\ref{fig:experiment_result}e). We found that the preference structure of host categories in terms of BPB and JSD is not entirely consistent. The median BPB for D1 (humans and primates) is approximately 1.823, relatively low, while the median JSD is 0.138, moderate among all models, indicating higher explainability at the genomic-statistical level with non-negligible but not worst coding-level deviations. Category B (fungi/oomycetes/plant pathogens) stands out with particularly low JSD and a narrow interquartile range, indicating more consistent and stable $K$-mer statistics across generations. Category F (others) also has a low JSD value, but its extremely narrow BPB distribution may reflect the concentration of lineage or data sources rather than a causal effect of host type. Overall, the purpose of host analysis is not to give a simple conclusion about which host is dangerous, but to identify which host categories are more likely to simultaneously meet the conditions of low global fit and low local statistical fidelity, thus corresponding to higher availability and risk windows that require priority consideration in different application scenarios. Of course, even if its JSD is not the lowest globally, a stable low BPB or small dispersion still suggests stronger statistical generativeness, which is worth paying attention to in risk control and capability assessment.

\subsubsection{Structure}

We evaluated whether generated coding sequences preserve protein-level structural constraints by comparing AlphaFold3 predicted structures of generated sequences against their matched ground-truth counterparts~\cite{Abramson2024}. Protein sequences were aligned and superposed on C$\alpha$ atoms to quantify fold similarity, while AlphaFold3 confidence scores were used to assess structural plausibility.For instance, the Yersinia phage vB\_YpM\_MLG42 shows a near-native match between generated and ground-truth structures, achieving a TM-score of 0.99 (Figure~\ref{fig:experiment_result}f).

Overall structural fidelity was low across the 1,143 paired targets. Only a small subset of sequence pairs exhibits strong fold-level concordance, with 22 pairs achieving TM-like $\ge 0.50$ and 44 pairs exhibiting C$\alpha$-RMSD $\leq$ 5~\AA{}. Generated proteins also exhibited lower AlphaFold3 confidence than their matched truths, and that the most consistent matches were enriched among shorter targets and phage-associated hosts, suggesting that current models more reliably preserve structural constraints for simpler viral proteins. For more analysis and structural comparison charts, please see the Appendix \ref{app:af3}.

\subsubsection{Application of Benchmarking Insights}
To evaluate whether our benchmarking findings can guide practical model development under resource constraints, we construct an in-house pre-training corpus, \textbf{ViroBland}, and use it to train lightweight viral nucleotide models.

\paragraph{\textbf{Pre-training corpus.}}
ViroBland is a 216M-nucleotide pre-training dataset designed to combine broad genomic context with virus-enriched in-domain information. It integrates three data sources:
\begin{itemize}
    \item \textbf{Human Reference:} Selected intervals from GRCh38 (Chr1--22 and ChrX), providing stable eukaryotic genomic context.
    \item \textbf{Multi-species Diversity:} A cross-species collection derived from the Nucleotide Transformer~\cite{nucleotidetransformer} dataset, covering bacteria, fungi, invertebrates, and vertebrates.
    \item \textbf{Viral In-domain Data:} A curated subset from the OpenVirus (LucaVirus-Gene) corpus, providing high-density viral nucleotide sequences.
\end{itemize}

To balance the three sources, we perform stratified sampling by selecting 12,000 training, 2,000 validation, and 2,000 test sequences from each source. After sequence-level deduplication, the final ViroBland corpus contains 32,023 training sequences, 4,000 validation sequences, and 2,662 test sequences. Source-specific statistics are summarized in Table~\ref{tab:virobland_source_bp}.

\begin{table}[htbp]
  \caption{Total base pairs by source in the ViroBland pre-training dataset.}
  \label{tab:virobland_source_bp}
  \centering
  \small
  \setlength{\tabcolsep}{6pt}
  \renewcommand{\arraystretch}{1.1}
  \begin{tabular*}{\columnwidth}{@{\extracolsep{\fill}}l r@{}}
    \toprule
    \textbf{Source} & \textbf{Total bases} \\
    \midrule
    Human reference genome (hg38/GRCh38) & 3.01 Mb \\
    Multi-species genomes (Nucleotide Transformer) & 163.84 Mb \\
    Viral sequences (OpenVirus) & 49.01 Mb \\
    \midrule
    \textbf{Total} & \textbf{216 Mb} \\
    \bottomrule
  \end{tabular*}
\end{table}

\paragraph{\textbf{Lightweight viral pretraining.}}
Using ViroBland, we developed ViroHyena, a series of lightweight Hyena-based models. The comprehensive pipeline, from stratified sampling to model training, is detailed in Appendix~\ref{app:ViroBland and ViroHyena}. By prioritizing virus-enriched and taxonomically diverse pre-training data over raw parameter scale, ViroHyena-436K improves the overall mean F1 on classification tasks to 39.32, corresponding to a 67.5\% gain over the original HyenaDNA-Large-1M (23.48). More detailed results are provided in Appendix~\ref{app:Pretrain Results}.

To examine whether this improvement is specific to the Hyena architecture, we further conduct architecture-level ablations by applying the same ViroBland pre-training strategy to DNABERT2 and Caduceus-PS, resulting in ViroDNABERT2 and ViroCaduceus. As shown in Appendix~\ref{app:architecture_ablation}, both ViroBland-adapted models improve over their corresponding original backbones across the evaluated classification settings. These results suggest that the benefit of ViroBland is not tied to a particular architecture, but reflects the broader value of virus-enriched and taxonomically diverse pre-training data.

Together, these preliminary results show that benchmark-derived insights can guide data-efficient viral model development, and that optimized data composition can partially compensate for limited parameter scale in the viral domain.

\section{Conclusions} 
In this work, we present \vb, the first comprehensive diagnostic benchmark for NFMs tailored to viral genomics, which evaluates \emph{biological understanding} and \emph{latent biosecurity risk} through 4 primary task types spanning 18 diverse scenarios. \vb~aims to address the critical gap in standardized evaluation for NFMs by instantiating diverse evaluation regimes, including genus-disjoint splits, temporal splits, and length-bucketed partitioning. We benchmark 66 NFMs, providing diagnostic insights into their performance across phylogenetic distances and evolutionary trajectories. Additionally, leveraging our finding that taxonomic diversity outweighs parameter scale, we establish a lightweight baseline that achieves a 67.5\% performance gain over significantly larger models. We further conduct a joint analysis (Appendix ~\ref{app:joint analysis}) to explore the synergy between classification and generation capabilities, and provide a Nipah-focused case study (Appendix ~\ref{app:case study}). By providing interpretable, diagnostic evaluations and a standardized, reproducible measurement framework, \vb is poised to accelerate research on viral nucleotide foundation models and to support viral genomic surveillance and responsible biosecurity governance.

\section*{Limitations and Ethical Considerations}
\vb~has several limitations.
Host prediction is framed as a single-label classification over coarse categories, prioritizing primary-host annotations for maximal reproducibility. Future iterations may expand to multi-label prediction as metadata matures. Additionally, while the temporal split uses NCBI deposit dates for uniformity, we acknowledge these reflect sequencing intensity rather than true emergence; future versions could integrate molecular-clock estimates for refined dating.

While \vb~is designed as an evaluation benchmark rather than a de novo virus design system, its generative tasks may reveal whether nucleotide-based models can produce sequences with plausible viral genomic statistics or protein-coding properties. Our goal is not to achieve actionable pathogen generation, but to make such risks measurable and visible; therefore, we do not provide guidance on constructing, rescuing, or validating generated viruses.

Ethics approval is not required as this study uses only publicly available viral sequences and non-identifying metadata.

\section*{GenAI Disclosure}
We used Generative AI tools to assist with manuscript preparation in language polishing. 
GenAI tools were not used to generate or manipulate experimental data, to perform statistical analyses, or to draw scientific conclusions. 
All AI-assisted text was reviewed, edited, and verified by the authors, who take full responsibility for the content.

\section*{Acknowledgments}
This work was partially supported by the New Generation Artificial Intelligence-National Science and Technology Major Project of China (2025ZD0121801) and the Prevention and Control of Emerging and Major Infectious Diseases-National Science and Technology Major Project of China (2025ZD01901102).
\bibliographystyle{ACM-Reference-Format}
\bibliography{1reference}

\newpage
\appendix

\label{sec:appendix}
\startcontents[appendix]
\section*{Appendix}
\printcontents[appendix]{l}{1}{\section*{Table of Contents}}
\newpage

\section{Detailed Data Curation Pipeline}\label{app:Detailed Data Preprocessing}
\subsection{Data Sources and Quality Filters}
The construction of the \vb~corpus followed a systematic pipeline designed to ensure both biological grounding and genomic integrity. We initiated the process by enumerating 273,974 virus-associated TaxIDs from the NCBI database. For each entry, complete taxonomic lineages—ranging from Kingdom to Species—were reconstructed via the NCBI Taxonomy hierarchy. To establish a reliable temporal dimension, we extracted the earliest discovery dates from NCBI viral data reports, adopting these first-seen timestamps as the canonical ``recorded time''. Following an initial quality-control phase that excluded entries with truncated taxonomic fields or missing temporal metadata, 204,603 TaxIDs remained. Specifically, we acquired complete whole-genome nucleotide sequences, Coding Sequences (CDS), and corresponding metadata from RefSeq~\cite{o2016reference} and GenBank~\cite{benson2012genbank}. By enforcing a strict requirement for valid assemblies with comprehensive annotation, the candidate pool was refined to 67,749 TaxIDs.All genomic data and associated metadata were programmatically retrieved from NCBI databases, with a final collection cutoff of January 10, 2026.

To address the challenge of one species-level TaxID mapping to multiple genomic versions, we implemented a hierarchical tie-breaking policy to select a single representative assembly. RefSeq assemblies were prioritized; in their absence, GenBank records were considered. Candidates were then ranked through a lexicographic sorting key based on: (1) RefSeq status (Reference > Representative > others), (2) assembly level (Complete Genome > Chromosome > Scaffold > Contig), (3) annotation completeness, (4) update recency, and (5) accession version. We augmented these entries with host annotations by aggregating metadata from NCBI reports, successfully deriving explicit species-level host labels for 61,410 virus species.

Given that the raw metadata yielded an unwieldy label space of 8,170 fine-grained host categories, we consolidated these into eight coarse-grained classes to facilitate stable modeling and evaluation. To automate this complex mapping, we leveraged the Qwen3-235B~\cite{yang2025qwen3} model under a specialized prompting framework, as detailed below:

\begin{tcolorbox}[promptbox,title=\textbf{Host Categorization Prompt}]
You are a researcher in bioinformatics and virology.

Given a “host/source” field (\texttt{host}), it may be a Latin scientific name, an English common name, a cell line, a tissue, or another type of description.
Please assign this \texttt{host} to \textbf{one} of the following categories (\textbf{output only the label letter}, with no explanation):

\begin{tabular}{@{}p{0.04\linewidth}p{0.92\linewidth}@{}}
\texttt{A} & Bacterial host (clinically, environmentally, or foodborne common bacteria; including species/strains/serotypes)\\
\texttt{B} & Fungi/oomycetes/plant pathogens (fungi, oomycetes, molds; including strains/isolates)\\
\texttt{C} & Plant host (crop or wild plants; including species/varieties/tissues)\\
\texttt{D} & Vertebrate host (mammals/birds/reptiles/amphibians/fish)\\
\texttt{E} & Arthropod vector / invertebrate host (insects/arachnids/crustaceans/mollusks/nematodes)\\
\texttt{F} & Other or uncertain (environment/food/sample descriptions, or unclear)\\
\end{tabular}

\medskip
\texttt{host: \{host\}}
\end{tcolorbox}

\begin{tcolorbox}[promptbox,title=\textbf{Vertebrate Host Subtype Prompt}]
You have already determined that the given \texttt{host} belongs to the \textbf{vertebrate host} category (D). 
Now further assign it to one of the following subtypes (\textbf{output only the number 1/2/3}, with no explanation):

\begin{tabular}{@{}p{0.04\linewidth}p{0.92\linewidth}@{}}
\texttt{1} & Humans and non-human primates (\emph{e.g.},~\textit{Homo sapiens}, human, apes, monkeys)\\
\texttt{2} & Livestock or companion animals (\emph{e.g.},~cattle, sheep, pigs, chickens, ducks, geese, cats, dogs, horses, camels)\\
\texttt{3} & Wild vertebrates (\emph{e.g.},~bats, rodents, carnivores, deer, marine mammals, and other non-domesticated vertebrates)\\
\end{tabular}

\medskip
\texttt{host: \{host\}}
\end{tcolorbox}

To assess the reliability of the LLM-assisted host categorization, we constructed a manually verified validation subset. Specifically, we randomly sampled 100 instances from each of the eight final host classes and manually examined their original host/source descriptions to establish gold-standard labels. We then evaluated the original Qwen3-235B annotations against this manually verified subset and further obtained independent annotations from two additional large language models, GLM-5 and Kimi-K2.5, using the same label definitions. As shown in Table~\ref{tab:host_label_stats}, Qwen3-235B achieved an overall accuracy of 96.25\%, while GLM-5 and Kimi-K2.5 achieved 94.25\% and 95.63\%, respectively. Most host classes exhibited near-perfect agreement across models, suggesting that the coarse-grained host labels are generally robust. The remaining errors were mainly concentrated in D2, D3, and F, which are intrinsically more ambiguous because the raw metadata may involve mixed animal-source descriptions, wild--domestic boundary cases, environmental samples, cell lines, or underspecified host annotations. These results indicate that label noise introduced by the LLM-assisted annotation pipeline is limited and largely localized to ambiguous host categories. Through this curation and validation process, the final \vb~corpus was established with 58,314 high-quality labeled viral species.
\begin{figure}[htbp]
  \centering
  \includegraphics[width=\columnwidth]{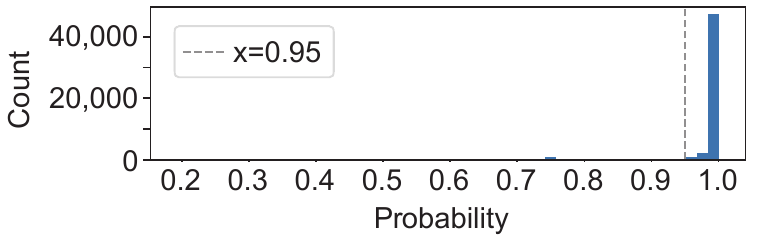}
  \caption{Distribution of the model’s maximum predicted class probability across samples. The dashed vertical line indicates the confidence threshold (0.95) used to filter out low-confidence predictions.}
  \label{fig:prob}
  \Description{Probability distribution plot for Qwen3-235B outputs.}
\end{figure}

\begin{table*}[htbp]
  \caption{Host label distribution and validation of LLM-assisted annotations.}
  \label{tab:host_label_stats}
  \centering
  \small
  \setlength{\tabcolsep}{4.5pt}
  \renewcommand{\arraystretch}{1.08}
  \begin{tabular}{@{}l r ccc l@{}}
    \toprule
    \textbf{Label} & \textbf{Count} 
    & \multicolumn{3}{c}{\textbf{Validation accuracy (\%)}} 
    & \textbf{Representative host examples} \\
    \cmidrule(lr){3-5}
    & & \textbf{Qwen} & \textbf{GLM} & \textbf{Kimi} & \\
    \midrule
    A  
      & 6{,}534  
      & 100 & \phantom{0}99 & \phantom{0}99 
      & Escherichia coli; Streptococcus pneumoniae; Bacillus subtilis \\

    B  
      & 820 
      & 100 & 100 & 100 
      & Ustilago maydis; Cryphonectria parasitica; Saccharomyces cerevisiae \\

    C  
      & 2{,}864 
      & 100 & 100 & 100 
      & Chlorella variabilis; Abutilon sellovianum; cassava \\

    D1 
      & 19{,}426 
      & 100 & \phantom{0}99 & 100 
      & Homo sapiens; Cercopithecus aethiops; Macaca silenus \\

    D2 
      & 12{,}744 
      & \phantom{0}99 & \phantom{0}87 & \phantom{0}84 
      & veal; sheep; Bos taurus \\

    D3 
      & 6{,}736 
      & \phantom{0}90 & \phantom{0}89 & \phantom{0}97 
      & raccoon; Columba livia; Rana pipiens \\

    E  
      & 1{,}411 
      & \phantom{0}96 & 100 & 100 
      & Choristoneura biennis; Orgyia pseudotsugata; Spodoptera exigua \\

    F  
      & 7{,}779 
      & \phantom{0}85 & \phantom{0}80 & \phantom{0}85 
      & Goutoucheng sour; cell lines; human gender \\
    \midrule
    Overall 
      & 58{,}314 
      & \phantom{0}96.25 & \phantom{0}94.25 & \phantom{0}95.63 
      & -- \\
    \bottomrule
  \end{tabular}

  \vspace{2pt}
  \footnotesize
  \emph{Note:}  
  A: bacterial host; 
  B: fungi/oomycetes/plant pathogens; 
  C: plant host; 
  D1: primates; 
  D2: livestock/companion animals; 
  D3: wild vertebrates; 
  E: arthropod/invertebrate host; 
  F: other or uncertain.
\end{table*}

\subsection{Data Partitioning}
Taxonomy and Host Classification are evaluated under two distinct splitting regimes. The specific partition logic, label cardinality, and temporal cutoff points are detailed in Table \ref{tab:dataset_stats_cls}.


The generation tasks are evaluated under three length regimes.We summarize the length regimes and sample counts for the generation tasks in Table~\ref{tab:dataset_stats_gen}.

Genome Modeling evaluation uses three length tiers based on global percentile thresholds of the entire collection: short ($P_{5} \le L \le P_{33}$), medium ($P_{33} < L \le P_{66}$), and long ($L > P_{66}$). For each TaxID, sequences are assigned to these buckets to stratify modeling difficulty while maintaining the dataset's overall length distribution. This enables a tiered assessment of the model's capacity for long-range dependencies.For CDS Generation, we curated a representative subset ($n=500$ per host category) using a two-stage sampling strategy to balance species diversity and recency. We first prioritized the most recent record for each unique species. If the budget $n$ was not met, we backfilled the remaining slots with the next most recent records.Following sampling, sequences were partitioned into short, medium, and long buckets using the same thresholds as in Genome Modeling. To further control redundancy, we applied evenly spaced subsampling ($k=3$) within each TaxID and length bucket.

\subsection{Recommended Lightweight Evaluation Subset}

We also provide a lightweight classification subset, ViroBench-CLS-Lite, for researchers working with limited computational resources. This subset supports rapid prototyping, hyperparameter screening, and preliminary model comparison, while retaining the main classification settings used in the full ViroBench benchmark.

ViroBench-CLS-Lite was constructed from the curated ViroBench corpus of 58,314 samples using time-balanced sampling. Each host class was sampled along the recorded-time axis. For each of the eight host categories, we set the target size to 1,000 samples, resulting in 8,000 records in total. Samples were assigned to the training, validation, and test periods using an approximate 8:1:1 ratio. Within each temporal window, records were selected at roughly even intervals according to their recorded time, which helped preserve broad temporal coverage and avoid overrepresenting densely sampled periods. For host classes with too few unique sequences, limited resampling was allowed to keep the class sizes consistent.

We then generated the same classification task settings as in the full benchmark. The subset was organized into ALL, DNA, and RNA settings, and included both taxonomy and host classification tasks. For each task and nucleic-acid setting, we provided two split types: a genus-based split to evaluate generalization across related taxonomic groups, and a temporal split to test whether models trained on earlier viral records can generalize to later ones. Before export, multi-contig genomes were aggregated by TaxID to produce sequence-level model inputs.

ViroBench-CLS-Lite is intended for model debugging, fast experimental iteration, hyperparameter screening, and preliminary comparisons among nucleotide foundation models. It is not meant to replace the full benchmark. Instead, it provides a standardized low-cost evaluation setting that can be used before running full-scale experiments. By keeping host-category sizes balanced and using consistent temporal boundaries across tasks, ViroBench-CLS-Lite offers a practical trade-off between computational efficiency and fidelity to the full benchmark.





\section{Implementation and Reproducibility}

\subsection{Model Specifications.}
To ensure the reproducibility of our benchmark results, we provide comprehensive specifications for all NFMs evaluated in \vb. Table~\ref{tab:appendix_model_stats} summarizes the architectural types, parameter scales, and specific versions utilized in our study.
\begin{table*}[!t]
  \centering
  \begin{threeparttable}
    \caption{Detailed specifications of NFMs evaluated in \vb.}
    \label{tab:appendix_model_stats}
    \footnotesize
    \setlength{\tabcolsep}{5pt}
    \renewcommand{\arraystretch}{1}

    \begin{tabularx}{\textwidth}{l c c c c c c >{\RaggedRight\arraybackslash}X}
      \toprule
      \textbf{Model Name} & \textbf{Lin.\tnote{*}} & \textbf{Cls.} & \textbf{Gen.} & \textbf{Max Params} & \textbf{Tokenizer} & \textbf{Model Type} & \textbf{Evaluated Sub-Models} \\
      \midrule

      AIDO.DNA\cite{ellington_accurate_2024} & N & \ding{51} & \ding{55} & 7B & Single & BERT & AIDO.DNA-300M/7B \\
      \addlinespace[0.1em]

      AIDO.RNA\cite{Zou2024.11.28.625345} & R & \ding{51} & \ding{55} & 1.6B & Single & BERT & AIDO.RNA-650M/1.6B, AIDO.RNA-650M/1.6B-CDS \\
      \addlinespace[0.1em]

      BiRNA-BERT\cite{Tahmid2024.07.02.601703} & R & \ding{51} & \ding{55} & 117M & BPE + Single & BERT & BiRNA-BERT \\
      \addlinespace[0.1em]

      Caduceus\cite{schiff2024caduceus} & N & \ding{51} & \ding{55} & 7.73M & Single & Bi-Mamba & Caduceus-ph-131k, Caduceus-ps-131k \\
      \addlinespace[0.1em]

      DNABERT\cite{ji2021dnabert} & N & \ding{51} & \ding{55} & 110M & Overlapping K-mer & BERT & DNABERT (3/4/5/6-mer) \\
      \addlinespace[0.1em]

      DNABERT-2\cite{zhou2023dnabert2} & N & \ding{51} & \ding{55} & 117M & BPE & BERT & DNABERT-2-117M \\
      \addlinespace[0.1em]

      DNABERT-S\cite{zhou2025dnabert} & D & \ding{51} & \ding{55} & - & BPE & BERT & DNABERT-S \\
      \addlinespace[0.1em]

      Evo1\cite{nguyen2024sequence} & P & \ding{51} & \ding{51} & 6.45B & Single & StripedHyena & evo-1-8k-base/131k-base \\
      \addlinespace[0.1em]

      Evo1.5\cite{merchant2025semantic} & P & \ding{51} & \ding{51} & 6.45B & Single & StripedHyena & evo-1.5-8k-base \\
      \addlinespace[0.1em]

      Evo2\cite{evo2} & P & \ding{51} & \ding{51} & 40B & Single & StripedHyena2 & evo2-1b-base/7b-base/40b-base, evo2-7b/40b \\
      \addlinespace[0.1em]

      GENA-LM\cite{GENA_LM} & N & \ding{51} & \ding{55} & 336M & BPE & BERT & gena-lm-bigbird-base-t2t, gena-lm-bert-base/large-t2t \\
      \addlinespace[0.1em]

      GENERator v2\cite{wu2025generator} & N & \ding{51} & \ding{51} & 3B & Non-overlapping K-mer & Transformer Decoder & \makecell[l]{GENERator-v2-eukaryote-1.2b/3b-base, \\ GENERator-v2-prokaryote-1.2b/3b-base} \\
      \addlinespace[0.1em]

      Genos\cite{10.1093/gigascience/giaf132} & N & \ding{51} & \ding{51} & 10B & Single & MoE Transformer & Genos-1.2B/10B/10B-v2 \\
      \addlinespace[0.1em]

      GenomeOcean\cite{zhou2025genomeocean} & D & \ding{51} & \ding{51} & 4B & BPE & Transformer Decoder & GenomeOcean-100M/500M/4B \\
      \addlinespace[0.1em]

      Grover\cite{sanabria2024dna} & N & \ding{51} & \ding{55} & - & BPE & BERT & Grover \\
      \addlinespace[0.1em]

      HyenaDNA\cite{nguyen2023hyenadna} & N & \ding{51} & \ding{51} & 54.6M & Single & Hyena & \makecell[l]{HyenaDNA-Tiny-1k/16k, HyenaDNA-Small-32k,\\ HyenaDNA-Medium-160k/450k, HyenaDNA-Large-1M} \\
      \addlinespace[0.1em]

      LucaOne\cite{he2025generalized} & D & \ding{51} & \ding{55} & 1.8B & Single & BERT & LucaOne-default-step36M, LucaOne-gene-step36.8M \\
      \addlinespace[0.1em]

      LucaVirus\cite{pan2025predicting} & D & \ding{51} & \ding{55} & 1.8B & Single & BERT & LucaVirus-default/gene-step3.8M \\
      \addlinespace[0.1em]

      MP-RNA\cite{yang2024mp} & N & \ding{51} & \ding{55} & 186M & Single & Transformer & MP-RNA \\
      \addlinespace[0.1em]

      NT v1\cite{nucleotidetransformer} & N & \ding{51} & \ding{55} & 2.5B & Non-overlapping K-mer & BERT & NT-500M-Human/1000G, NT-2.5B-1000G/MS \\
      \addlinespace[0.1em]

      NT v2\cite{nucleotidetransformer} & N & \ding{51} & \ding{55} & 500M & Non-overlapping K-mer & BERT & NTv2-50M-MS-3kmer, NTv2-50M/100M/250M/500M-MS \\
      \addlinespace[0.1em]

      NT v3\cite{boshar2025foundational} & N & \ding{51} & \ding{55} & 650M & Single & U-Net+Diffusion & NTv3-8M/100M/650M-pre, NTv3-100M/650M-post \\
      \addlinespace[0.1em]

      OmniReg-GPT\cite{wang2025omnireg} & N & \ding{51} & \ding{51} & 270M & BPE & GPT & omniReg-gpt-270M \\
      \addlinespace[0.1em]

      RNA-FM\cite{chen2022interpretable} & R & \ding{51} & \ding{55} & 99.52M & Single & BERT & RNA-FM \\
      \addlinespace[0.1em]

      RiNALMo\cite{penic2025rinalmo} & R & \ding{51} & \ding{55} & 650.88M & Single & BERT & RiNALMo \\
      \addlinespace[0.1em]

      RNABERT\cite{akiyama2022informative} & N & \ding{51} & \ding{55} & 0.48M & Single & BERT & RNABERT \\
      \bottomrule
    \end{tabularx}

    \begin{tablenotes}
      \item[*] \textbf{Pretraining Coverage Lineage}: (\textbf{D}) Diverse Viral Coverage; (\textbf{P}) Phage-specific Coverage; (\textbf{R}) RNA-specific Coverage; (\textbf{N}) Non-viral Coverage.
    \end{tablenotes}
  \end{threeparttable}
\end{table*}

\subsection{Standardized Evaluation Framework}

We evaluate all assessed foundation models using a frozen-backbone protocol to isolate and compare their representational quality. In this pipeline, the pretrained weights of the NFMs are kept fixed, and only a lightweight multi-task classification head is trained on the extracted sequence embeddings. To optimize computational efficiency, these embeddings are precomputed and cached for all data splits.

\paragraph{Embedding Extraction and Pooling.} To ensure representational integrity, pooling strategies strictly follow each model's original implementation (detailed in Table~\ref{tab:embedding_strategy_ordered_4col}). Specifically, most BERT-style models utilize mean pooling, while long-context architectures (\emph{e.g.}, Evo, HyenaDNA, Nucleotide Transformer) and the Evo2 family rely on the final token representation. DNABERT models utilize the [CLS] token, whereas our CNN baseline is trained end-to-end with global pooling. For sequences exceeding a model's native context limit, we employ a window-based approach: during training, we perform random sub-sampling of sequence windows; during validation and testing, we utilize fixed-count sampling or full-sequence coverage, aggregating window-level representations via mean pooling to produce final sequence-level predictions.

\paragraph{Unified Benchmarking Interfaces.} To ensure cross-model consistency, we implement two standardized interfaces: \begin{itemize} \item \emph{get\_embedding}: A unified wrapper that standardizes embedding extraction and caching across all discriminative tasks. \item \emph{generate}: A framework for autoregressive models to assess generative behaviors—such as K-mer spectrum deviation and CDS validity—under standardized prompt construction and length-control rules. \end{itemize} By maintaining this rigorous consistency, \vb enables a fair comparison across diverse architectures, ranging from classification accuracy to generative fidelity, within a fully reproducible pipeline. Specific hyperparameter configurations and the internal architecture of the MLP head are detailed in Appendix~\ref{app:Model Architecture Specifications} and Table~\ref{tab:train_config}.

\begin{table}[htbp]
  \centering
  \caption{Embedding extraction or pooling strategies for all evaluated models.}
  \label{tab:embedding_strategy_ordered_4col}
  \small
  \setlength{\tabcolsep}{9pt}
  \renewcommand{\arraystretch}{1.1}

  \resizebox{\linewidth}{!}{%
  \begin{tabular}{@{}lc|lc@{}}
    \toprule
    \textbf{Model} & \textbf{Strategy} & \textbf{Model} & \textbf{Strategy} \\
    \midrule
    Evo1 & Final & Evo1.5 & Final \\
    Evo2 & Final\textsuperscript{*} & NT V1 & Final \\
    NT V2 & Final & NT V3 & Mean \\
    Caduceus & Mean & DNABERT & CLS \\
    DNABERT-2 & Mean & DNABERT-S & Mean \\
    HyenaDNA & Final & Genos & Mean \\
    OmniReg-GPT & Mean & Gena-LM & Mean \\
    Grover & Mean & GenomeOcean & Mean \\
    GENERator V2 & Last Token & AIDO.DNA & Mean \\
    AIDO.RNA & Mean & LucaOne & Mean \\
    LucaVirus & Mean & RNA-FM & Mean \\
    RiNALMo & Mean & BiRNA-BERT & Mean \\
    RNABERT & Mean & MP-RNA & Mean \\
    \bottomrule
    \addlinespace[2pt]
    \multicolumn{4}{@{}l@{}}{\scriptsize \textsuperscript{*}\shortstack[l]{Evo2 uses layer name outputs: 1B$\rightarrow$24, 7B$\rightarrow$26, 40B$\rightarrow$20.}}
  \end{tabular}%
  }
\end{table}

\begin{table}[htbp]
  \centering
  \caption{Training configurations and optimization hyperparameters.}
  \label{tab:train_config}
  \small
  \setlength{\tabcolsep}{10pt}
  \renewcommand{\arraystretch}{1.1}
  \resizebox{\columnwidth}{!}{%
  \begin{tabular}{lc}
    \toprule
    \textbf{Hyperparameter} & \textbf{Value (Default)} \\
    \midrule
    Learning Rate             & $\{10^{-2}, 10^{-3}, 10^{-4}\}$ \\
    Weight Decay              & $0.01$ \\
    Maximum Epochs            & $300$ \\
    Head Batch Size           & $64$  \\
    Early Stopping Patience   & $30$ \\
    Early Stopping Metric     & Accuracy \\
    Minimum Improvement ($\Delta$) & $10^{-4}$ \\
    Class Weights             & Balanced \\
    Window Length             & $\{512, 1024, 2048\}$ \\
    Training Windows          & $\{8, 4, 2\}$ \\
    Evaluation Windows        & $\{64, 32, 16\}$ \\
    \bottomrule
  \end{tabular}%
  }
\end{table}

\subsection{Model Architecture Details}\label{app:Model Architecture Specifications}
\begin{figure*}[t]
  \centering
  \includegraphics[width=\textwidth]{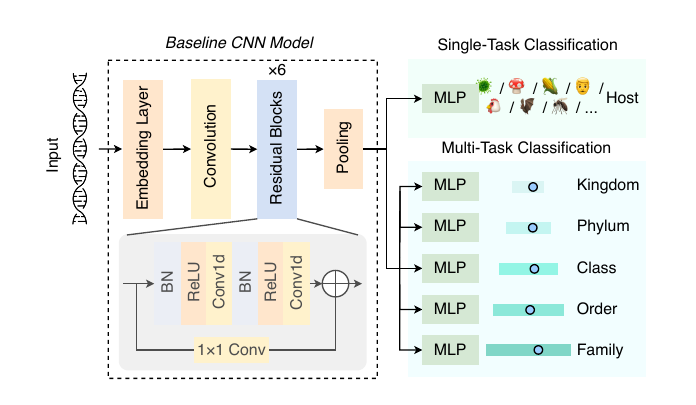}
  \caption{Detailed configuration of the baseline 1D-ResNet architecture. The schematic illustrates the end-to-end processing pipeline, from nucleotide embedding to task-specific outputs. The backbone comprises a 1D convolutional stem followed by six residual blocks ($N=6$), each employing a bottleneck-free BN--ReLU--Conv1d sequence. Skip connections incorporate $1\times1$ convolutions for dimensionality matching where necessary. The global average pooling layer compresses feature maps into a fixed-length embedding for the prediction heads. On the right, the dual-pathway head configuration is shown: a single MLP for host classification and five parallel MLP heads for hierarchical taxonomic prediction across five ranks (Kingdom, Phylum, Class, Order, Family).}
  \label{fig:CNN}
\end{figure*}

\paragraph{CNN Baseline}
We employ a 1D ResNet as our baseline architecture for genomic sequence analysis (see Fig.~\ref{fig:CNN} for the full architecture). The model first maps discrete nucleotides (A/C/G/T/N) into continuous vector representations. These are then processed through a convolutional stem and a series of residual blocks to progressively extract local motifs and higher-order compositional patterns, forming a hierarchical feature representation. To handle variable-length sequences, a global pooling layer aggregates these features into a fixed-dimensional embedding. This embedding is then fed into lightweight MLP heads for prediction. Our design supports both single-task and multi-task learning: multiple parallel heads can share the same backbone to jointly predict various taxonomic ranks (\emph{e.g.}, from \emph{kingdom} to \emph{Family}), facilitating parameter-efficient feature sharing and better generalization. Detailed configurations are provided in Table~\ref{tab:genomecnn1d_config}.
\begin{table}[htbp]
  \centering
  \caption{Hyperparameter specifications for the baseline CNN.}
  \label{tab:genomecnn1d_config}
  \small
  \setlength{\tabcolsep}{5pt}
  \renewcommand{\arraystretch}{1.1}
  \begin{tabular}{@{}lc | lc@{}}
    \toprule
    \textbf{Parameter} & \textbf{Value} & \textbf{Parameter} & \textbf{Value} \\
    \midrule
    Vocabulary Size     & 5                & Kernel Size         & 7 \\
    Padding Index       & 0                & Normalization       & BatchNorm1D \\
    Embedding Dimension & 64               & GN Groups           & 8 \\
    Hidden Channels     & (64, 128, 256)   & Dropout Rate        & 0.2 \\
    Blocks Per Stage    & (2, 2, 2)        & Global Pooling      & \texttt{Avg} \\
    Head Hidden Units   & 256              & Head Dropout        & 0.3 \\
    \bottomrule
  \end{tabular}
\end{table}

\paragraph{Classification Head}
To ensure a fair comparison, we attach a standardized classification head to all assessed NFMs, varying only the input sequence embeddings. Following the adaptation protocol of Evo2, we employ a lightweight Multi-Layer Perceptron (MLP) as the prediction head. This setup ensures that performance variations are primarily driven by the backbone's representational quality rather than differences in the head architecture.

\begin{table}[htbp]
  \centering
  \caption{Dynamic MLP classification head size as a function of label cardinality.}
  \label{tab:mlp_head_config}
  \small
  \setlength{\tabcolsep}{4pt}
  \renewcommand{\arraystretch}{1.1}
  \begin{tabular}{@{}p{0.45\columnwidth}p{0.55\columnwidth}@{}}
    \toprule
    \textbf{Label size $C$} & \textbf{Hidden widths $(h_1,h_2,h_3)$} \\
    \midrule
    $C<100$ & $(512,128,64)$ \\
    $100\le C<1000$ & $(512,256,128)$ \\
    \midrule
    \textbf{Output} & Logits (activation applied in the loss) \\
    \bottomrule
  \end{tabular}
\end{table}

The MLP head maps a $D$-dimensional input vector through three feed-forward blocks, each consisting of a linear transformation, ReLU activation, and Layer Normalization. Dropout ($p=0.3$) is applied after the first two blocks to mitigate overfitting. The final layer outputs raw logits, which are fed directly into the loss function (e.g., CrossEntropyLoss) without internal softmax or sigmoid activations. To balance capacity and parameter efficiency, we dynamically scale the hidden layer widths based on the label cardinality $C$ (Table~\ref{tab:mlp_head_config}); smaller label spaces utilize narrower layers to prevent overfitting. All weights are Kaiming-initialized to ensure training stability alongside ReLU activations.

\subsection{Evaluation Metrics}\label{subsec:Evaluation Metrics}
We employ the following formulations for performance evaluation.
\subsubsection{Evaluation Metrics for Taxonomy and Host Classification}
\paragraph{\textbf{Precision}} Precision is computed as:\begin{equation}P_i = \frac{TP_i}{TP_i + FP_i}\end{equation}where $TP_i$ and $FP_i$ denote the number of true positives and false positives for the $i$-th class, respectively. This metric quantifies the model's reliability in identifying specific viral families without introducing excessive false alarms.

\paragraph{\textbf{Recall}} Recall is computed as:\begin{equation}R_i = \frac{TP_i}{TP_i + FN_i}\end{equation}where $FN_i$ denotes the number of false negatives for the $i$-th class. Recall is particularly critical in the context of ViroBench to ensure that divergent or novel viral sequences are not overlooked by the model.
\paragraph{\textbf{Macro-F1 Score}}To ensure balanced evaluation across imbalanced viral categories, we report the Macro-average F1 score, which treats all classes with equal weight regardless of their sample size:\begin{equation}\text{Macro-F1} = \frac{1}{C} \sum_{i=1}^{C} \frac{2 \cdot P_i \cdot R_i}{P_i + R_i}\end{equation}
where $C$ denotes the total number of taxonomic or host categories, and $P_i$ and $R_i$ represent the precision and recall for the $i$-th class, respectively.

\paragraph{\textbf{Area Under the Precision-Recall Curve (AUPRC)}}AUPRC is computed as:\begin{equation}\text{AUPRC} = \sum_{n} (R_n - R_{n-1}) P_n\end{equation}
where $P_n$ and $R_n$ denote precision and recall at the $n$-th threshold, respectively. This metric summarizes the precision--recall trade-off across all classification thresholds.

\subsubsection{Evaluation Metrics for Genome Modeling}
\paragraph{\textbf{Bits Per Base (BPB)}}For Genome Modeling, BPB serves as the primary metric to quantify sequence likelihood across different tokenization schemes:\begin{equation}\mathrm{BPB} = \frac{\overline{\mathcal{L}}_{\text{tok}} \cdot T}{L \cdot \ln 2}\end{equation}
where $\overline{\mathcal{L}}_{\text{tok}}$ is the average token-level negative log-likelihood (in nats), $T$ is the total token count, and $L$ is the sequence length in bases. A lower BPB indicates superior modeling of genomic dependencies.

\subsubsection{Evaluation Metrics for CDS Generation}
\paragraph{\textbf{Edit Distance (ED)}}
To quantify the error rate in sequence reconstruction, we employ the Levenshtein distance $\mathrm{LD}(y, \hat{y})$, defined as the minimum number of single-nucleotide operations—specifically insertions, deletions, and substitutions—required to transform the generated sequence $\hat{y}$ into the target $y$. In the context of viral genomes, this metric accounts for potential frameshifts or point mutations during generation. To ensure comparability across sequences of varying lengths, we normalize this distance by the ground-truth length $|y|$:\begin{equation}\mathrm{ED}(y, \hat{y}) = \frac{\mathrm{LD}(y, \hat{y})}{|y|}.\end{equation}Under this formulation, an ED of $0$ indicates a perfect verbatim reconstruction, while higher values reflect increasing divergence from the reference. Note that ED can exceed $1.0$ if the model generates excessively long sequences compared to the target.

\paragraph{\textbf{Exact Match Accuracy (EMA)}}
EMA measures the character-level identity between the ground-truth continuation $y$ and the generated sequence $\hat{y}$:
\begin{equation}
\mathrm{EMA}(y,\hat{y}) = \frac{1}{|y|}\sum_{i=1}^{|y|}\mathbb{I}!\left[\hat{y}_i = y_i\right]
\end{equation}
This metric captures the model's ability to recover the precise nucleotide composition of the original viral sequence.

\paragraph{\textbf{CDS Success Rate (CSR)}}CSR evaluates the biological functional integrity of the generated sequences. A continuation is considered successful if the concatenated sequence $x_{1:p} \Vert \hat{y}$ maintains coding validity (e.g., proper reading frame and absence of internal stop codons):\begin{equation}\mathrm{CSR} = \frac{1}{|\mathcal{D}|}\sum_{(x,y)\in\mathcal{D}} \mathbb{I}{\mathrm{CDS}}!\left(x{1:p}\Vert \hat{y}\right)\end{equation}where $\mathbb{I}_{\mathrm{CDS}}(\cdot)$ is an indicator for CDS validity.

\paragraph{\textbf{K-mer Jensen--Shannon Divergence (kmer-JSD)}}Beyond surface-level alignment, we use the $k$-mer spectrum to assess the distributional plausibility of the generated sequence. The $k$ value is adaptively determined as $k = \mathrm{clamp}(\mathrm{round}(0.7\log_{4} L_{\mathrm{eff}}), 1, 13)$. Let $p$ and $q$ be the $k$-mer frequency distributions of $y$ and $\hat{y}$, respectively. With $m=\tfrac{1}{2}(p+q)$, the JSD is computed as:\begin{equation}\mathrm{kmer\text{-}JSD}(p,q) = \frac{1}{2}\sum_i p_i \log_2\frac{p_i}{m_i} + \frac{1}{2}\sum_i q_i \log_2\frac{q_i}{m_i}\end{equation}Lower JSD values indicate that the generated sequence better mimics the higher-order dependency patterns of real viral genomes.

\paragraph{\textbf{K-mer Kolmogorov--Smirnov Statistic (kmer-KS)}}To further quantify the distance between $k$-mer distributions, we employ the KS statistic on the cumulative distribution functions (CDFs) of the spectrum, $P(t)$ and $Q(t)$:\begin{equation}\mathrm{kmer\text{-}KS}(p,q) = \sup_{t}, |P(t)-Q(t)|\end{equation}The kmer-KS statistic measures the maximum deviation between the observed and generated $k$-mer counts, providing a robust assessment of biological plausibility.

\section{Additional Results}
\subsection{Classification Results}
\subsubsection{Detailed Performance Benchmarking and Extended Metrics}\label{app:Comprehensive Performance Metrics}
We provide the complete evaluation results for all benchmarked models across viral taxonomy and host classification tasks. The following tables present the exhaustive performance metrics:

\begin{itemize} 
\item \textbf{Precision (Table~\ref{tab:precision_results})}: Detailed precision scores for both taxonomy and host classification tasks across all evaluated models.
\item \textbf{Recall (Table~\ref{tab:recall_results})}: Detailed recall scores for both taxonomy and host classification tasks across all evaluated models.
\item \textbf{F1-Score (Table~\ref{tab:virus_cls_results_cont})}: Detailed macro-F1 scores for both taxonomy and host classification tasks across all evaluated models. 
\item \textbf{ALL-Taxon Macro-F1 (Table~\ref{tab:taxon_rank_results})}: Detailed macro-F1 scores across taxonomic ranks under G-split and T-split.
\end{itemize}

\subsubsection{Additional phylogenetic analyses}\label{app:phylo_trees}
\begin{figure*}[htbp]
  \centering
  \includegraphics[width=0.8\textwidth]{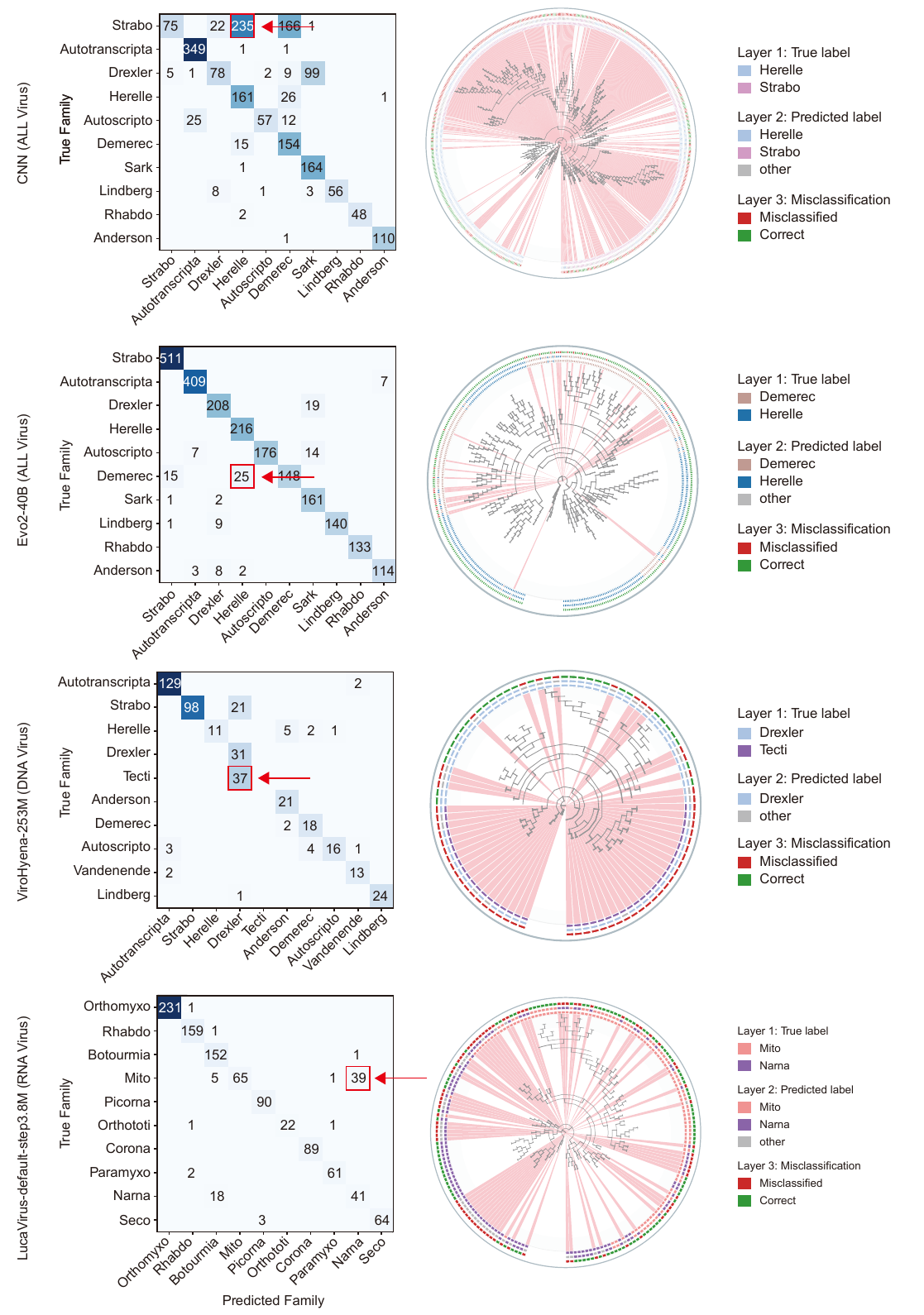}
  \caption{Performance evaluation and phylogenetic analysis of viral family classification across four models. The left column displays confusion matrices for each model, where the diagonal represents correct assignments and red boxes indicate the most frequent misclassification pair (True Label vs. Predicted Label). The right column features circular phylogenetic trees constructed from the sequences within these specific misclassified pairs. These trees visualize the relationship between biological ancestry and model performance across three layers: Layer 1 (True Label), Layer 2 (Predicted Label), and Layer 3 (Classification Result: Green for correct, Red for misclassified).}
  \label{fig:tree-Appendix}
\end{figure*}

We performed an extended analysis of misclassified sequences across all four models in Figure~\ref{fig:tree-Appendix}. Our results reveal that classification errors are not randomly distributed across the taxonomic landscape but are instead concentrated within specific ``conflict nodes''. For instance, the CNN model exhibits a pronounced collapse at the Strabo-Herelle interface, while Evo2-40B, ViroHyena, and LucaVirus encounter similar bottlenecks when distinguishing between Demerec/Herelle, Tecti/Drexler, and Mito/Narna clusters. Crucially, as evidenced by the phylogenetic trees, misclassified sequences  consistently form dense clusters at the boundaries of closely related lineages or nest deeply within the clades of the predicted family, rather than appearing as isolated outliers.

This systematic clustering confirms that model failures are fundamentally rooted in the evolutionary continuity of viral genomes. These "evolutionary gray zones" represent regions where genomic divergence has not yet produced the discrete sequence signatures required for models to establish stable latent boundaries. Rather than reflecting arbitrary algorithmic artifacts, these non-random biases suggest that current models are capturing genuine biological signals—such as ancestral motifs or convergent evolutionary traits—that confound standard taxonomic assignment. These findings imply that simply scaling model parameters is insufficient to resolve such deep-seated ambiguities; instead, future iterations must integrate phylogenetic topology directly into the training objective to navigate the fine-grained distinctions of viral evolution.

\subsubsection{Detailed Performance Deltas across Diagnostic Scenarios}\label{app:hotmap}
\begin{figure*}[htbp]
  \centering
  \includegraphics[width=\textwidth]{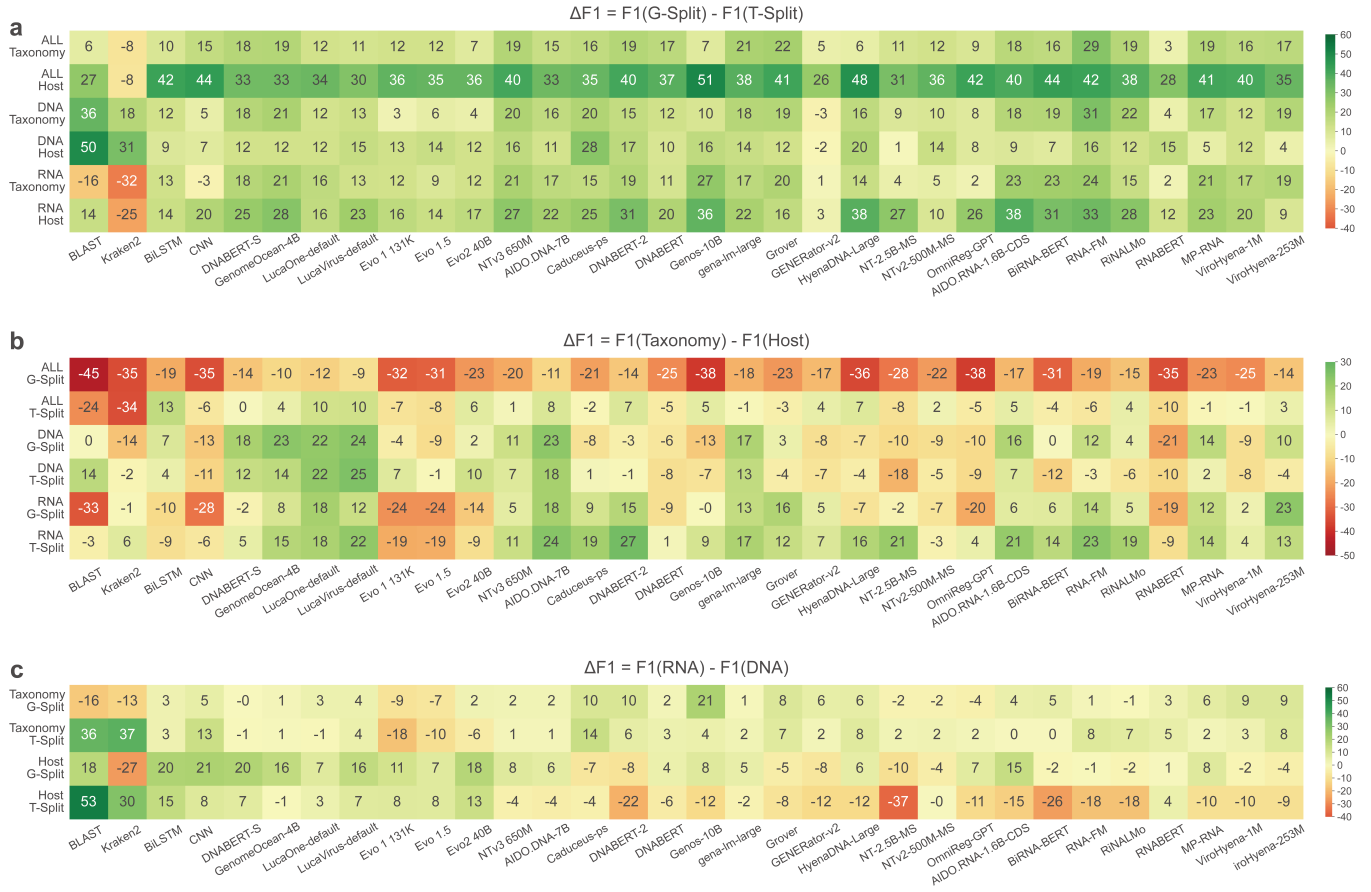}
  \caption{Multi-dimensional diagnostic analysis of performance disparities ($\Delta$F1). The heatmaps visualize the performance gaps across 12 diagnostic scenarios for various NFMs.
(a) Generalization gap ($\Delta$F1 = F1(G-Split) - F1(T-Split)), quantifying the performance decay when transitioning from temporal to genus-disjoint evaluations across taxonomy and host classification tasks.
(b) Task-wise disparity ($\Delta$F1 = F1(Taxonomy) - F1(Host)), illustrating the relative difficulty of host classification compared to taxonomy under different data splits.
(c) Modality bias ($\Delta$F1 = F1(RNA) - F1(DNA)), highlighting the performance differences between RNA and DNA sequence processing across tasks and splits.
Numerical values indicate the percentage point difference in F1 score.}
  \label{fig:appendix_heatmap}
\end{figure*}

We provide a comprehensive breakdown of the performance disparities (measured by $\Delta$F1) across 12 distinct diagnostic variants, encompassing different data splits, classification tasks, and sequence modalities (Figure~\ref{fig:appendix_heatmap}).
\paragraph{Performance Decay across Evaluation Splits (Figure~\ref{fig:appendix_heatmap}a)}A consistent "generalization tax" is observed when transitioning from the Temporal split (T-Split) to the Genus-disjoint split (G-Split). This performance decay is universal, appearing in the ALL dataset and across both DNA and RNA subsets. In taxonomy classification, the DNA subset consistently exhibits a more pronounced decay than the RNA subset across most models. In host classification, the performance drop is severe across all modalities, with multiple models (e.g., AIDO.DNA-7B, Genos-10B) showing $\Delta$F1 decreases exceeding 40. These results indicate that current models rely heavily on temporal proximity for host prediction, a dependency that persists regardless of sequence type.\paragraph{Task-Wise Asymmetry under Distribution Shifts (Figure~\ref{fig:appendix_heatmap}b)}The analysis reveals a significant disparity between taxonomy and host prediction performance. Under the T-Split, the performance gap ($\Delta$F1 = F1(Taxonomy) - F1(Host)) remains narrow across ALL, DNA, and RNA categories. However, under the G-Split, host prediction performance decreases significantly more than taxonomy performance. Notably, the RNA subset frequently maintains a smaller task gap in G-Split compared to the DNA subset. Conversely, for specific DNA models such as NT-2.5B-MS, the gap widens to -32 in G-Split, highlighting a systemic difficulty in maintaining host-specificity when evaluated on novel genera.\paragraph{Modality Bias and Model Heterogeneity (Figure~\ref{fig:appendix_heatmap}c)}The cross-modality comparison ($\Delta$F1 = F1(RNA) - F1(DNA)) demonstrates significant architectural divergence. In taxonomy tasks, the majority of models exhibit higher performance on RNA sequences. This trend is not uniform across tasks; in host classification under T-Split, certain models (e.g., HyenaDNA-Large and NT-2.5B-MS) show a substantial performance bias toward DNA ($\Delta$F1 reaching -37), while others like AIDO.DNA-7B show a more balanced profile. This heterogeneity confirms that modality preference is not solely determined by data distribution but is also influenced by specific model architectures and their capacity to capture genomic dependencies.
\subsubsection{Characterizing the Intrinsic Structure of Model Embeddings}
\begin{figure*}[htbp]
  \centering
  \includegraphics[width=\textwidth]{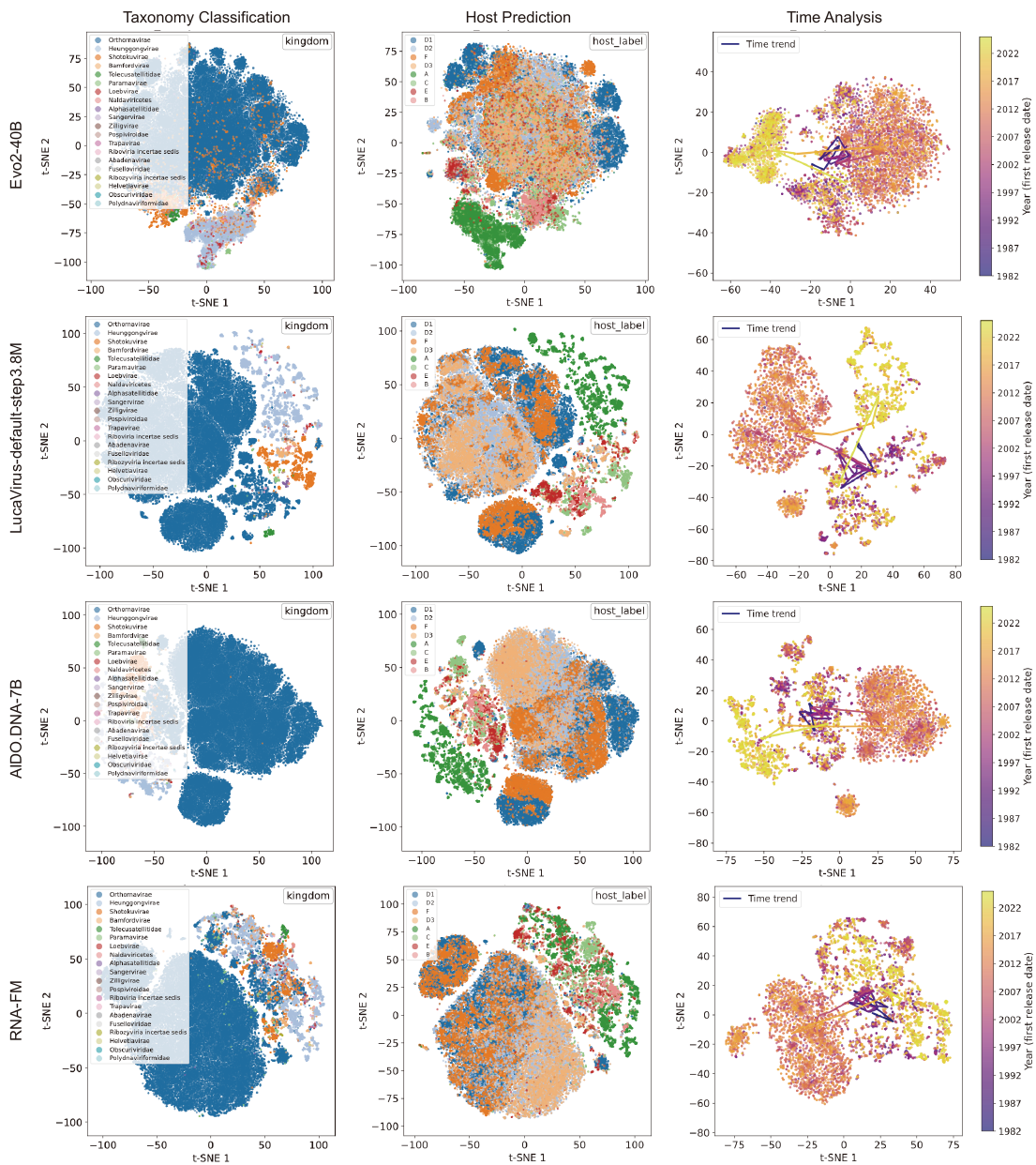}
  \caption{The t-SNE dimensionality reduction distributions of viral genome embedding vectors generated by four models (Evo2-40B, LucaVirus-default-step3.8M, AIDO.DNA-7B, and RNA-FM) in classification tasks and temporal evolution contexts.}
  \label{fig:embedding}
\end{figure*}

To investigate the representational capacity of the evaluated NFMs, we projected the high-dimensional embeddings generated by Evo2-40B, LucaVirus, AIDO.DNA, and RNA-FM into a two-dimensional space using t-SNE (Figure~\ref{fig:embedding}). This visualization demonstrates that the models' latent spaces possess inherent discriminative power, even without explicit fine-tuning for downstream tasks.

As shown in the Taxonomy and Host columns, sequences belonging to the same viral kingdom or host category form distinct, well-separated clusters. This structural organization suggests that the models capture fundamental genomic signatures such as codon usage bias or conserved functional motifs. The degree of clustering varies across architectures, reflecting differences in how models prioritize genomic features.

Furthermore, we performed a temporal embedding analysis to track the evolution of viral sequences over four decades (1982–2024). By coloring the embeddings according to their first release date and plotting the centroid trajectory (Time trend), we observe a clear "temporal drift" in the latent space. This shift indicates that the genomic features extracted by the models are sensitive to the progressive mutations and evolutionary adaptations of viruses over time. The non-overlapping distribution of early and contemporary virus sequences reinforces the idea that model embeddings can serve as a molecular clock of sorts, capturing the trajectory of viral divergence in a low-dimensional representation. 

\subsection{Generation Results}
\subsubsection{Comprehensive Performance Metrics.}
We report the full suite of generation metrics for all models and length buckets to ensure transparency and reproducibility. Table~\ref{tab:bpb_genome_bucket_stats} summarizes the BPB statistics for genome modelling across length buckets (lower is better), and Table~\ref{tab:cds_gen_results_detail} reports the CDS continuation results, covering sequence-level fidelity (edit distance and exact-match accuracy), distributional similarity (K-mer JSD and K-mer KS), and biological validity (CDS success rate).

\subsubsection{Generate Error Analysis}
This section analyzes the error patterns generated by CDS, identifying high-impact failure types that directly disrupt sequences that can be interpreted as reasonable encoded sequences, and providing specific cases to pinpoint the problems. The analysis focuses on truth-gen sequence pairs labeled CDS in the dataset, primarily examining the presence of illegal characters, disruption of codon structure, premature termination due to stop codons, and severe length anomalies.

From the overall error rate perspective, the generated sequences maintain formal encoding integrity in most samples. First, at the level of missing and empty sequences, no missing or empty strings were observed in the generated sequences, indicating that the generation process for this dataset is generally stable at the I/O level. Second, at the level of codon structure, after pruning the sequences to multiples of 3 and removing stop codons, if present, at the ends, no instances of the generated sequences failing to align with codon boundaries were observed, meaning that explicit bitshifting length errors are not the primary issue. Third, at the stop codon level, the dominant failures are associated with violations of CDS termination and internal coding consistency rather than simple format corruption. Under the CDS-Short setting, only 0.98\% of generated continuations (35/3,575) satisfy the CDS validity criteria. Canonical terminal stop codons appear at the expected terminal position in only 5.01\% of sequences, whereas premature internal stop codons occur in 76.84\% of generations. Moreover, 72.81\% of generated sequences simultaneously miss the expected terminal stop codon and contain at least one premature internal stop codon. These results indicate that the main failure mode is not invalid output formatting, but the inability to consistently preserve coding-frame and termination constraints during autoregressive decoding.

Although such errors are not the primary cause of failures, we still examine typical cases of invalid character output and severe length anomalies because they directly impact downstream parsing, translation, and ORF verification. Regarding invalid characters, only one generated sequence was found to contain the character N, which is not A/C/G/T, corresponding to taxid 3128054, model OmniReg-base, belonging to short bucketing. The sequence length is consistent with the reference (198 nt), and the end still uses a standard stop codon, indicating that the error did not originate from a failure in length control or the termination mechanism, but rather from a violation of alphabetical constraints. Further localization revealed that N appears at the 103rd base position of the generated sequence (counting from 0 to 102), with the local context "TGGGGACAAAAAAAAAAAATNCATCTCTGAAGGGCTGGGT". In the generative model output, this symbol may be an anomalous product of the decoding or post-processing stages. This type of error has a direct and severe impact on downstream processes because any analysis relying on explicitly defined codons, such as translation, ORF verification, and codon substitution statistics, will fail or produce indeterminate results at this position. Therefore, although this error is extremely rare overall, it should be given high engineering priority and typically needs to be eliminated through strict character constraints, post-output filtering, and triggered regeneration.

As a complementary case analysis beyond the CDS-Short causal study, we further inspect severe length anomalies in longer CDS generations. Four generated sequences were found to be inconsistent in length with the reference CDS, all of which were significantly shorter than the reference sequence. All four samples are from long buckets, and the corresponding models are concentrated in the original kernel version of GENERator. Among them, GENERator-v2-prokaryote-3b-base accounts for 3 cases, and GENERator-v2-prokaryote-1.2b-base accounts for 1 case. Specifically, the reference CDS length for taxid 2793733 is 6102 nt, while the generated length is only 204 nt; the reference length for taxid 2735919 is 4845 nt, while the generated length is only 138 nt; the reference length for taxid 2917257 is 6129 nt, while the generated length is only 180 nt; and the reference length for taxid 2810802 is 1551 nt, while the generated length is only 132 nt. Using the ratio of the generated length to the reference length as a visual characterization of truncation strength, we can see that the ratio ranges from only 2.85\% to 8.51\%, indicating that these outputs are closer to generating only a very short prefix of the reference CDS rather than a slight deviation. Further examination of the terminal codons of these truncated sequences reveals that they still end with stop codons (e.g., TAA or TAG) and there are no internal stop codons. This means that this failure mode is not a nonsense error caused by premature internal termination, but rather that termination occurs at the end of the sequence but too early, resulting in a formally translatable but much shorter ORF than the reference. In conditional generation tasks targeting the reference CDS, this type of output should be considered a generation failure because it fails to cover the main region of the target CDS and renders any comparison metrics based on full-length consistency incomparable.

It is noteworthy that although the proportion of length truncation in the overall sample is not high (4/1095, approximately 0.37\%), it accounts for 4/47 (approximately 8.51\%) within the long bucket, exhibiting a clear bucket concentration. This phenomenon suggests that length control failures may be related to scenarios with longer target sequences: when the model needs to maintain the coding structure and continue generation over a long span, it is more likely to provide a termination signal or trigger premature stopping at an earlier position. Considering that the prompt sequence length is 129 nt and the generated length of the truncated samples is close to the prompt length (e.g., 132 nt, 138 nt, 180 nt, 204 nt), it is reasonable to infer that this type of failure is related to the behavior of quickly generating termination and ending the output during the decoding process. Since these samples still end with the standard stop codon, this behavior is not random truncation, but more like the model quickly closing a short ORF after the prompt. For long CDS generation tasks, this may reflect the increased uncertainty of the model when extending generation, leading to a greater tendency to output a stop codon to complete a self-consistent but too short coding segment.

To further explore the causal mechanism behind the failure of CDS generation, we conducted gradient-based attribution analysis at the terminal decision point, which is defined as the decoding step before the generation of the final three nucleotides. We first measure the total probability mass assigned to the three canonical stop codons, TAA, TAG, and TGA, at this step. The stop-codon probability is nearly identical between successful and failed cases (0.0481 vs. 0.0482), suggesting that the failures are not simply caused by a local inability to emit stop codons.

We then quantify how strongly the terminal decision depends on the original input prompt, rather than only on the recently generated suffix. Successful cases exhibit substantially stronger prompt dependence than failed cases (0.4766 vs. 0.3590). This difference is also observed in the last quarter of the prompt, which is closest to the generation boundary, where successful cases assign higher attribution than failed cases (0.1472 vs. 0.1037).

Taken together, these results indicate that the primary causal mechanism is the insufficient preservation of long-range, immediate conditional encoding and termination constraints during the decoding process. Although the decoder can locally assign similar probability to stop codons in both successful and failed cases, failed generations are less conditioned on the original CDS context when making terminal decisions. As generation progresses, local decoding errors accumulate, weakening the global encoding framework and termination plan, leading to premature internal termination or missing terminal termination. This explains why likelihood levels or local sequence statistics alone are insufficient to assess CDS generation quality and further underscores the necessity of employing biologically based metrics such as CDS success rate and structural-level verification.

\subsubsection{Time trend analysis.}\label{app:time_gen}

\begin{figure*}[htbp]
  \centering
  \includegraphics[width=\textwidth]{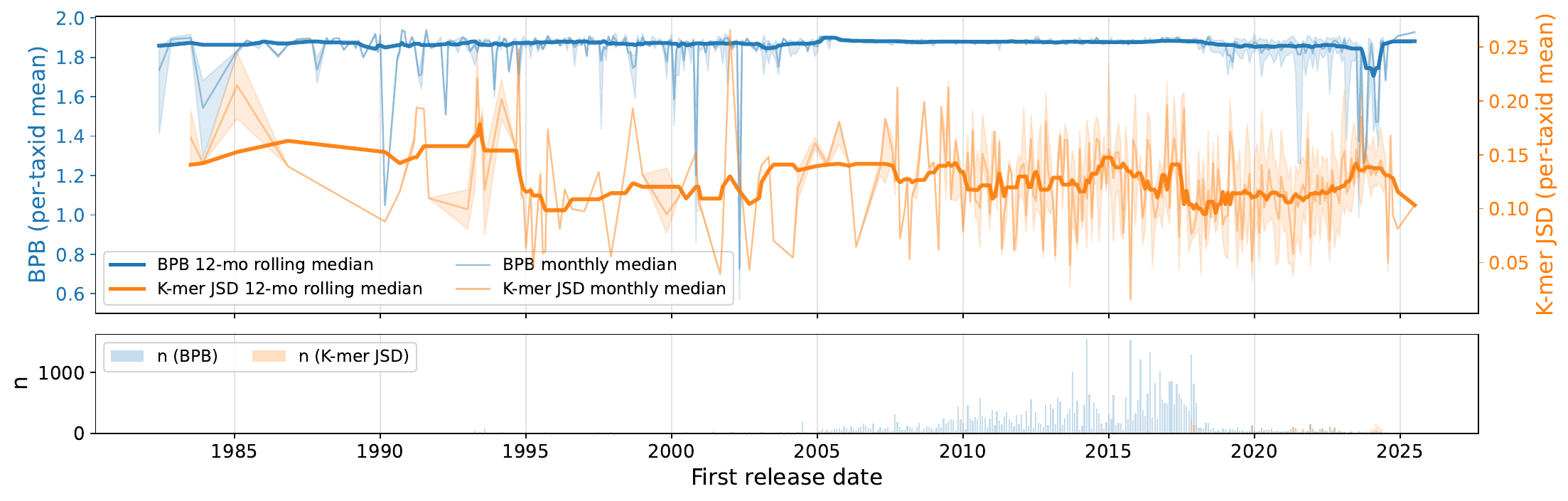}
  \caption{The monthly aggregate time trend of Evo2-40B. The top chart shows the BPB (left axis, blue) and K-mer JSD (right axis, orange), as well as the monthly median (thin line), 12-mo (month) rolling median (thick line), and interquartile range (IQR). The bottom chart shows the monthly sample counts for these two tasks.}
  \label{fig:time}
  \Description{Monthly median and 12-month rolling median trends of BPB and CDS K-mer JSD for Evo2-40B, with corresponding monthly sample sizes.}
\end{figure*}

To avoid interpreting temporal patterns as evidence that the data become intrinsically harder or easier over time, we make conservative statements only within year windows where sample sizes are relatively sufficient, and we explicitly emphasize the uncertainty in the most recent period. As shown in Fig.~\ref{fig:time}, using Evo2-40B as a representative example, both BPB and CDS $K$-mer JSD exhibit a gradual downward trend in the year-aggregated median series (BPB slope $\approx -1.80\times10^{-3}$ per year; JSD slope $\approx -1.27\times10^{-3}$ per year), suggesting that, on long time scales, the average likelihood behavior and compositional consistency with newly added sequences have not systematically deteriorated. Restricting the analysis to years with at least 50 samples, the median-of-yearly-medians for BPB is approximately 1.881 for 2005--2010 and 1.879 for 2011--2017, and decreases to about 1.862 for 2018--2022 (an improvement on the order of $\sim$0.02). Over the same stable-sample regime, JSD decreases from about 0.139 in 2011--2017 to about 0.096 in 2018--2022, indicating a more pronounced improvement in compositional agreement during this interval. 

In contrast, the 2023--2025 window in Fig.~\ref{fig:time} is characterized by fewer available years and larger sample-size fluctuations, such that the corresponding statistics may be dominated by a small number of months or by concentrated deposition from particular lineages; JSD may also show rebounds or increased volatility. Therefore, stronger causal attributions in time, for instance to sequencing technologies, annotation practices, or the continual emergence of new lineages, should be supported by stratified controls over host and lineage composition, rather than inferred from aggregate temporal curves alone. Within our evaluation framework, the most robust conclusion is that BPB and JSD do not show systematic degradation in periods with stable sample sizes, whereas the increased variability in recent years motivates finer-grained stratified analyses to identify the drivers of the observed fluctuations.

\subsubsection{Alphafold3 Structure Verification}\label{app:af3}
We evaluated whether models generating coding sequence continuations from fixed 129 nt starter sequences preserved protein-level constraints. For each sequence pair, we first performed a global protein sequence alignment to determine residue correspondences; all subsequent structural similarity statistics were calculated based on this alignment to avoid spurious error inflation caused by shifts, insertions, deletions, or length mismatches.

\begin{figure*}[htbp]
  \centering
  \includegraphics[width=0.9\textwidth]{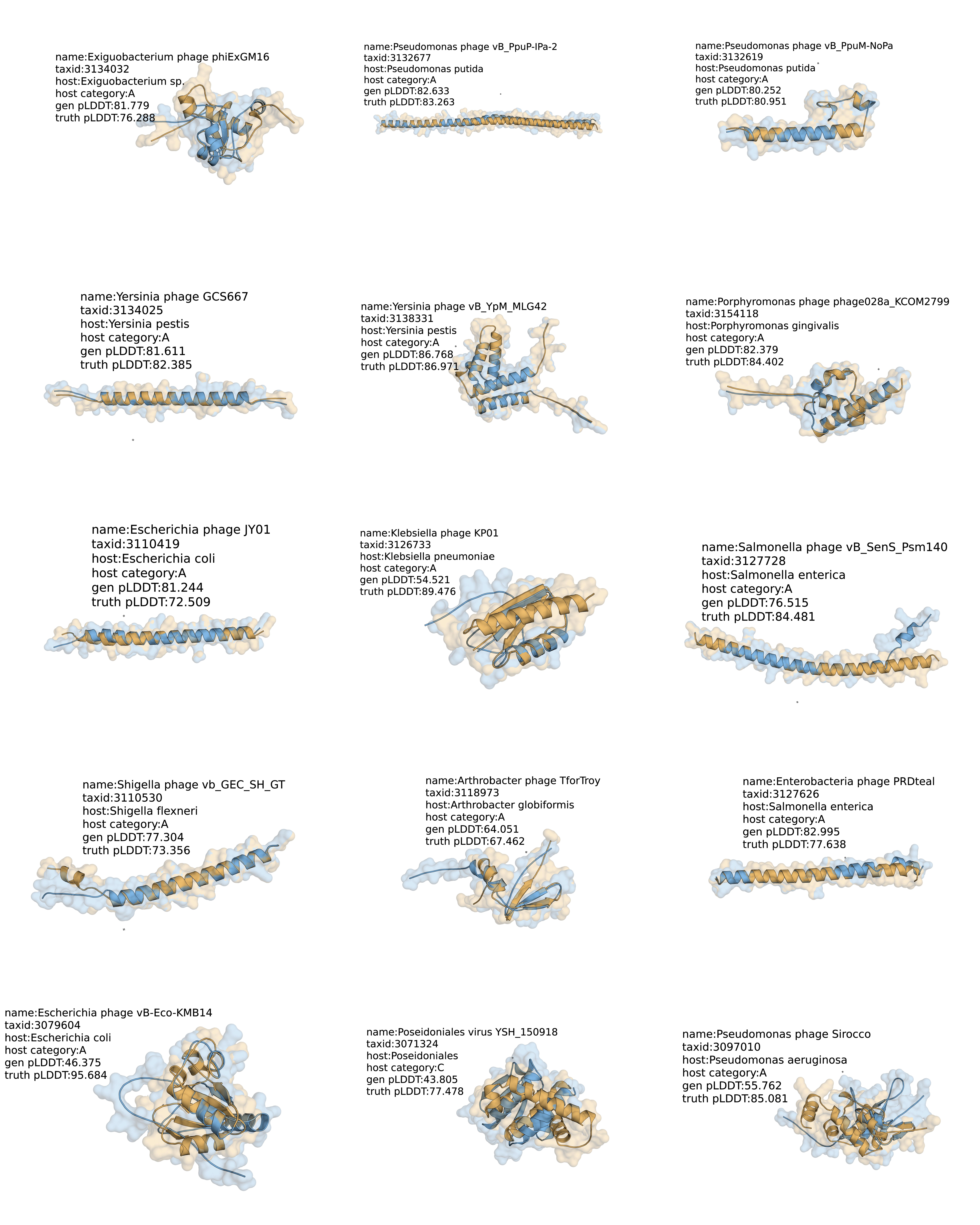}
  \caption{The generated and ground sequence structures of the top three targets in each model group are overlaid. Each panel overlays the AlphaFold3 predicted structures of the aligned ground sequence (orange, truth) and generated sequence (blue, gen), and annotates the corresponding taxid/host metadata and average pLDDT value. The models from top to bottom are: Evo1, Evo2, HyenaDNA-large-1M, Genos-10B, and GENERator-v2-3B.}
  \label{fig:af3_all}
\end{figure*}

\begin{figure*}[htbp]
  \centering
  \includegraphics[width=\textwidth]{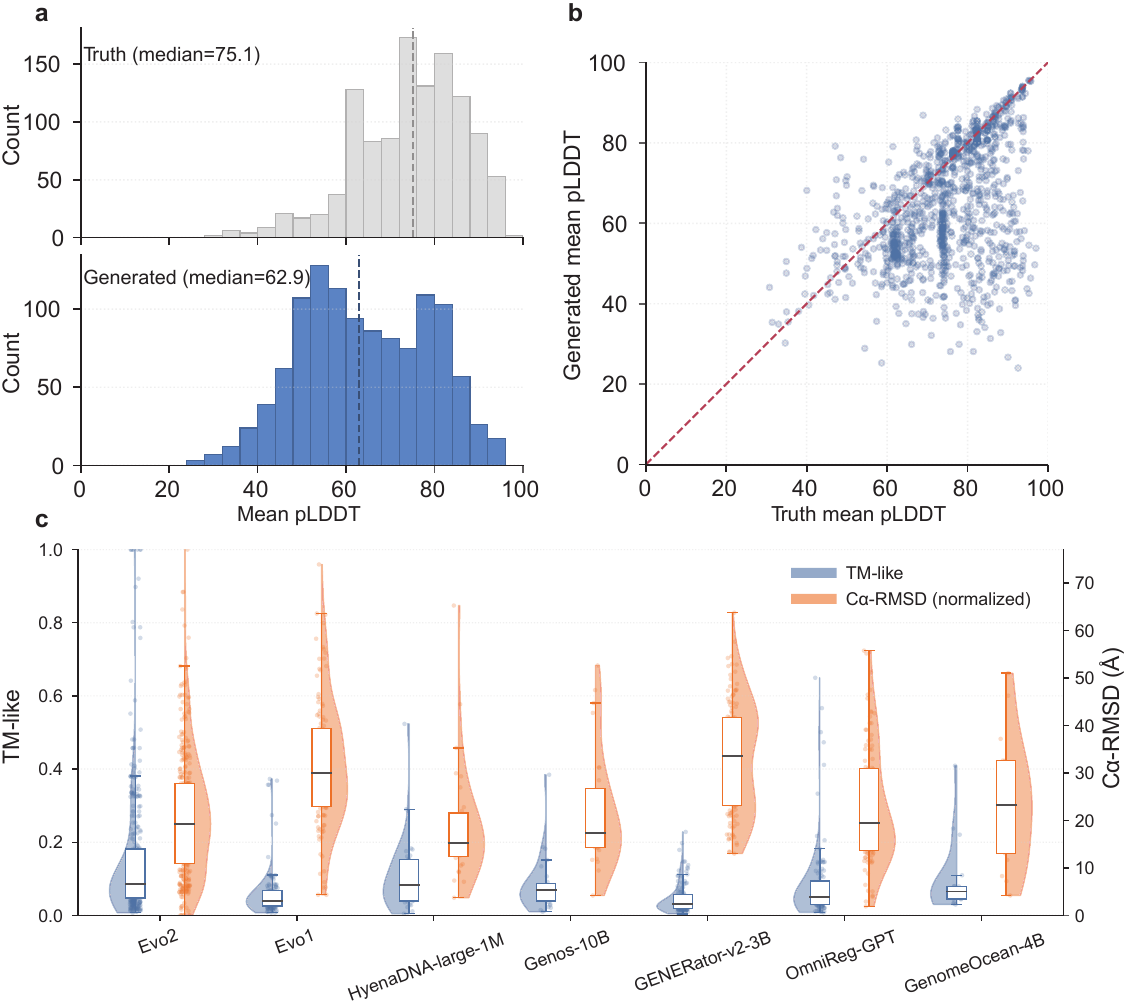}
  \caption{Structural confidence and fold similarity between generated and ground-truth proteins. (a) Distributions of per-target mean pLDDT for ground-truth (top, gray) and generated (bottom, blue) structures; dashed vertical lines indicate medians. (b) Pairwise comparison of generated versus ground-truth mean pLDDT for each target; the dashed diagonal denotes equality ($y = x$). (c) Model-group stratification of structural agreement, shown as raincloud summaries of TM-like similarity (higher is better) and C$\alpha$-RMSD (lower is better) for sequence-aligned, C$\alpha$-superposed structures. }
  \label{fig:plddt}
\end{figure*}

\begin{table*}[t]
  \centering
  \caption{Comparison of 3D structural similarity between generated and real sequences.}
  \label{tab:af3-table}
  \footnotesize
  \setlength{\tabcolsep}{6pt}
  \renewcommand{\arraystretch}{1.05}
  \sisetup{detect-weight=true, detect-family=true}
  \resizebox{0.8\textwidth}{!}{%
  \begin{tabular}{
    l
    S[table-format=3.0]
    S[table-format=1.3]
    S[table-format=2.2]
    S[table-format=2.2]
    S[table-format=2.2]
  }
    \toprule
    \textbf{Model group} &
    {\textbf{$n$}} &
    {\textbf{TM-like}} &
    {\textbf{C$\alpha$-RMSD}} &
    {\textbf{truth pLDDT}} &
    {\textbf{gen pLDDT}} \\
    \midrule
    Evo1                     & 121 & 0.040 & 30.04 & 73.73 & 58.94 \\
    Evo2                     & 353 & 0.085 & 19.33 & 82.03 & 76.33 \\
    Genos-10B                &  27 & 0.069 & 17.40 & 74.17 & 64.05 \\
    GenomeOcean-4B           &  17 & 0.065 & 23.24 & 77.63 & 72.99 \\
    GENERator-v2-3B & 122 & 0.032 & 33.50 & 72.94 & 53.88 \\
    HyenaDNA-large-1M        &  31 & 0.082 & 15.22 & 76.39 & 71.23 \\
    OmniReg-GPT              & 104 & 0.050 & 19.48 & 74.53 & 66.39 \\
    \bottomrule
  \end{tabular}%
  }
\end{table*}

\textbf{Global Structural Fidelity.}
In 1143 paired targets, the distribution of fold similarity was highly concentrated near zero, with a TM-like median of 0.054 and an interquartile range of 0.029--0.102. This was accompanied by large geometric biases, with a C$\alpha$-RMSD median of 23.74~\AA{} and an interquartile range of 15.66--35.19. Only 22 out of 1143 pairs had TM-like values $\ge 0.50$, and 13 out of 1143 exceeded 0.70, which is consistent with a small number of reconstructed structures that closely approximate the native structure. The relationship between sequence identity and structural similarity was moderate (Pearson $r = 0.370$), indicating that relatively small amino acid differences can significantly alter the predicted folded structure, and that plausibility at the nucleotide/codon level alone is insufficient to guarantee folded structure preservation.

The confidence-based signal further indicates reduced folding ability of the generated proteins. Compared to the true structures, the generated structures show a left-shifted pLDDT distribution (median 75.14$\rightarrow$62.88) and reduced global confidence (median pTM 0.50$\rightarrow$0.37), along with a higher proportion of predicted disorder (0.64$\rightarrow$0.88), as shown in Figure~\ref{fig:plddt}.

\textbf{Performance Differences Between Models.}
We stratified the performance by model group (Table~\ref{tab:af3-table}). Evo2 had the highest median fold recovery rate (17 out of 353 targets with TM-like $\ge 0.50$), followed by HyenaDNA-large-1M (1 out of 31 targets with TM-like $\ge 0.50$). OmniReg-GPT, despite a lower median TM-like, still produced a small number of high-fidelity results (3 out of 104 targets with TM-like $\ge 0.50$), indicating sporadic strong reconstruction results. These results suggest that larger capacity models can improve the consistency of folding levels, but structural fidelity is still far from stable across different targets.

\textbf{Preferred Patterns.}
Structural fidelity is clearly biased towards short proteins. The median true length of the top 5\% subset and the subset with TM-like~\mbox{$\ge 0.50$} were approximately 54 amino acids and approximately 52 amino acids, respectively (the longest sequence in the TM-like~\mbox{$\ge 0.50$} subset reached 169 amino acids). In contrast, transmembrane segment similarity for very long targets (e.g., \mbox{$>2000$} amino acids) approached zero (median 0.0057), indicating that maintaining long-range constraints and domain architecture remains extremely challenging under cue-constrained continuation.

The subsets with the highest similarity were enriched for phage-like entries with bacterial hosts, as shown in Figure~\ref{fig:af3_all}. Using ViroBench metadata, 95.7\% of the top three overall targets for each model belonged to host class A. The primary hosts in these subsets included clinically and environmentally significant bacterial species such as \textit{Yersinia pestis}, \textit{Escherichia coli}, \textit{Pseudomonas}, \textit{Klebsiella pneumoniae}, and \textit{Salmonella enteritidis}. Correspondingly, the best-performing families were enriched for common phage families (e.g., \textit{Peduoviridae}), while examples of eukaryotic host viruses were relatively rare (e.g., only one example from the \textit{Poxviridae} family among viruses with TM-like~\mbox{$\ge 0.50$}). This enrichment suggests that the model's reliable preservation of folded structures occurs primarily in relatively restricted phage proteins, while broader viral diversity and potentially more complex host-related restrictions are difficult to capture consistently.

\textbf{Implications for generation.}
These results highlight a persistent gap between sequence-level continuation and protein-level structural preservation. While prompt conditioning effectively stabilizes output length, fold fidelity depends on maintaining long-range, higher-order constraints that are not enforced by local sequence similarity alone. The strong concentration of successes among short proteins and bacteriophage-associated entries suggests that incorporating additional inductive biases, such as structure-aware objectives, conserved motif or domain constraints, or protein-language priors, will be essential for generating biologically interpretable CDS at scale. In practical terms, structural screening, for example by AlphaFold3 confidence and fold-similarity proxies, appears necessary to identify the small subset of outputs likely to retain native-like structure.

\subsection{Joint Analysis} \label{app:joint analysis}

\begin{figure*}[htbp]
  \centering
  \includegraphics[width=\textwidth]{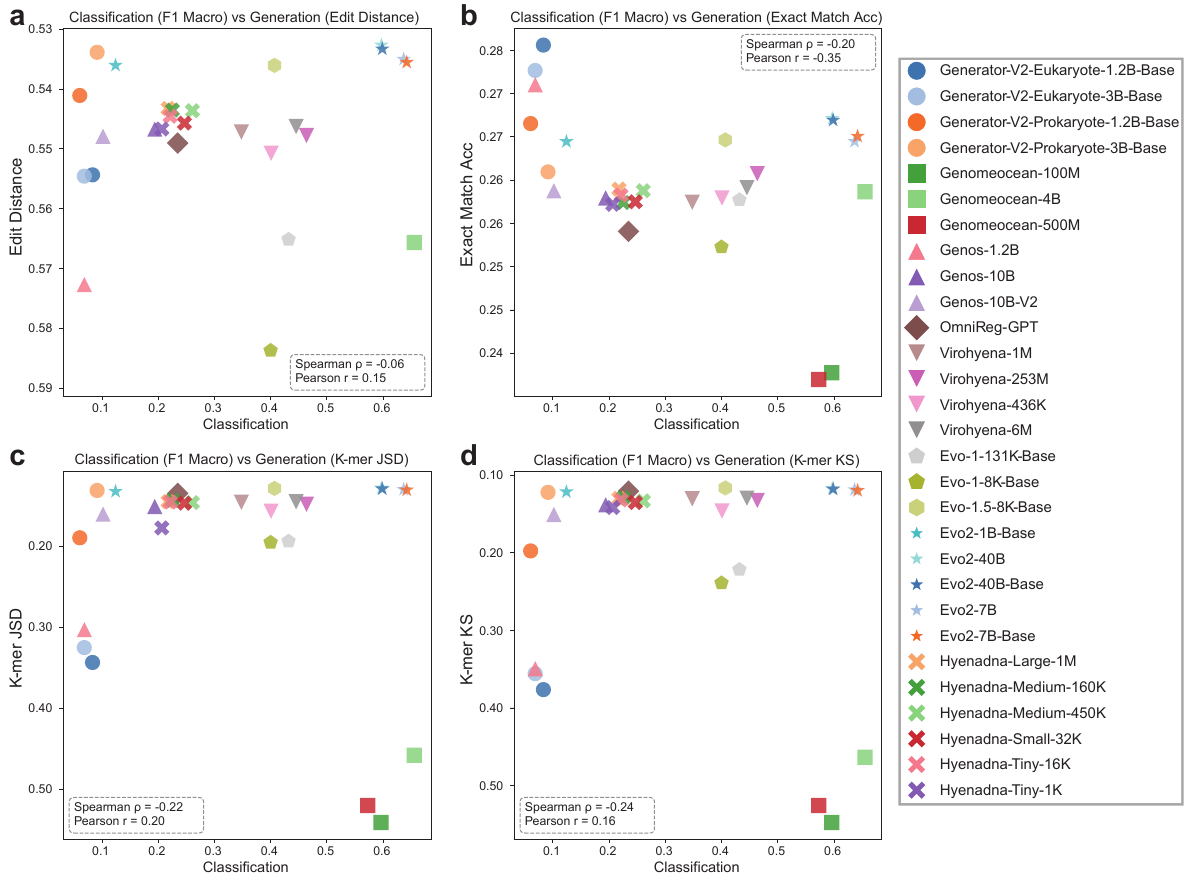}
  \caption{Joint analysis of classification and generation performance. Each panel plots macro-F1 (x-axis) against a generation metric (y-axis): (a) Edit Distance (lower is better), (b) Exact Match Accuracy (higher is better), (c) K-mer Jensen--Shannon Divergence (JSD; lower is better), and (d) K-mer Kolmogorov--Smirnov statistic (KS; lower is better). For panels with ``lower is better'' metrics, the y-axis is inverted so that higher values consistently indicate better performance. Each point denotes a model (colors/markers indicate model families and scales; see legend); Spearman's $\rho$ and Pearson's $r$ summarize the correlation in each panel.}
  \label{fig:cls_vs_gen}
\end{figure*}

We further perform a joint analysis that places models’ classification and generation abilities in the same view (Fig.~\ref{fig:cls_vs_gen}). Classification performance is summarized by the mean macro-F1 across 12 scenarios, and generation is assessed with complementary metrics including edit distance, exact match accuracy, and K-mer distribution measures. Overall, many models exhibit a clear trade-off between the two. Evo2 stands out as consistently strong across metrics, occupying the upper-right region of the plots . Our re-pretrained model, ViroHyena, does not reach Evo2’s peak performance, but it lies closer to the diagonal trend, indicating a more balanced profile that supports both discriminative and generative objectives. Together, these results provide additional evidence that our pre-training strategy improves overall capability rather than optimizing for a single metric.

\subsection{Case Study: Viral Sequence Analysis in Nipah} \label{app:case study}

We further showcase a case study on Nipah virus for viral sequence analysis. Using the pretrained base model, we perform both hierarchical taxonomic classification that infers labels across multiple ranks (e.g., realm, phylum, class, order, and family) and sliding-window PPL profiling along the genome. This yields an interpretable perplexity landscape that complements discrete rank-wise predictions by highlighting genomic regions that are well modeled versus atypical under the base model.
\begin{figure*}[htbp]
  \centering
  \includegraphics[width=\textwidth]{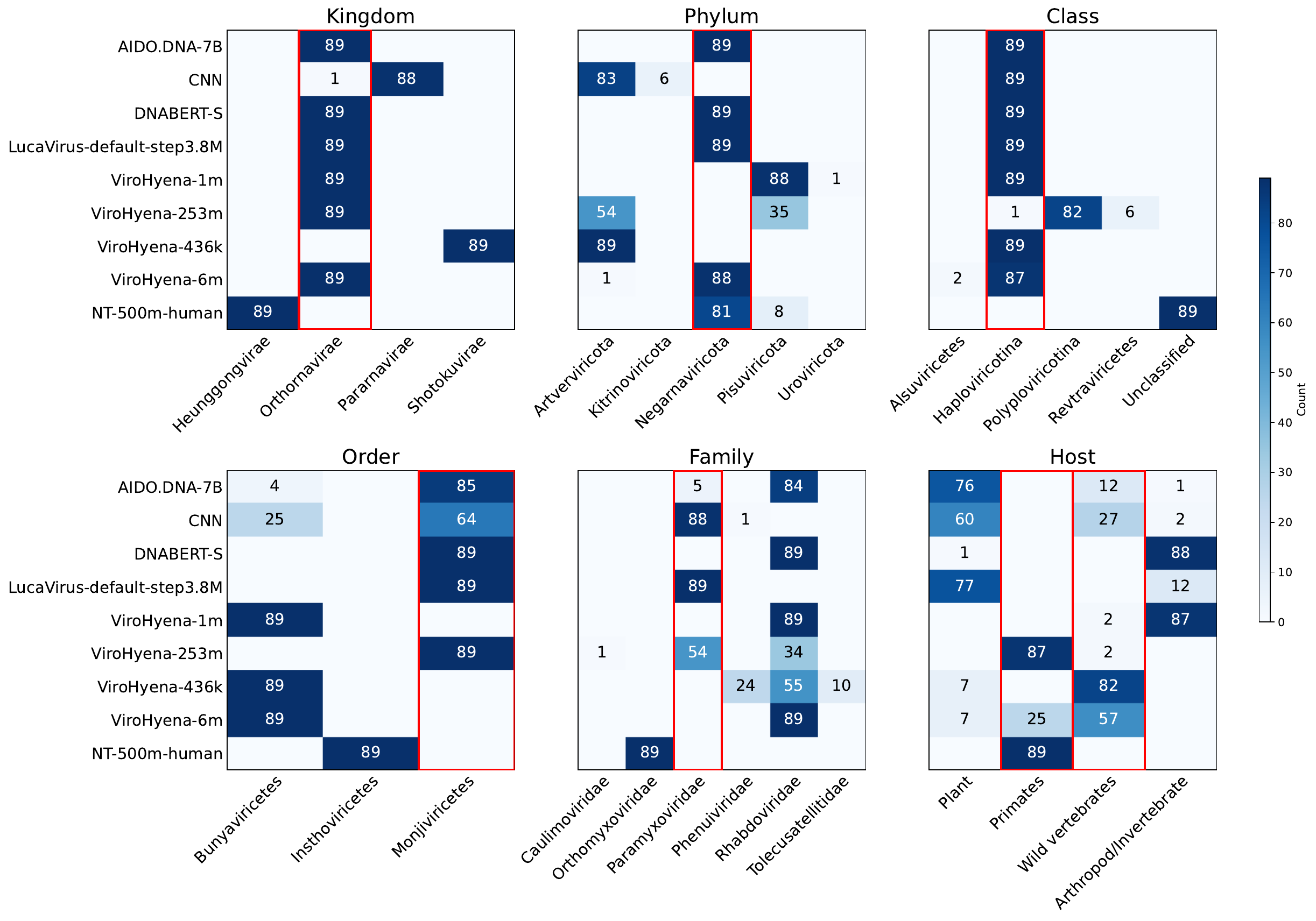}
  \caption{Confusion matrices for hierarchical taxonomic and host classification on 89 Nipah virus genomes across five taxonomic ranks (Kingdom–Family) and host groups. Rows denote models and cell values indicate the number of genomes predicted for each label; red boxes mark the correct label in each panel.}
  \label{fig:taxon_all_levels}
\end{figure*}

We curated 89 complete Nipah virus genomes, each identified by a GenBank accession (e.g., AF212302.2), and performed model-based analysis using a 512-nt sliding-window inference pipeline. Importantly, the evaluated Nipah virus (TaxID 3052225) was not included in any of our training or benchmark construction data, making this a strict out-of-distribution test. For each genome, we obtained window-level outputs from different pretrained models and aggregated them into genome-level predictions across hierarchical taxonomic ranks (kingdom, phylum, class, order, family) and host group. Figure~\ref{fig:taxon_all_levels} summarizes the resulting confusion matrices, where the red boxes mark the ground-truth labels. Overall, several virus-aware foundation models produce highly concentrated predictions at higher ranks, with most genomes assigned to the correct kingdom, phylum, and class columns, indicating robust recovery of coarse phylogenetic placement. In contrast, domain-mismatched baselines exhibit systematic off-target shifts already at higher ranks, reflecting weaker transfer to viral sequence semantics. As the taxonomy becomes more fine-grained, the task becomes noticeably harder: at the family level, predictions for many models disperse across closely related RNA viruses families rather than remaining in the red-box column, consistent with increased inter-family similarity under short nucleotide contexts. Host prediction shows the greatest ambiguity, with outputs often split between plausible vertebrate-associated categories and, for some baselines, substantial spurious assignments to unrelated host groups, highlighting that host signals are weaker and more confounded than taxonomic signals when inferred from nucleotide windows alone. Despite this strict held-out setting, the strong rank-wise concentration exhibited by several NFMs supports their effectiveness for analyzing previously unseen viral genomes and for providing actionable, hierarchy-aware sequence understanding from raw nucleotides.

\begin{figure*}[htbp]
  \centering
  \includegraphics[width=\textwidth]{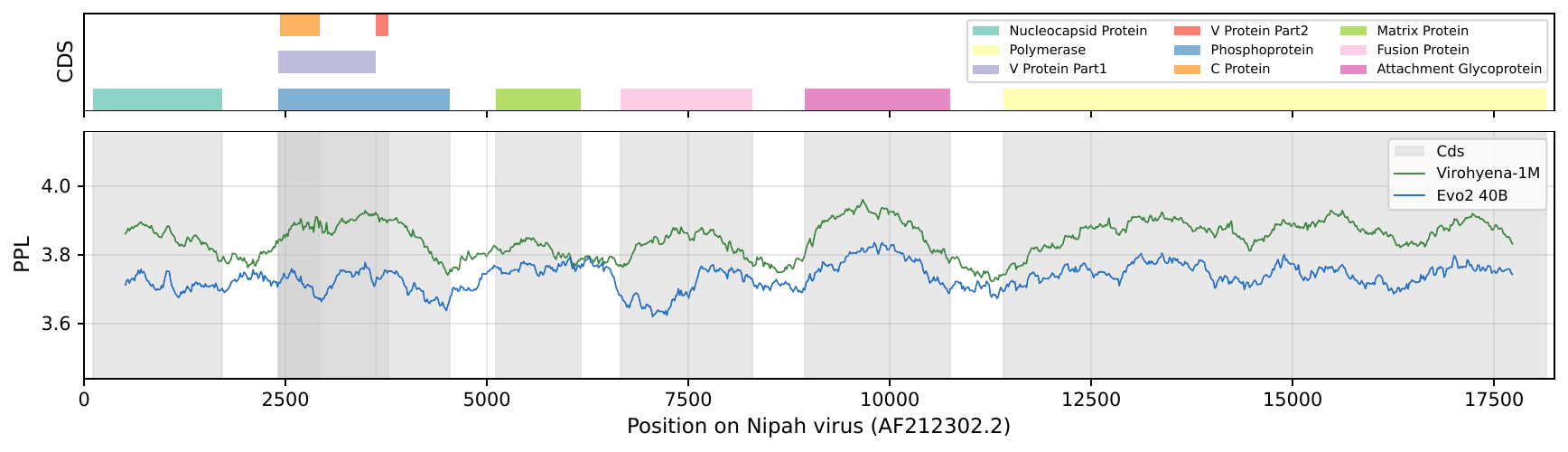}
  \caption{Sliding-window PPL profiles along the Nipah virus genome (AF212302.2). The top track shows annotated CDS regions, and the bottom panel compares position-wise PPL from ViroHyena-1M and Evo2-40B, with shaded intervals indicating CDS locations.}
  \label{fig:tracks_cds_lowcomp}
\end{figure*}

We further conducted a fine-grained likelihood profiling analysis on the Nipah virus reference genome AF212302.2. Specifically, we applied a sliding-window scheme with window size 512 nt and, for each window, computed the model perplexity only on the last 16 bases (i.e., evaluating next-token prediction under a fixed 512-nt context) to obtain a position-resolved PPL landscape along the genome. Overlaying this landscape with the genome annotation revealed a clear and reproducible pattern: PPL is systematically higher within annotated CDS regions (shaded intervals) than in non-coding segments, and this trend is consistent across models, while differing in overall calibration. These results indicate that coding regions are intrinsically harder for nucleotide language models to predict under local context, likely due to their richer compositional structure and stronger functional constraints compared with non-coding sequence. Overall, the analysis supports the conclusion that sliding-window PPL profiling can serve as an interpretable diagnostic signal, complementing discrete classification outputs by highlighting genomic regions with increased modeling difficulty that align with functional (protein-coding) organization.

Together, these analyses show that NFMs yield complementary signals on previously unseen viral genomes. Rank-wise classification recovers the correct taxonomic placement under a strict held-out setting, whereas sliding-window PPL profiling provides an interpretable, genome-resolved view that tracks functional organization and highlights coding regions as systematically harder to model. Collectively, these results support the pretrained base model as a practical tool for both taxonomy-oriented inference and fine-grained likelihood-based genome characterization.

\section{In-domain Pre-training with ViroBland}\label{app:ViroBland and ViroHyena}

\subsection{ViroHyena Pre-training Protocol}

We perform self-supervised pre-training for a HyenaDNA-style model based on the open-source Hyena architecture.
We adopt \emph{causal language modeling}, treating a DNA sequence as a character-level token sequence and
performing next-token prediction: given a prefix, the model predicts the next nucleotide, enabling it to learn
both local and long-range statistical regularities in DNA.

\paragraph{Objective and loss.}
Given a token sequence of length $L$, $\mathbf{x}=(x_1,\ldots,x_L)$, the model predicts the next token in an
autoregressive manner:
\begin{equation}
p(\mathbf{x})=\prod_{t=1}^{L-1} p(x_{t+1}\mid x_{\le t}).
\end{equation}
We minimize the cross-entropy (negative log-likelihood) over valid positions:
\begin{equation}
\mathcal{L} = -\sum_{t\in\Omega}\log p(x_{t+1}\mid x_{\le t}),
\end{equation}
where $\Omega$ denotes the set of valid training positions. Padding positions and ambiguous bases (e.g., \texttt{N})
are masked by setting their targets to \texttt{ignore\_index=-100}, and thus do not contribute to the loss.

\paragraph{Data and input construction.}
Pre-training is conducted on the \textit{ViroBland} corpus, utilizing the BED+FASTA format with pre-established data splits. During each training iteration, we sample genomic intervals $(\textit{contig}, \textit{start}, \textit{end})$ from the BED file and extract the corresponding sequences from the FASTA reference. Each sequence is standardized to a maximum length of $\texttt{max\_length}=8192$ tokens via truncation or padding. We employ a character-level tokenizer for the nucleotide alphabet $\{A, C, G, T, N\}$ and append an end-of-sequence (\texttt{<EOS>}) token to mark boundary conditions. Training pairs are generated using a causal one-position shift:
\begin{equation}
\mathbf{x}{\mathrm{in}}=\mathbf{x}{1:L-1},\qquad \mathbf{y}=\mathbf{x}_{2:L},
\end{equation}
ensuring that the model's prediction at each spatial index corresponds to the subsequent nucleotide.

\paragraph{Training configuration.}
The model and optimization hyperparameters follow the Hyena pre-training setup. Detailed configurations are reported
in Table~\ref{tab:pretrain_config}.

\begin{table}[htbp]
  \centering
  \caption{Pre-training configurations of our ViroHyena models.}
  \label{tab:pretrain_config}
  \small
  \renewcommand{\arraystretch}{1.12}
  \resizebox{\columnwidth}{!}{%
  \begin{tabular}{@{}lccccc@{}}
    \toprule
    \textbf{Model} & \textbf{\#Params} & \textbf{$d_{\text{model}}$} & \textbf{\#Layers} & \textbf{Max len} & \textbf{LR} \\
    \midrule
    ViroHyena-436K & 0.436M & 128  & 2  & 8192 & $3\times10^{-4}$ \\
    ViroHyena-1.6M & 1.6M   & 256  & 2  & 8192 & $3\times10^{-4}$ \\
    ViroHyena-6.6M & 6.6M   & 256  & 8  & 8192 & $3\times10^{-4}$ \\
    ViroHyena-253M & 253M   & 1024 & 20 & 8192 & $3\times10^{-4}$ \\
    \bottomrule
  \end{tabular}}
\end{table}

\subsection{Pre-training Results}\label{app:Pretrain Results}

\begin{figure}[htbp]
  \centering
  \includegraphics[width=\linewidth,height=0.45\textheight,keepaspectratio]{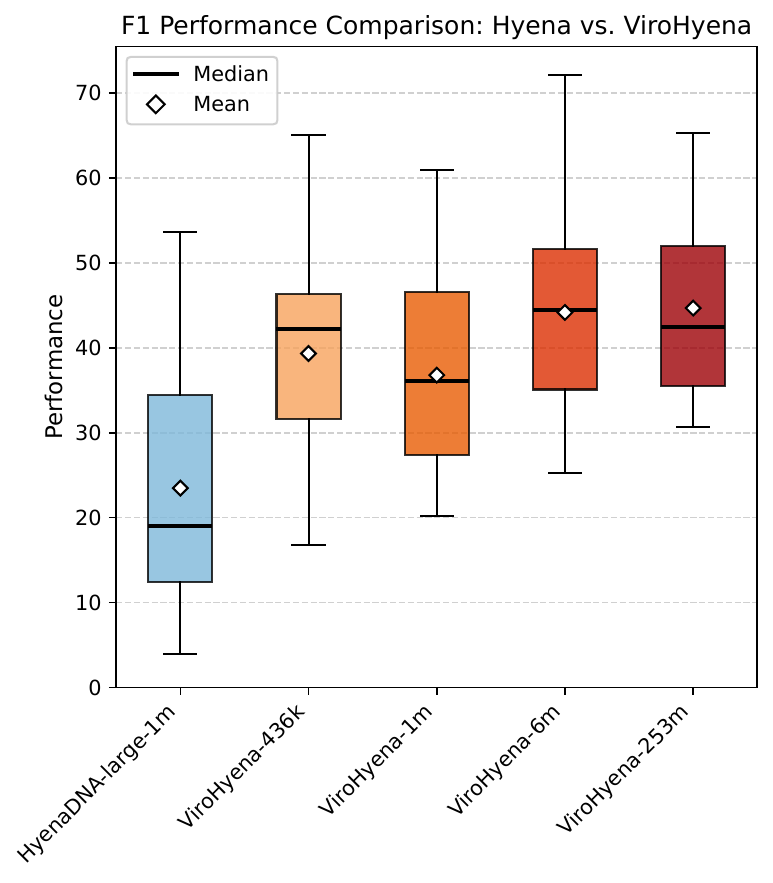}
  \caption{Distribution of Macro-F1 scores across all classification tasks for HyenaDNA-Large-1M and ViroHyena variants. Boxes indicate the interquartile range; horizontal lines and diamonds denote the median and mean, respectively; whiskers show the full range across tasks.}
  \label{fig:f1_distribution}
  \Description{Task-level F1 score distributions for HyenaDNA-Large-1M and four ViroHyena variants, showing median lines, mean diamonds, interquartile ranges, and whiskers.}
\end{figure}

\begin{figure}[htbp]
  \centering
  \includegraphics[width=\linewidth,height=0.45\textheight,keepaspectratio]{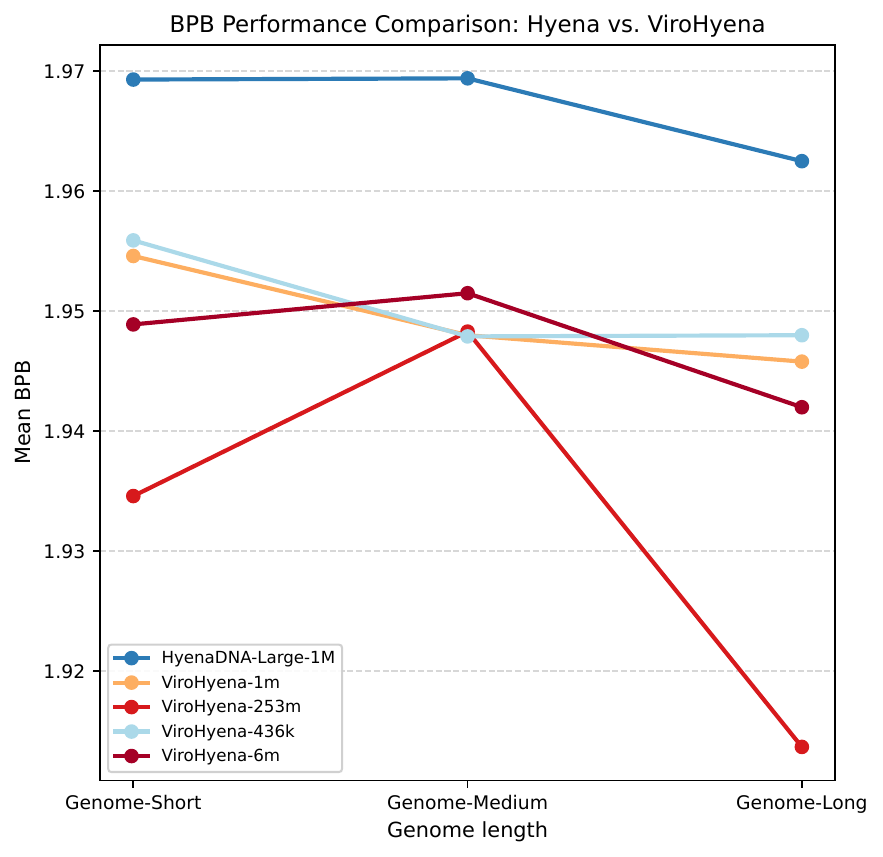}
  \caption{Mean BPB across genome-length buckets (short, medium, and long) for HyenaDNA-Large-1M and ViroHyena variants. Lower BPB indicates better likelihood modelling of nucleotide sequences; lines show how modelling quality varies with sequence length after in-domain pre-training.}
  \label{fig:bpb_length}
  \Description{Line plot of mean BPB over short/medium/long genome-length buckets for HyenaDNA-Large-1M and four ViroHyena variants.}
\end{figure}

We systematically evaluate the impact of in-domain pre-training along two dimensions: classification and generation. Specifically, we analyze changes in the Macro-F1 score (F1) across classification tasks to quantify the model's ability to capture virus-related functional signals, and examine BPB to assess improvements in modeling the underlying nucleotide distribution. By comparing the overall shifts of these metrics before and after pre-training, we aim to determine whether ViroBland pre-training yields consistent benefits on both the “classification–generation” axes, and how these gains vary with model scale.

Across all classification tasks, continued pre-training on ViroBland yielded substantial and consistent performance gains. The HyenaDNA-Large-1M baseline achieved a mean Macro-F1 of 23.48, whereas our virus-adapted ViroHyena family improved markedly even at the smallest scales. Specifically, ViroHyena-436K and ViroHyena-1M reached mean Macro-F1 scores of 39.32 and 36.77, corresponding to absolute gains of +15.84 and +13.29 points (a +67.5\% and +56.6\% relative improvement), respectively.

As we scaled the architecture, performance continued to increase: ViroHyena-6M achieved a mean Macro-F1 of 44.16 (+20.68 points; +88.1\% relative). Notably, the much larger ViroHyena-253M attained a highly similar score of 44.67 (+21.19 points; +90.2\% relative), suggesting that discriminative performance largely saturates beyond moderate scales on this dataset. Beyond the averages, the boxplot results (Fig.~\ref{fig:f1_distribution}) show a consistent upward shift of the entire performance distribution, with both the median and interquartile range moving toward higher scores. This indicates that ViroHyena's advantage is not confined to a small subset of tasks, but reflects broad and stable improvements in capturing virus-specific functional signals.

From generation results, in-domain pre-training on ViroBland yields lower BPB across genome-length buckets, indicating improved likelihood modeling of nucleotide sequences. The pre-training baseline HyenaDNA-Large-1M attains BPB values of 1.9693,1.9694,\-1.9625 on the short/medium/long genome buckets, whereas the ViroHyena family is consistently lower. This trend is further visualized in Fig.~\ref{fig:bpb_length}, which shows how modeling quality varies with sequence length after in-domain pre-training. For example, ViroHyena-1M improves to 1.9546/1.9480/1.9458, and ViroHyena-436k reaches 1.9559/1.9479/1.9480. As model scale increases, some buckets further benefit: ViroHyena-6M achieves 1.9420 on the long-genome bucket. Notably, ViroHyena-253M attains the lowest BPB on long genomes, 1.9137 (a reduction of 0.0488 relative to the long-genome baseline), suggesting that larger models exhibit a stronger advantage in long-range sequence modeling. Overall, ViroHyena shows BPB reductions across all three buckets, with larger improvements on the long-genome regime.

\section{Ablation Studies}
We conduct additional ablation studies to examine whether the main conclusions of ViroBench are sensitive to input segmentation, window configuration, model architecture, pretraining data composition, tokenization strategy, and model scale. Unless otherwise specified, all ablations use the same data splits, downstream classifier, pooling strategy, and evaluation protocol as the main classification experiments. We report Macro-F1 scores in percentages.
\subsection{Effect of Sequence Segmentation}
In the main experiments, we segment viral genomes into non-overlapping fixed-length windows, with an additional tail window to ensure coverage of the sequence end. Although fixed-length windows are not always aligned with biological units, this design reflects a practical viral surveillance scenario in which models often need to identify viruses from local genomic fragments rather than complete genomes. To directly assess whether the fixed-window design affects our comparative conclusions, we evaluate a contig-based segmentation strategy on the ALL-virus classification tasks. Instead of slicing sequences into fixed-length windows, this setting preserves contig boundaries as the input units. Table~\ref{tab:contig_summary} summarizes the average Macro-F1 over the four ALL-virus classification scenarios, including taxonomy and host prediction under both Genus-disjoint and Temporal splits. Full per-task results are provided in Table~\ref{tab:contig_segmentation_ablation}.

\begin{table}[t]
  \caption{Comparison between fixed-window and contig-based segmentation on ALL-virus classification tasks. F1 denotes the average Macro-F1 over ALL Taxonomy-G, ALL Taxonomy-T, ALL Host-G, and ALL Host-T. Ranks are computed within each segmentation setting.}
  \label{tab:contig_summary}
  \centering
  \small
  \setlength{\tabcolsep}{5pt}
  \renewcommand{\arraystretch}{1.08}
  \begin{tabular}{@{}lcc@{}}
    \toprule
    \textbf{Model} & \textbf{Fixed F1 (Rank)} & \textbf{Contig F1 (Rank)} \\
    \midrule
    CNN & 37.11 (8) & 31.79 (9) \\
    BiLSTM & 62.45 (4) & 32.66 (7) \\
    LucaOne-Default-Step36M & 64.18 (3) & 50.00 (3) \\
    LucaVirus-Default-Step3.8M & 70.05 (1) & 57.21 (1) \\
    GenomeOcean-100M & 57.87 (6) & 45.67 (4) \\
    Evo2-1B-Base & 13.42 (15) & 13.84 (15) \\
    NTv3-650M-Pre & 27.80 (12) & 21.54 (13) \\
    AIDO.DNA-300M & 64.54 (2) & 50.91 (2) \\
    Caduceus-PH & 31.79 (10) & 41.83 (6) \\
    Caduceus-PS & 31.43 (11) & 22.65 (11) \\
    HyenaDNA-Large-1M & 22.47 (14) & 18.30 (14) \\
    DNABERT-2-117M & 27.58 (13) & 22.08 (12) \\
    DNABERT-6 & 35.93 (9) & 32.32 (8) \\
    ViroHyena-6M & 43.77 (7) & 25.45 (10) \\
    \bottomrule
  \end{tabular}
\end{table}

Contig-based segmentation generally leads to lower absolute performance than fixed-window segmentation. This may be because contigs introduce greater variation in input length, reduce the number of training instances, and make the downstream classifier more sensitive to highly uneven sequence coverage. Nevertheless, the relative ordering of models remains highly consistent across the two segmentation strategies. The Spearman rank correlation between fixed-window and contig-based results is 0.93, indicating that segmentation mainly affects absolute performance rather than changing the comparative conclusions of our benchmark.

\begin{table*}[t]
  \caption{Ablation on contig-based sequence segmentation. Values are reported as Macro-F1 scores with standard deviation in parentheses (\%).}
  \label{tab:contig_segmentation_ablation}
  \centering
  \scriptsize
  \setlength{\tabcolsep}{3pt}
  \renewcommand{\arraystretch}{1.08}
  \resizebox{\textwidth}{!}{
  \begin{tabular}{@{}lcccccccccccc@{}}
    \toprule
    \multirow{2}{*}{\textbf{Model}} 
    & \multicolumn{4}{c}{\textbf{ALL}} 
    & \multicolumn{4}{c}{\textbf{DNA}} 
    & \multicolumn{4}{c}{\textbf{RNA}} \\
    \cmidrule(lr){2-5} \cmidrule(lr){6-9} \cmidrule(lr){10-13}
    & \textbf{Taxon-G} & \textbf{Taxon-T} & \textbf{Host-G} & \textbf{Host-T}
    & \textbf{Taxon-G} & \textbf{Taxon-T} & \textbf{Host-G} & \textbf{Host-T}
    & \textbf{Taxon-G} & \textbf{Taxon-T} & \textbf{Host-G} & \textbf{Host-T} \\
    \midrule
    CNN & 22.40(11.64) & 9.63(3.79) & 67.44(7.37) & 27.68(9.46) & 21.54(11.96) & 17.67(6.26) & 31.37(11.55) & 15.06(12.01) & 22.53(8.06) & 22.03(5.71) & 56.29(3.77) & 47.93(3.19) \\
    LucaOne-Default-Step36M & 50.35(25.27) & 36.40(22.34) & 75.22(12.35) & 38.03(21.21) & 73.70(10.33) & 50.58(8.46) & 55.63(0.71) & 47.78(1.11) & 62.91(23.50) & 51.27(25.35) & 64.47(22.52) & 27.73(22.41) \\
    LucaVirus-Default-Step3.8M & 58.81(21.22) & 45.34(20.04) & 80.32(8.07) & 44.36(16.25) & 74.41(13.33) & 56.69(6.22) & 59.61(1.22) & 42.22(3.01) & 71.58(14.01) & 58.23(27.41) & 74.65(9.17) & 45.28(17.46) \\
    DNABERT-S & 43.41(14.19) & 29.78(13.10) & 76.36(4.35) & 26.50(3.57) & 63.99(4.47) & 42.45(5.63) & 63.07(6.93) & 34.97(4.71) & 57.58(17.21) & 45.02(14.65) & 72.82(4.42) & 34.84(3.99) \\
    GenomeOcean-100M & 50.07(21.37) & 33.63(19.04) & 65.76(20.70) & 33.20(16.75) & 71.32(7.03) & 44.06(6.16) & 56.37(1.80) & 37.93(3.25) & 57.27(23.70) & 41.87(20.66) & 60.39(24.01) & 28.11(22.30) \\
    Evo2-1B-Base & 5.16(2.51) & 4.52(2.14) & 40.98(8.50) & 4.70(2.21) & 7.92(0.37) & 9.59(1.23) & 22.82(3.51) & 21.99(0.82) & 11.39(2.60) & 13.44(3.00) & 30.76(4.97) & 13.06(7.43) \\
    NTv3-650M-Pre & 26.60(25.50) & 17.28(18.54) & 41.29(9.04) & 1.00(0.00) & 25.36(23.17) & 23.84(14.22) & 40.08(4.52) & 30.78(12.71) & 34.28(28.58) & 24.26(19.97) & 26.47(14.67) & 5.78(0.00) \\
    AIDO.DNA-300M & 55.37(24.06) & 38.02(22.62) & 72.98(15.36) & 37.25(20.01) & 77.46(5.34) & 50.27(5.73) & 58.32(1.12) & 46.36(6.56) & 63.69(20.65) & 50.94(27.33) & 64.69(21.52) & 36.40(18.67) \\
    Caduceus-PH & 42.87(0.00) & 26.34(0.00) & 69.36(0.00) & 28.75(0.00) & 35.72(0.00) & 16.15(0.00) & 40.61(0.00) & 31.16(0.00) & 52.73(0.00) & 37.97(0.00) & 55.79(0.00) & 39.66(0.00) \\
    Caduceus-PS & 24.54(21.26) & 11.12(9.17) & 51.78(20.51) & 3.17(3.06) & 38.79(10.12) & 19.56(0.00) & 49.19(1.03) & 32.28(0.80) & 37.10(22.11) & 41.64(50.58) & 39.91(20.13) & 14.98(15.94) \\
    HyenaDNA-Large-1M & 14.50(17.30) & 8.66(10.21) & 44.01(21.50) & 6.02(8.69) & 19.93(22.68) & 13.61(5.58) & 37.94(7.68) & 37.80(5.08) & 21.55(13.84) & 18.58(9.77) & 32.10(20.21) & 5.78(0.00) \\
    DNABERT-2-117M & 22.26(20.27) & 13.70(12.80) & 46.27(9.54) & 6.09(8.81) & 37.51(7.09) & 15.26(4.86) & 42.80(3.98) & 25.27(8.01) & 30.30(17.38) & 21.19(8.80) & 25.33(8.73) & 5.78(0.00) \\
    DNABERT-6 & 8.35(4.51) & 5.22(2.36) & 47.85(6.49) & 7.84(6.02) & 17.16(2.57) & 7.16(0.58) & 25.70(3.50) & 17.92(0.36) & 13.98(3.02) & 14.90(3.05) & 47.03(8.51) & 29.12(9.06) \\
    BiLSTM & 36.92(49.43) & 22.95(31.27) & 44.23(52.41) & 26.55(22.75) & 47.57(41.39) & 31.84(25.47) & 38.71(25.06) & 35.07(16.63) & 39.45(43.64) & 33.24(32.19) & 41.11(47.80) & 32.22(37.39) \\
    ViroHyena-6M & 25.15(20.52) & 18.01(15.18) & 43.83(13.25) & 14.79(11.94) & 36.39(23.69) & 19.32(5.56) & 46.54(4.40) & 47.42(4.65) & 35.25(23.97) & 25.01(13.63) & 34.16(18.36) & 12.82(12.20) \\
    \bottomrule
  \end{tabular}
  }
\end{table*}

\subsection{Effect of Window Configuration}

\begin{table}[htbp]
  \caption{Ablation on windowing configuration on \texttt{ALL taxon genus}. Each setting is denoted as $W/N/K$, where $W$ is the window size, $N$ is the number of windows, and $K$ is the number of selected windows for validation and testing. Values are reported in \%.}
  \label{tab:window_ablation}
  \centering
  \small
  \renewcommand{\arraystretch}{1.1}

  \begin{tabular*}{\columnwidth}{@{\extracolsep{\fill}}lccc@{}}
    \toprule
    \textbf{Model} & \textbf{512/8/64} & \textbf{1024/4/32} & \textbf{2048/2/16} \\
    \midrule
    AIDO.DNA-7B & 95.19 & 95.34 & 94.57 \\
    BiRNA-BERT & 72.25 & 67.75 & 50.20 \\
    Caduceus-PS & 75.54 & 72.84 & 72.97 \\
    CNN & 77.17 & 67.71 & 60.22 \\
    DNABERT-2-117M & 74.43 & 72.69 & 79.03 \\
    DNABERT-S & 92.90 & 93.96 & 94.82 \\
    Evo2 1B-Base & 56.01 & 51.75 & 39.43 \\
    Evo2 7B & 96.25 & 95.63 & 95.48 \\
    Gena-lm-bert-Base-t2t & 90.89 & 90.28 & 91.61 \\
    GENERator-v2-prokaryote-3b-Base & 23.09 & 20.13 & 30.24 \\
    GenomeOcean-4B & 95.51 & 95.96 & 96.26 \\
    GROVER & 82.38 & 79.79 & 77.58 \\
    HyenaDNA-Large-1M & 46.00 & 38.84 & 43.13 \\
    LucaVirus-default-step3.8M & 97.61 & 97.57 & 97.52 \\
    NT-2.5B-1000g & 16.76 & 63.88 & 72.77 \\
    NT-2.5B-ms & 13.86 & 50.29 & 58.41 \\
    NTv2-500M-ms & 14.83 & 87.08 & 91.62 \\
    NTv3-650M-post & 89.65 & 89.92 & 86.76 \\
    OmniReg-GPT & 47.68 & 46.70 & 58.36 \\
    RiNALMo & 80.96 & 80.50 & 76.94 \\
    RNABERT & 57.43 & 50.53 & 38.47 \\
    ViroHyena-253M & 75.64 & 75.95 & 81.07 \\
    \bottomrule
  \end{tabular*}
\end{table}

We conduct ablation studies on the windowing strategy under a fixed base budget. Each configuration is defined by a triplet $(W,N,K)$, where $W$ is the window size (in bases), $N$ is the number of concatenated windows per input, and $K$ is the number of windows sampled during validation and testing. To maintain a constant total input length of $W \times N = 4096$, we evaluate three specific settings:
$$(W,N,K) \in \{(512,8,64), (1024,4,32), (2048,2,16)\}.$$
All other experimental components, including data splits, training protocols and classifier head architectures, remain identical across all settings. As shown in Table~\ref{tab:window_ablation}, the optimal windowing configuration varies across models; accordingly, rather than enforcing a single universal setting, we select the best-performing $(W,N,K)$ for each model in subsequent experiments to avoid underestimating performance due to a suboptimal input construction. We additionally ablate the learning rate for the downstream classification head, comparing values in $\{10^{-2},10^{-3},10^{-4}\}$ while keeping all other hyperparameters constant.

\subsection{Effect of Architecture and Viral Pretraining}\label{app:architecture_ablation}

To examine how model architecture and viral-domain pretraining affect downstream performance, we provide controlled comparisons within the same backbone family: DNABERT-2-117M versus DNABERT2-ViroBench, and Caduceus-PS versus Caduceus-ViroBench.

The DNABERT comparison shows a consistent benefit from viral-domain adaptation. DNABERT2-ViroBench outperforms DNABERT-2-117M in every reported setting. The gains are especially clear for host prediction, where DNABERT2-ViroBench improves ALL Host-T from 43.01 to 56.73, DNA Host-G from 36.43 to 62.35, RNA Host-G from 50.89 to 70.21, and RNA Host-T from 31.59 to 44.90. Improvements are also observed for taxonomy classification, including ALL Taxon-G, DNA Taxon-G, and RNA Taxon-G.

A similar pattern appears for the Caduceus family. Caduceus-ViroBench improves over Caduceus-PS across all columns in Table~\ref{tab:architecture_ablation}. The improvement is particularly pronounced for taxonomy prediction, including ALL Taxon-G from 48.88 to 58.43, ALL Taxon-T from 25.71 to 41.75, DNA Taxon-G from 43.06 to 58.37, and RNA Taxon-G from 62.54 to 73.79. Host prediction also benefits, although the magnitude is more moderate in some DNA and RNA host settings.

These results indicate that, across different model architectures, viral-domain pretraining consistently improves performance on ViroBench classification tasks.

\begin{table*}[h]
  \caption{Ablation on model architecture. Values are reported as Macro-F1 scores with standard deviation in parentheses (\%).}
  \label{tab:architecture_ablation}
  \centering
  \scriptsize
  \setlength{\tabcolsep}{3pt}
  \renewcommand{\arraystretch}{1.08}
  \resizebox{\textwidth}{!}{
  \begin{tabular}{@{}lcccccccccccc@{}}
    \toprule
    \multirow{2}{*}{\textbf{Model}} 
    & \multicolumn{4}{c}{\textbf{ALL}} 
    & \multicolumn{4}{c}{\textbf{DNA}} 
    & \multicolumn{4}{c}{\textbf{RNA}} \\
    \cmidrule(lr){2-5} \cmidrule(lr){6-9} \cmidrule(lr){10-13}
    & \textbf{Taxon-G} & \textbf{Taxon-T} & \textbf{Host-G} & \textbf{Host-T}
    & \textbf{Taxon-G} & \textbf{Taxon-T} & \textbf{Host-G} & \textbf{Host-T}
    & \textbf{Taxon-G} & \textbf{Taxon-T} & \textbf{Host-G} & \textbf{Host-T} \\
    \midrule
    DNABERT-2-117M & 47.03(3.41) & 32.10(2.50) & 66.84(0.76) & 43.01(0.77) & 47.65(4.52) & 25.96(2.26) & 36.43(2.51) & 35.64(1.44) & 61.79(2.49) & 37.19(3.47) & 50.89(1.86) & 31.59(0.84) \\
    ViroDNABERT2 & 53.72(2.85) & 32.43(2.22) & 77.57(0.74) & 56.73(0.80) & 59.02(6.55) & 30.03(2.03) & 62.35(3.92) & 38.95(3.95) & 73.79(2.50) & 41.25(3.33) & 70.21(7.00) & 44.90(1.71) \\
    Caduceus-PS & 48.88(4.11) & 25.71(2.53) & 67.91(0.62) & 43.17(3.19) & 43.06(4.16) & 22.96(2.87) & 43.27(11.55) & 35.49(2.11) & 62.54(2.56) & 31.02(2.73) & 59.74(0.15) & 35.30(2.35) \\
    ViroCaduceus & 58.43(2.42) & 41.75(5.05) & 70.90(0.50) & 50.13(1.42) & 58.37(1.37) & 31.95(2.68) & 47.70(0.48) & 39.47(1.07) & 73.79(2.74) & 39.49(2.87) & 63.09(9.67) & 38.44(2.34) \\
    \bottomrule
  \end{tabular}
  }
\end{table*}

\subsection{Effect of Tokenization and Model Scale}

We also study tokenization and scale under the Hyena architecture. This controlled setting allows us to compare three tokenization strategies: BPE, fixed Kmer6 tokenization, and Char-level tokenization. 

As shown in Table~\ref{tab:tokenization_ablation}, BPE achieves the best overall performance, followed by Kmer6 and then Char-level tokenization. This pattern suggests that viral sequence modeling benefits from adaptive tokenization. BPE can merge recurring viral subsequences into variable-length units, which may preserve informative local motifs while avoiding overly long fixed vocabularies. In contrast, Kmer6 imposes a fixed segmentation regardless of sequence context, and Char-level tokenization decomposes motifs into individual nucleotides, forcing the model to reconstruct local biological patterns from very short units. Scaling the Hyena model generally provides some benefit, but the gain is smaller and less systematic than the gain from choosing an appropriate tokenization strategy. These findings indicate that tokenizer design is a central modeling choice for viral NFMs and should be considered alongside architecture and parameter count.

\begin{table*}[h]
  \caption{Ablation on tokenization strategy. Values are reported as Macro-F1 scores with standard deviation in parentheses (\%).}
  \label{tab:tokenization_ablation}
  \centering
  \scriptsize
  \setlength{\tabcolsep}{3pt}
  \renewcommand{\arraystretch}{1.08}
  \resizebox{\textwidth}{!}{
  \begin{tabular}{@{}lcccccccccccc@{}}
    \toprule
    \multirow{2}{*}{\textbf{Model}} 
    & \multicolumn{4}{c}{\textbf{ALL}} 
    & \multicolumn{4}{c}{\textbf{DNA}} 
    & \multicolumn{4}{c}{\textbf{RNA}} \\
    \cmidrule(lr){2-5} \cmidrule(lr){6-9} \cmidrule(lr){10-13}
    & \textbf{Taxon-G} & \textbf{Taxon-T} & \textbf{Host-G} & \textbf{Host-T}
    & \textbf{Taxon-G} & \textbf{Taxon-T} & \textbf{Host-G} & \textbf{Host-T}
    & \textbf{Taxon-G} & \textbf{Taxon-T} & \textbf{Host-G} & \textbf{Host-T} \\
    \midrule
    Hyena-Local-BPE-253M & 65.24(2.05) & 41.54(3.69) & 81.86(0.44) & 56.66(1.50) & 70.43(3.24) & 39.62(2.38) & 70.23(3.02) & 46.58(3.25) & 79.17(1.71) & 46.63(2.83) & 84.85(0.93) & 52.52(3.27) \\
    Hyena-Local-BPE-436K & 64.77(2.00) & 37.82(1.96) & 79.05(1.03) & 54.18(3.09) & 67.19(2.57) & 38.94(2.48) & 70.90(2.86) & 47.22(2.49) & 81.68(5.94) & 45.08(2.58) & 81.69(1.18) & 51.02(5.69) \\
    Hyena-Local-BPE-6p6M & 65.21(1.71) & 40.51(2.03) & 80.52(0.74) & 56.01(5.92) & 70.88(3.71) & 40.50(2.73) & 69.96(4.21) & 46.95(2.76) & 79.61(1.82) & 43.63(2.31) & 80.43(3.03) & 45.50(0.26) \\
    Hyena-Local-Char-1p6M & 51.78(5.14) & 29.96(2.77) & 68.59(1.29) & 44.25(2.30) & 52.95(6.01) & 28.11(4.93) & 47.85(12.29) & 38.90(2.35) & 59.98(3.29) & 35.99(2.27) & 54.69(4.20) & 34.03(4.01) \\
    Hyena-Local-Char-253M & 57.26(3.15) & 36.37(2.62) & 72.04(0.84) & 49.42(4.45) & 55.59(4.98) & 29.54(5.32) & 48.64(2.11) & 38.56(0.29) & 65.08(3.80) & 39.10(3.10) & 67.83(8.30) & 33.58(1.15) \\
    Hyena-Local-Char-436K & 53.01(3.26) & 31.26(2.19) & 70.16(2.27) & 43.72(4.91) & 47.68(2.78) & 24.89(2.07) & 57.01(8.24) & 39.78(3.37) & 62.93(4.57) & 34.27(2.50) & 65.85(7.18) & 35.31(1.86) \\
    Hyena-Local-Char-6p6M & 58.71(2.35) & 38.43(1.27) & 74.28(1.96) & 48.50(0.78) & 55.81(2.89) & 34.89(5.54) & 55.38(9.97) & 36.99(0.48) & 77.54(6.55) & 38.84(1.98) & 74.98(4.81) & 39.37(5.66) \\
    Hyena-Local-Kmer6-1p6M & 64.58(1.50) & 43.07(2.92) & 77.56(0.19) & 51.79(0.28) & 67.46(3.04) & 40.62(4.20) & 71.42(1.00) & 47.30(1.10) & 76.27(2.41) & 46.00(1.39) & 80.98(2.07) & 41.77(0.81) \\
    Hyena-Local-Kmer6-253M & 59.71(1.98) & 42.25(3.93) & 80.37(0.24) & 53.75(3.68) & 61.57(3.69) & 37.72(2.75) & 69.75(2.66) & 41.17(1.54) & 75.02(2.66) & 44.10(2.42) & 73.47(7.17) & 52.14(3.29) \\
    Hyena-Local-Kmer6-436K & 63.01(1.98) & 42.93(1.89) & 75.64(1.45) & 49.13(0.75) & 64.00(3.99) & 39.65(2.32) & 67.16(1.40) & 46.52(1.58) & 77.40(2.65) & 45.28(2.14) & 77.59(0.89) & 41.84(2.36) \\
    Hyena-Local-Kmer6-6p6M & 59.98(1.98) & 39.71(3.84) & 76.94(0.58) & 51.39(2.54) & 59.94(4.98) & 38.72(2.63) & 74.01(2.33) & 39.78(3.68) & 73.61(3.14) & 42.63(0.79) & 73.23(5.16) & 39.95(0.51) \\
    \bottomrule
  \end{tabular}
  }
\end{table*}

\subsection{Effect of prefix length ablation on CDS generation}
We conduct prefix length ablation experiments to validate the rationale for using a 129-bp prompt in CDS generation. Since CDSs are organized by codons, all tested prefix lengths are multiples of three. We compare three settings, including a shorter 90-bp prefix, the default 129-bp prefix, and a slightly longer 135-bp prefix.

Table~\ref{tab:prefix_ablation} summarizes the results on representative autoregressive models across the short and medium CDS regimes. Compared with the 90-bp prefix, the 129-bp prefix consistently improves CDS validity. For ViroHyena-6M, the CDS success rate increases from 0.34/0.03 to 1.03/0.04 on the short/medium regimes. For Evo2-7B-base, it increases from 0.56/0.19 to 0.88/0.27. This suggests that 90-bp may provide insufficient upstream coding context for models to reliably infer that the continuation should follow CDS-like structural constraints rather than generic nucleotide statistics. In contrast, extending the prompt to 135-bp does not yield consistent further gains. Although it slightly improves CDS validity for ViroHyena-6M, it leads to a clear increase in K-mer JSD on the short regime for both ViroHyena-6M and Evo2-7B-base, indicating degraded distributional fidelity. For Evo2-7B-base, the short-regime CDS success rate also drops from 0.88 to 0.73 when increasing the prefix from 129-bp to 135-bp.

These results indicate that 129-bp is a stable empirical trade-off: it provides more coding context information than-90 bp while avoiding the poor stability observed at 135-bp. Therefore, we use 129-bp as the default CDS prefix length in our generation evaluation.

\begin{table}[t]
\centering
\caption{Prefix length ablation for CDS generation. CDS success rate is reported in percentage (\%, higher is better). K-mer JSD is scaled by 100 for readability (lower is better). S and M denote the short and medium CDS regimes, respectively.}
\label{tab:prefix_ablation}
\resizebox{\linewidth}{!}{
\begin{tabular}{lccc}
\toprule
Model & Prefix length (bp) & CDS Success (S/M) & K-mer JSD (S/M) \\
\midrule
ViroHyena-6M & 90  & 0.34 / 0.03 & 15.03 / 14.40 \\
ViroHyena-6M & 129 & 1.03 / 0.04 & 15.88 / 13.16 \\
ViroHyena-6M & 135 & 1.21 / 0.06 & 21.29 / 13.03 \\
\midrule
Evo2-7B-base & 90  & 0.56 / 0.19 & 14.71 / 13.73 \\
Evo2-7B-base & 129 & 0.88 / 0.27 & 15.47 / 12.72 \\
Evo2-7B-base & 135 & 0.73 / 0.31 & 20.60 / 12.33 \\
\bottomrule
\end{tabular}
}
\end{table}

\section{Computational Cost and Efficiency}
All experiments were conducted on the QiZhi Cluster (Shanghai Institute of Intelligent Computing), utilizing NVIDIA H200 GPUs.To assess computational overhead, we use the end-to-end wall-clock time (in minutes) for taxonomy classification on the all-virus dataset under the genus-disjoint split strategy as a proxy for relative cost. For models requiring precomputed embeddings, this duration encompasses batch extraction, caching, MLP head training, and final evaluation. For the baseline CNN, we report the total end-to-end training and evaluation time. These statistics, summarized in Table~\ref{tab:time_all_taxon_genus}, reveal several key insights.

Runtime is broadly correlated with model scale; larger backbones inevitably incur higher costs for embedding computation and forward inference. Within the Evo2 family, for instance, runtime scales from approximately 328 minutes for the 7B model to nearly 1,490 minutes for the 40B variant. A similar monotonic increase is observed in the NT V2 series, where runtime grows from 322 to 586 minutes as the parameter count increases from 50M to 500M.

Beyond parameter count, architectural paradigms significantly influence practical throughput. Notably, NT V3 (U-Net + Diffusion) achieves substantially shorter runtimes than its Transformer- or Hyena-based counterparts. Even at 650M parameters, NT V3 completes the evaluation pipeline in approximately 75 minutes, considerably faster than smaller models in other families. This efficiency likely stems from its distinct computational structure and feature-extraction pathway, which may offer superior parallelization and reduced sensitivity to sequence length during embedding extraction.

These results underscore that under a frozen-backbone protocol, inference efficiency is shaped by the interplay of parameter scale, architecture, and embedding strategy. Consequently, wall-clock time remains a critical metric alongside accuracy for evaluating the real-world deployability of genomic foundation models.

\begin{table*}[htbp]
  \centering
  \caption{Computational efficiency and throughput across foundation model backbones. Reported values represent the average end-to-end wall-clock time (minutes) on a single NVIDIA H200 GPU, covering embedding extraction and downstream evaluation for all-virus classification tasks. Timings are averaged across four representative settings: Kingdom-to-Genus and Host classification under both G-split and T-split strategies. Paradigm abbreviations include: \texttt{Trans-Enc} (Transformer encoder), \texttt{Trans-Dec} (Transformer decoder), \texttt{Trans-MoE} (MoE Transformer), \texttt{Hyena/SSM} (Hyena-style SSM), and \texttt{Mamba/SSM} (Mamba-style SSM).}
  \label{tab:time_all_taxon_genus}
  \small
  \setlength{\tabcolsep}{3pt}
  \renewcommand{\arraystretch}{1}

  \begin{tabular*}{\textwidth}{@{\extracolsep{\fill}}
    p{0.32\textwidth} >{\centering\arraybackslash}p{1.8cm} r
    @{\hspace{0.55em}\vrule\hspace{0.55em}}
    p{0.32\textwidth} >{\centering\arraybackslash}p{1.8cm} r
  @{}}
    \toprule
    \textbf{Name} & \textbf{Paradigm} & \textbf{Time} &
    \textbf{Name} & \textbf{Paradigm} & \textbf{Time} \\
    \midrule

    AIDO.DNA-300M & Trans-Enc & 155.33 & AIDO.DNA-7B & Trans-Enc & 659.80 \\
    AIDO.RNA-1.6B & Trans-Enc & 266.35 & AIDO.RNA-1.6B-CDS & Trans-Enc & 261.48 \\
    AIDO.RNA-650M & Trans-Enc & 175.43 & AIDO.RNA-650M-CDS & Trans-Enc & 180.73 \\
    BiRNA-BERT & Trans-Enc & 22.58 & Caduceus-ph & Mamba/SSM & 122.80 \\
    Caduceus-ps & Mamba/SSM & 120.68 & DNABERT (3mer) & Trans-Enc & 11.00 \\
    DNABERT (4mer) & Trans-Enc & 11.22 & DNABERT (5mer) & Trans-Enc & 11.31 \\
    DNABERT (6mer) & Trans-Enc & 12.63 & DNABERT-2 & Trans-Enc & 10.48 \\
    DNABERT-S & Trans-Enc & 12.13 & evo-1.5-8k-Base & Hyena/SSM & 359.88 \\
    Evo1 7B (131k) & Hyena/SSM & 361.03 & Evo1 7B (8k) & Hyena/SSM & 360.96 \\
    Evo2 1B Base & Hyena/SSM & 170.89 & Evo2 40B & Hyena/SSM & 1462.18 \\
    Evo2 40B Base & Hyena/SSM & 1441.43 & Evo2 7B & Hyena/SSM & 355.82 \\
    Evo2 7B Base & Hyena/SSM & 338.54 & gena-lm-bert-Base-t2t & Trans-Enc & 4.99 \\
    gena-lm-bert-large-t2t & Trans-Enc & 8.46 & gena-lm-bigbird-Base-t2t & Trans-Enc & 7.08 \\
    GENERator-v2-eukaryote-1.2b-Base & Trans-Dec & 20.54 & GENERator-v2-eukaryote-3b-Base & Trans-Dec & 30.33 \\
    GENERator-v2-prokaryote-1.2b-Base & Trans-Dec & 20.99 & GENERator-v2-prokaryote-3b-Base & Trans-Dec & 33.19 \\
    GenomeOcean-100M & Trans-Dec & 9.83 & GenomeOcean-4B & Trans-Dec & 36.64 \\
    GenomeOcean-500M & Trans-Dec & 12.40 & Genos-1.2B & Trans-MoE & 57.05 \\
    Genos-10B & Trans-MoE & 149.70 & Genos-10B-v2 & Trans-MoE & 157.70 \\
    Grover & Trans-Enc & 5.97 & HyenaDNA-Large-1M & Hyena/SSM & 9.38 \\
    HyenaDNA-Medium-160k & Hyena/SSM & 9.48 & HyenaDNA-Medium-450k & Hyena/SSM & 9.46 \\
    HyenaDNA-Small-32k & Hyena/SSM & 8.33 & HyenaDNA-Tiny-16k-d128 & Hyena/SSM & 7.19 \\
    HyenaDNA-Tiny-1k & Hyena/SSM & 7.09 & MP-RNA & Trans & 327.56 \\
    NT-2.5B-1000G & Trans-Enc & 138.35 & NT-2.5B-MS & Trans-Enc & 158.76 \\
    NT-500M-1000G & Trans-Enc & 42.99 & NT-500M-Human & Trans-Enc & 43.04 \\
    NTv2-100M-MS & Trans-Enc & 142.61 & NTv2-250M-MS & Trans-Enc & 185.75 \\
    NTv2-500M-MS & Trans-Enc & 238.59 & NTv2-50M-MS & Trans-Enc & 78.74 \\
    NTv2-50M-MS-3kmer & Trans-Enc & 80.92 & NTv3\_100M\_post & Diffusion & 20.37 \\
    NTv3\_100M\_pre & Diffusion & 11.07 & NTv3\_650M\_post & Diffusion & 22.14 \\
    NTv3\_650M\_pre & Diffusion & 16.74 & NTv3\_8M\_pre & Diffusion & 9.09 \\
    OmniReg-GPT & Trans-Dec & 27.13 & RiNALMo & Trans-Enc & 153.83 \\
    RNA-FM & Trans-Enc & 196.29 & RNABERT & Trans-Enc & 4.38 \\
    ViroHyena-1M & Hyena/SSM & 2.32 & ViroHyena-253M & Hyena/SSM & 26.23 \\
    ViroHyena-436k & Hyena/SSM & 1.95 & ViroHyena-6M & Hyena/SSM & 3.90 \\
    \bottomrule
  \end{tabular*}
\end{table*}

\section{Future Work and Limitations}\label{app:Data and Evaluation Limitations}
While \vb~ aims to provide a rigorous and biologically grounded benchmark, several simplifying assumptions merit discussion.

Our host classification pipeline assigns each virus to exactly one coarse-grained host category. In practice, many viruses exhibit broad or multi-host tropism. For example, influenza A circulates among avian, swine, and human. Framing host prediction as a single-label classification task does not capture this multiplicity and may penalize models that produce biologically reasonable but "incorrect" secondary host associations. Extending \vb~ to a multi-label host prediction setting is a natural direction for future work.

Our temporal partitioning relies on the NCBI record date for each virus, which reflects when a virus was first sequenced and deposited rather than when it actually emerged in nature. Many ancient viruses were only sequenced in recent decades, while heavily surveilled pathogens such as influenza are densely sampled in recent years. This conflation of sequencing effort with genuine evolutionary novelty means that the T-split may partly test a model's robustness to surveillance bias rather than purely to mutational drift. Incorporating molecular clock estimates or independent phylogenetic dating could help decouple these factors in future iterations.

The genus-disjoint split assumes that sequences from distinct genera share no significant homology, thereby enforcing phylogenetic extrapolation. However, recombination events have been documented across RNA viruses lineages (e.g., coronaviruses) and among bacteriophages, which may introduce shared genomic regions across genus boundaries. Reassortment in segmented viruses such as influenza can likewise produce chimeric genomes combining segments from different lineages. These events introduce shared genomic regions across genus boundaries, potentially allowing models to exploit partial homology and inflating apparent cross-genus generalization performance. Future benchmarks could incorporate recombination-aware filtering or breakpoint masking to enforce stricter phylogenetic isolation.

ViroHyena is currently trained with a maximum context length of $\sim$8k tokens, providing a practical trade-off between coverage and efficiency for many viral sequences in \vb. Extending pre-training to longer contexts is an important next step to better model genome-scale dependencies, especially for long genomes and tasks that require cross-locus reasoning.



\begin{table*}[htbp]
  \caption{Precision for viral taxonomy and host classification are reported for the full suite of models.}
  \label{tab:precision_results}
  \centering
  \scriptsize
  \setlength{\tabcolsep}{3pt}
  \renewcommand{\arraystretch}{1.08}

  \resizebox{\textwidth}{!}{%
  \begin{tabular}{l*{12}{c}}
    \toprule
    \multirow{3}{*}{\textbf{Model Name}} &
    \multicolumn{4}{c}{\textbf{ALL Viruses}} &
    \multicolumn{4}{c}{\textbf{DNA Viruses}} &
    \multicolumn{4}{c}{\textbf{RNA Viruses}} \\
    \cmidrule(lr){2-5}\cmidrule(lr){6-9}\cmidrule(lr){10-13}
    &
    \multicolumn{2}{c}{\textbf{Taxonomy}} & \multicolumn{2}{c}{\textbf{Host}} &
    \multicolumn{2}{c}{\textbf{Taxonomy}} & \multicolumn{2}{c}{\textbf{Host}} &
    \multicolumn{2}{c}{\textbf{Taxonomy}} & \multicolumn{2}{c}{\textbf{Host}} \\
    \cmidrule(lr){2-3}\cmidrule(lr){4-5}\cmidrule(lr){6-7}\cmidrule(lr){8-9}\cmidrule(lr){10-11}\cmidrule(lr){12-13}
    &
    G-Split & T-Split & G-Split & T-Split &
    G-Split & T-Split & G-Split & T-Split &
    G-Split & T-Split & G-Split & T-Split \\

    \grouphead{Baseline}{13}
    BLAST & 47.69\std{0.00} & 41.25\std{0.00} & 93.19\std{0.00} & 64.19\std{0.00} & 75.86\std{0.00} & 39.89\std{0.00} & 75.88\std{0.00} & 29.36\std{0.00} & 59.44\std{0.00} & 75.90\std{0.00} & 93.76\std{0.00} & 86.87\std{0.00} \\

    Kraken2 & 26.82\std{0.00} & 34.92\std{0.00} & 70.15\std{0.00} & 68.57\std{0.00} & 52.62\std{0.00} & 34.12\std{0.00} & 65.87\std{0.00} & 38.54\std{0.00} & 39.41\std{0.00} & 71.32\std{0.00} & 56.38\std{0.00} & 70.40\std{0.00} \\
    BiLSTM & 68.77\std{1.89} & 60.67\std{1.32} & 84.46\std{0.55} & 47.78\std{0.97} & 73.88\std{2.72} & 72.54\std{5.13} & 66.57\std{5.82} & 63.25\std{1.92} & 79.38\std{3.80} & 60.73\std{1.53} & 83.69\std{0.44} & 67.47\std{2.96} \\
    CNN & 37.91\std{11.78} & 26.61\std{17.66} & 73.75\std{3.01} & 31.46\std{6.04} & 28.37\std{22.36} & 25.34\std{8.80} & 42.36\std{7.23} & 35.09\std{5.52} & 34.85\std{24.69} & 39.57\std{6.32} & 66.12\std{10.70} & 47.82\std{11.29} \\
\bottomrule
  \end{tabular}%
  }
\end{table*}

\begin{table*}[htbp]
\addtocounter{table}{-1}
  \caption{Precision results on viral taxonomy and host classification (continued).}
  \label{tab:precision_results_continued}
  \centering
  \scriptsize
  \setlength{\tabcolsep}{3pt}
  \renewcommand{\arraystretch}{1.08}

  \resizebox{\textwidth}{!}{%
  \begin{tabular}{l*{12}{c}}
    \toprule
    \multirow{3}{*}{\textbf{Model Name}} &
    \multicolumn{4}{c}{\textbf{ALL Viruses}} &
    \multicolumn{4}{c}{\textbf{DNA Viruses}} &
    \multicolumn{4}{c}{\textbf{RNA Viruses}} \\
    \cmidrule(lr){2-5}\cmidrule(lr){6-9}\cmidrule(lr){10-13}
    &
    \multicolumn{2}{c}{\textbf{Taxonomy}} & \multicolumn{2}{c}{\textbf{Host}} &
    \multicolumn{2}{c}{\textbf{Taxonomy}} & \multicolumn{2}{c}{\textbf{Host}} &
    \multicolumn{2}{c}{\textbf{Taxonomy}} & \multicolumn{2}{c}{\textbf{Host}} \\
    \cmidrule(lr){2-3}\cmidrule(lr){4-5}\cmidrule(lr){6-7}\cmidrule(lr){8-9}\cmidrule(lr){10-11}\cmidrule(lr){12-13}
    &
    G-Split & T-Split & G-Split & T-Split &
    G-Split & T-Split & G-Split & T-Split &
    G-Split & T-Split & G-Split & T-Split \\

    \grouphead{RNA Foundation Models (RNA-specific Coverage)}{13}
    AIDO.RNA-650M & 54.93\std{7.62} & 38.89\std{6.39} & 74.43\std{1.92} & 39.06\std{0.87} & 58.07\std{5.50} & 49.37\std{5.49} & 55.08\std{0.44} & 41.27\std{2.67} & 66.87\std{4.84} & 46.15\std{3.82} & 54.52\std{6.18} & 14.14\std{15.90} \\
    AIDO.RNA-1.6B & 52.71\std{6.64} & 35.84\std{5.90} & 70.68\std{0.93} & 35.92\std{3.84} & 56.50\std{2.55} & 44.66\std{4.67} & 52.73\std{6.34} & 42.90\std{3.42} & 64.02\std{4.79} & 44.07\std{4.09} & 46.34\std{2.59} & 11.74\std{16.32} \\
    AIDO.RNA-650M-CDS & 64.20\std{4.61} & 47.22\std{6.97} & 78.59\std{0.26} & 46.89\std{3.75} & 71.12\std{3.80} & 64.76\std{5.05} & 57.01\std{1.72} & 42.29\std{2.60} & 75.44\std{4.47} & 56.14\std{4.15} & 71.06\std{1.82} & 41.36\std{7.30} \\
    AIDO.RNA-1.6B-CDS & 60.22\std{6.72} & 45.20\std{4.98} & 77.18\std{1.76} & 41.20\std{0.71} & 67.36\std{3.25} & 58.06\std{4.22} & 56.41\std{2.51} & 43.49\std{1.33} & 74.86\std{3.81} & 51.99\std{2.93} & 71.37\std{1.45} & 30.79\std{2.70} \\
\bottomrule
  \end{tabular}%
  }
\end{table*}


\begin{table*}[htbp]
  \caption{Recall for viral taxonomy and host classification are reported for the full suite of models.}
  \label{tab:recall_results}
  \centering
  \scriptsize
  \setlength{\tabcolsep}{3pt}
  \renewcommand{\arraystretch}{1}

  \resizebox{\textwidth}{!}{%
  \begin{tabular}{l*{12}{c}}
    \toprule
    \multirow{3}{*}{\textbf{Model Name}} &
    \multicolumn{4}{c}{\textbf{ALL Viruses}} &
    \multicolumn{4}{c}{\textbf{DNA Viruses}} &
    \multicolumn{4}{c}{\textbf{RNA Viruses}} \\
    \cmidrule(lr){2-5}\cmidrule(lr){6-9}\cmidrule(lr){10-13}
    &
    \multicolumn{2}{c}{\textbf{Taxonomy}} & \multicolumn{2}{c}{\textbf{Host}} &
    \multicolumn{2}{c}{\textbf{Taxonomy}} & \multicolumn{2}{c}{\textbf{Host}} &
    \multicolumn{2}{c}{\textbf{Taxonomy}} & \multicolumn{2}{c}{\textbf{Host}} \\
    \cmidrule(lr){2-3}\cmidrule(lr){4-5}\cmidrule(lr){6-7}\cmidrule(lr){8-9}\cmidrule(lr){10-11}\cmidrule(lr){12-13}
    &
    G-Split & T-Split & G-Split & T-Split &
    G-Split & T-Split & G-Split & T-Split &
    G-Split & T-Split & G-Split & T-Split \\

    \grouphead{Baseline}{13}
    BLAST & 47.75\std{0.00} & 41.35\std{0.00} & 91.90\std{0.00} & 69.43\std{0.00} & 75.62\std{0.00} & 39.93\std{0.00} & 75.11\std{0.00} & 28.74\std{0.00} & 60.03\std{0.00} & 75.77\std{0.00} & 92.37\std{0.00} & 80.79\std{0.00} \\
    Kraken2 & 26.78\std{0.00} & 35.03\std{0.00} & 63.55\std{0.00} & 73.23\std{0.00} & 52.62\std{0.00} & 34.12\std{0.00} & 69.16\std{0.00} & 41.67\std{0.00} & 39.33\std{0.00} & 71.88\std{0.00} & 44.24\std{0.00} & 68.95\std{0.00} \\
    BiLSTM & 66.17\std{1.50} & 56.29\std{2.40} & 85.25\std{1.27} & 43.18\std{1.12} & 69.18\std{2.94} & 56.97\std{2.35} & 62.02\std{8.04} & 53.36\std{0.47} & 72.52\std{3.49} & 59.94\std{2.41} & 82.36\std{1.42} & 68.67\std{4.00} \\
    CNN & 37.98\std{9.58} & 21.85\std{14.27} & 70.18\std{2.42} & 27.50\std{4.39} & 31.72\std{23.19} & 26.10\std{6.33} & 41.47\std{6.28} & 30.90\std{5.27} & 32.28\std{21.22} & 38.58\std{4.53} & 60.04\std{6.38} & 44.48\std{10.55} \\
\bottomrule
  \end{tabular}%
  }
\end{table*}

\begin{table*}[htbp]
  \addtocounter{table}{-1}
  \caption{Recall for viral taxonomy and host classification are reported for the full suite of models (continued).}
  \label{tab:virus_cls_results_cont-rc}
  \centering
  \scriptsize
  \setlength{\tabcolsep}{3pt}
  \renewcommand{\arraystretch}{1.08}

  \resizebox{\textwidth}{!}{%
  \begin{tabular}{l*{12}{c}}
    \toprule
    \multirow{3}{*}{\textbf{Model Name}} &
    \multicolumn{4}{c}{\textbf{ALL Viruses}} &
    \multicolumn{4}{c}{\textbf{DNA Viruses}} &
    \multicolumn{4}{c}{\textbf{RNA Viruses}} \\
    \cmidrule(lr){2-5}\cmidrule(lr){6-9}\cmidrule(lr){10-13}
    &
    \multicolumn{2}{c}{\textbf{Taxonomy}} & \multicolumn{2}{c}{\textbf{Host}} &
    \multicolumn{2}{c}{\textbf{Taxonomy}} & \multicolumn{2}{c}{\textbf{Host}} &
    \multicolumn{2}{c}{\textbf{Taxonomy}} & \multicolumn{2}{c}{\textbf{Host}} \\
    \cmidrule(lr){2-3}\cmidrule(lr){4-5}\cmidrule(lr){6-7}\cmidrule(lr){8-9}\cmidrule(lr){10-11}\cmidrule(lr){12-13}
    &
    G-Split & T-Split & G-Split & T-Split &
    G-Split & T-Split & G-Split & T-Split &
    G-Split & T-Split & G-Split & T-Split \\

    \grouphead{RNA Foundation Models (RNA-specific Coverage)}{13}
    AIDO.RNA-650M & 61.88\std{5.80} & 41.54\std{6.70} & 76.10\std{1.39} & 34.18\std{0.98} & 72.41\std{4.93} & 41.87\std{2.76} & 49.58\std{0.69} & 42.76\std{1.06} & 73.66\std{2.88} & 53.57\std{3.84} & 51.18\std{0.81} & 23.87\std{11.49} \\
    AIDO.RNA-1.6B & 59.65\std{6.20} & 38.48\std{5.98} & 70.61\std{1.99} & 31.17\std{2.73} & 72.48\std{3.85} & 37.89\std{4.27} & 47.54\std{1.11} & 44.67\std{1.89} & 70.70\std{5.27} & 52.95\std{5.51} & 47.99\std{1.42} & 19.89\std{12.80} \\
    AIDO.RNA-650M-CDS & 70.87\std{3.34} & 51.57\std{7.22} & 80.46\std{0.54} & 42.34\std{1.53} & 79.36\std{2.71} & 57.94\std{3.82} & 53.81\std{1.53} & 42.66\std{4.09} & 77.78\std{4.44} & 61.51\std{5.13} & 71.09\std{3.13} & 40.26\std{1.88} \\
    AIDO.RNA-1.6B-CDS & 66.44\std{5.24} & 48.08\std{6.23} & 79.04\std{1.18} & 37.93\std{1.87} & 79.22\std{3.17} & 52.85\std{4.68} & 52.70\std{2.72} & 47.18\std{0.53} & 78.53\std{3.50} & 61.35\std{4.19} & 68.65\std{5.12} & 36.10\std{2.08} \\
\bottomrule
  \end{tabular}%
  }
\end{table*}

\begin{table*}[htbp]
  \caption{Macro-F1 scores for viral taxonomy and host classification. Evaluation covers ALL, DNA, and RNA viruses sets under genus-disjoint (G-split) and temporal (T-split) protocols. Means (standard deviations) are reported.}
  \label{tab:virus_cls_results_cont}
  \centering
  \scriptsize
  \setlength{\tabcolsep}{3pt}
  \renewcommand{\arraystretch}{1}

  \resizebox{\textwidth}{!}{%
  \begin{tabular}{l*{12}{c}}
    \toprule
    \multirow{3}{*}{\textbf{Model Name}} &
    \multicolumn{4}{c}{\textbf{ALL Viruses}} &
    \multicolumn{4}{c}{\textbf{DNA Viruses}} &
    \multicolumn{4}{c}{\textbf{RNA Viruses}} \\
    \cmidrule(lr){2-5}\cmidrule(lr){6-9}\cmidrule(lr){10-13}
    &
    \multicolumn{2}{c}{\textbf{Taxonomy}} & \multicolumn{2}{c}{\textbf{Host}} &
    \multicolumn{2}{c}{\textbf{Taxonomy}} & \multicolumn{2}{c}{\textbf{Host}} &
    \multicolumn{2}{c}{\textbf{Taxonomy}} & \multicolumn{2}{c}{\textbf{Host}} \\
    \cmidrule(lr){2-3}\cmidrule(lr){4-5}\cmidrule(lr){6-7}\cmidrule(lr){8-9}\cmidrule(lr){10-11}\cmidrule(lr){12-13}
    &
    G-Split & T-Split & G-Split & T-Split &
    G-Split & T-Split & G-Split & T-Split &
    G-Split & T-Split & G-Split & T-Split \\

    \grouphead{Baseline}{13}
    BLAST & 47.67\std{0.00} & 41.22\std{0.00} & 92.50\std{0.00} & 65.55\std{0.00} & 75.68\std{0.00} & 39.91\std{0.00} & 75.42\std{0.00} & 25.81\std{0.00} & 59.65\std{0.00} & 75.74\std{0.00} & 93.01\std{0.00} & 79.09\std{0.00} \\
    Kraken2 & 26.78\std{0.00} & 34.93\std{0.00} & 61.70\std{0.00} & 69.41\std{0.00} & 52.62\std{0.00} & 34.12\std{0.00} & 67.05\std{0.00} & 35.71\std{0.00} & 39.36\std{0.00} & 71.46\std{0.00} & 40.49\std{0.00} & 65.52\std{0.00} \\
    BiLSTM & 66.05\std{1.89} & 54.67\std{2.27} & 84.40\std{0.98} & 44.69\std{1.31} & 69.67\std{3.25} & 57.79\std{2.76} & 62.90\std{7.82} & 56.48\std{0.80} & 73.96\std{3.79} & 57.43\std{1.77} & 81.56\std{0.57} & 65.11\std{2.04} \\
    CNN & 34.72\std{10.96} & 19.26\std{13.92} & 69.29\std{2.51} & 25.16\std{5.52} & 26.63\std{20.35} & 21.45\std{5.73} & 39.87\std{6.87} & 32.62\std{5.47} & 32.07\std{21.49} & 34.81\std{4.19} & 60.46\std{7.49} & 40.71\std{13.21} \\
\bottomrule
  \end{tabular}%
  }
\end{table*}

\begin{table*}[htbp]
  \addtocounter{table}{-1}
  \caption{Macro-F1 scores for viral taxonomy and host classification (continued).}
  \label{tab:f1_cls_results_cont_dna}
  \centering
  \scriptsize
  \setlength{\tabcolsep}{3pt}
  \renewcommand{\arraystretch}{1}

  \resizebox{\textwidth}{!}{%
  \begin{tabular}{l*{12}{c}}
    \toprule
    \multirow{3}{*}{\textbf{Model Name}} &
    \multicolumn{4}{c}{\textbf{ALL Viruses}} &
    \multicolumn{4}{c}{\textbf{DNA Viruses}} &
    \multicolumn{4}{c}{\textbf{RNA Viruses}} \\
    \cmidrule(lr){2-5}\cmidrule(lr){6-9}\cmidrule(lr){10-13}
    &
    \multicolumn{2}{c}{\textbf{Taxonomy}} & \multicolumn{2}{c}{\textbf{Host}} &
    \multicolumn{2}{c}{\textbf{Taxonomy}} & \multicolumn{2}{c}{\textbf{Host}} &
    \multicolumn{2}{c}{\textbf{Taxonomy}} & \multicolumn{2}{c}{\textbf{Host}} \\
    \cmidrule(lr){2-3}\cmidrule(lr){4-5}\cmidrule(lr){6-7}\cmidrule(lr){8-9}\cmidrule(lr){10-11}\cmidrule(lr){12-13}
    &
    G-Split & T-Split & G-Split & T-Split &
    G-Split & T-Split & G-Split & T-Split &
    G-Split & T-Split & G-Split & T-Split \\

    \grouphead{DNA Foundation Models (Non-viral Coverage)}{13}
    AIDO.DNA-300M & 71.27\std{2.63} & 57.11\std{6.19} & 82.82\std{1.15} & 46.96\std{2.85} & 79.38\std{3.84} & 63.40\std{2.00} & 57.58\std{0.84} & 44.34\std{1.25} & 84.97\std{2.64} & 66.86\std{3.64} & 76.33\std{5.33} & 47.45\std{1.67} \\
    AIDO.DNA-7B & 69.87\std{3.05} & 55.27\std{6.59} & 80.65\std{1.23} & 47.47\std{1.38} & 79.37\std{2.04} & 63.52\std{4.08} & 56.61\std{2.01} & 45.13\std{1.25} & 81.28\std{2.17} & 64.64\std{4.66} & 62.90\std{0.69} & 40.72\std{6.88} \\
    Caduceus-ph & 30.87\std{6.17} & 19.39\std{3.97} & 55.21\std{1.34} & 21.68\std{4.35} & 30.22\std{6.96} & 23.07\std{2.05} & 44.13\std{1.31} & 31.59\std{0.73} & 47.90\std{2.27} & 20.24\std{6.41} & 42.38\std{1.92} & 26.67\std{1.58} \\
    Caduceus-ps & 33.56\std{6.61} & 18.04\std{3.17} & 54.57\std{5.57} & 19.54\std{2.97} & 36.78\std{3.11} & 16.90\std{8.42} & 44.42\std{2.67} & 16.11\std{0.00} & 46.68\std{5.25} & 31.39\std{2.20} & 37.34\std{3.58} & 12.45\std{14.80} \\
\bottomrule
  \end{tabular}%
  }
\end{table*}






\begin{table*}[htbp]
  \caption{Macro-F1 scores on the ALL-Taxon task across taxonomic ranks under genus-disjoint (G-split) and temporal (T-split) evaluation. Results are reported as mean (standard deviation).}
  \label{tab:taxon_rank_results}
  \centering
  \scriptsize
  \setlength{\tabcolsep}{3pt}
  \renewcommand{\arraystretch}{1}

  \resizebox{\textwidth}{!}{%
  \begin{tabular}{l*{10}{c}}
    \toprule
    \multirow{2}{*}{\textbf{Model}} &
    \multicolumn{5}{c}{\textbf{G-Split}} &
    \multicolumn{5}{c}{\textbf{T-Split}} \\
    \cmidrule(lr){2-6}\cmidrule(lr){7-11}
    &
    Kingdom & Phylum & Class & Order & Family &
    Kingdom & Phylum & Class & Order & Family \\

    \grouphead{Baseline}{11}
    BLAST & 67.75\std{0.00} & 64.60\std{0.00} & 30.66\std{0.00} & 21.92\std{0.00} & 53.41\std{0.00} & 63.53\std{0.00} & 52.55\std{0.00} & 19.81\std{0.00} & 16.91\std{0.00} & 53.28\std{0.00} \\
    
    Kraken2 & 39.29\std{0.00} & 35.29\std{0.00} & 15.58\std{0.00} & 12.22\std{0.00} & 31.51\std{0.00} & 51.88\std{0.00} & 44.53\std{0.00} & 17.47\std{0.00} & 14.98\std{0.00} & 45.79\std{0.00} \\
    BiLSTM & 68.34\std{3.99} & 68.92\std{0.54} & 69.02\std{2.62} & 62.25\std{0.36} & 61.70\std{1.95} & 57.58\std{1.32} & 58.55\std{4.04} & 51.94\std{2.73} & 53.15\std{2.20} & 52.13\std{1.05} \\
    CNN & 39.90\std{4.57} & 36.21\std{9.29} & 37.00\std{16.31} & 31.40\std{13.23} & 29.07\std{11.39} & 22.45\std{10.43} & 22.84\std{14.42} & 18.54\std{16.32} & 17.51\std{14.37} & 14.94\std{14.07} \\

    \grouphead{DNA Foundation Models (Diverse Viral Coverage)}{11}
    LucaOne-default-step36M & 74.77\std{1.18} & 76.29\std{4.33} & 70.80\std{6.22} & 60.48\std{5.00} & 66.60\std{1.11} & 61.09\std{5.28} & 67.55\std{1.92} & 52.54\std{2.96} & 47.93\std{5.04} & 58.14\std{1.86} \\
    LucaOne-gene-step36.8M & 67.73\std{12.70} & 70.97\std{5.37} & 64.67\std{12.15} & 49.89\std{6.98} & 43.98\std{37.96} & 50.86\std{8.71} & 55.95\std{8.99} & 40.76\std{8.54} & 30.19\std{27.32} & 45.71\std{14.01} \\
\bottomrule
  \end{tabular}%
  }
\end{table*}

\begin{table*}[htbp]
  \addtocounter{table}{-1}
  \caption{Macro-F1 scores across taxonomic ranks under G-split and T-split (continued).}
  \label{tab:taxon_rank_results_cont}
  \centering
  \scriptsize
  \setlength{\tabcolsep}{3pt}
  \renewcommand{\arraystretch}{1}

  \resizebox{\textwidth}{!}{%
  \begin{tabular}{l*{10}{c}}
    \toprule
    \multirow{2}{*}{\textbf{Model}} &
    \multicolumn{5}{c}{\textbf{G-Split}} &
    \multicolumn{5}{c}{\textbf{T-Split}} \\
    \cmidrule(lr){2-6}\cmidrule(lr){7-11}
    &
    Kingdom & Phylum & Class & Order & Family &
    Kingdom & Phylum & Class & Order & Family \\
    \grouphead{DNA Foundation Models (Non-viral Coverage)}{11}
    AIDO.DNA-300M & 73.01\std{2.67} & 77.56\std{3.14} & 72.22\std{4.48} & 63.54\std{0.95} & 70.01\std{1.92} & 62.62\std{7.79} & 62.74\std{4.39} & 51.94\std{6.20} & 50.71\std{7.33} & 57.52\std{5.23} \\
    AIDO.DNA-7B & 72.82\std{2.21} & 77.25\std{3.27} & 70.91\std{1.17} & 59.92\std{6.20} & 68.44\std{2.42} & 63.27\std{6.34} & 63.01\std{5.08} & 48.89\std{7.80} & 48.22\std{7.87} & 52.97\std{5.87} \\
    Caduceus-ph & 37.22\std{5.78} & 35.39\std{3.83} & 34.43\std{7.66} & 22.63\std{5.81} & 24.66\std{7.76} & 23.79\std{4.05} & 24.54\std{0.82} & 16.43\std{3.40} & 14.23\std{5.49} & 17.95\std{6.10} \\
    Caduceus-ps & 42.94\std{4.66} & 36.59\std{7.09} & 33.80\std{6.77} & 24.52\std{7.26} & 29.95\std{7.26} & 23.98\std{3.92} & 23.45\std{1.63} & 14.10\std{2.62} & 11.82\std{2.83} & 16.85\std{4.83} \\
    Genos-1.2B & 7.03\std{2.59} & 4.49\std{2.02} & 1.72\std{0.83} & 1.16\std{0.56} & 0.64\std{0.47} & 2.36\std{0.00} & 0.58\std{0.00} & 0.12\std{0.00} & 0.01\std{0.00} & 0.00\std{0.00} \\
    Genos-10B & 28.74\std{5.77} & 22.12\std{21.94} & 16.53\std{17.91} & 9.78\std{13.45} & 13.12\std{19.28} & 16.31\std{12.12} & 14.25\std{12.92} & 7.50\std{7.83} & 6.13\std{6.57} & 9.17\std{10.71} \\
\bottomrule
  \end{tabular}%
  }
\end{table*}


\begin{table*}[t]
  \caption{Summary statistics of BPB across genome-length buckets. For each model and bucket, we report the minimum, maximum, median, and mean BPB. Models are sorted in lexicographic order by name.}
  \label{tab:bpb_genome_bucket_stats}
  \centering
  \small
  \setlength{\tabcolsep}{2pt}
  \renewcommand{\arraystretch}{1.05}

  \begin{tabular*}{\textwidth}{@{\extracolsep{\fill}}
    >{\raggedright\arraybackslash}p{0.23\textwidth}
    *{12}{c}
  @{}}
    \toprule
    \textbf{Model} &
    \multicolumn{4}{c}{\textbf{Genome-Short}} &
    \multicolumn{4}{c}{\textbf{Genome-Medium}} &
    \multicolumn{4}{c}{\textbf{Genome-Long}} \\
    \cmidrule(lr){2-5}\cmidrule(lr){6-9}\cmidrule(lr){10-13}
    & \textbf{Min} & \textbf{Max} & \textbf{Median} & \textbf{Mean}
    & \textbf{Min} & \textbf{Max} & \textbf{Median} & \textbf{Mean}
    & \textbf{Min} & \textbf{Max} & \textbf{Median} & \textbf{Mean} \\
    \midrule

    Evo-1-131K-Base & 1.5276 & 3.0700 & 2.1651 & 2.1739 & 0.8368 & 3.2950 & 2.2040 & 2.1890 & 1.0725 & 3.1816 & 2.1487 & 2.1341 \\
    Evo-1-8K-Base   & 1.3742 & 2.7462 & 2.0567 & 2.0547 & 0.7556 & 3.0240 & 2.0637 & 2.0590 & 0.9759 & 3.0830 & 2.0559 & 2.0472 \\
    Evo-1.5-8K-Base & 1.2558 & 2.2288 & 1.9230 & 1.9230 & 0.7192 & 2.2036 & 1.9047 & 1.9035 & 0.5248 & 2.2861 & 1.8924 & 1.8772 \\
    Evo2 1B Base  & 1.2697 & 2.2213 & 1.9120 & 1.9113 & 0.7196 & 2.2325 & 1.8999 & 1.8998 & 0.4066 & 2.3316 & 1.8921 & 1.8840 \\
    Evo2 40B       & 0.8957 & 2.1941 & 1.9038 & 1.9010 & 0.7134 & 2.1422 & 1.8609 & 1.8651 & 0.1163 & 2.2749 & 1.8817 & 1.8660 \\
    Evo2 4B Base & 0.8632 & 2.1865 & 1.9067 & 1.9038 & 0.7274 & 2.1784 & 1.8795 & 1.8813 & 0.3170 & 2.3022 & 1.8816 & 1.8699 \\
\bottomrule
  \end{tabular*}
\end{table*}


\begin{table*}[t]
  \caption{Results on CDS continuation across length buckets. Exact Match Accuracy and CDS Success Rate are reported in \%, with the symbol moved to the column headers. Edit Distance / K-mer JSD / K-mer KS are unitless. For Edit Distance / K-mer JSD / K-mer KS, lower is better; for Exact Match Accuracy / CDS Success Rate, higher is better.}
  \label{tab:cds_gen_results_detail}
  \centering
  \small
  \setlength{\tabcolsep}{2pt}
  \renewcommand{\arraystretch}{1}

  \begin{tabular*}{\textwidth}{@{\extracolsep{\fill}}l*{15}{c}@{}}
    \toprule
    \textbf{Model Name} &
    \multicolumn{5}{c}{\textbf{CDS-Short}} &
    \multicolumn{5}{c}{\textbf{CDS-Medium}} &
    \multicolumn{5}{c}{\textbf{CDS-Long}} \\
    \cmidrule(lr){2-6}\cmidrule(lr){7-11}\cmidrule(lr){12-16}

    & \textbf{Edit $\downarrow$}
    & \textbf{Match $\uparrow$}
    & \textbf{JSD $\downarrow$}
    & \textbf{KS $\downarrow$}
    & \textbf{Succ. $\uparrow$}
    & \textbf{Edit $\downarrow$}
    & \textbf{Match $\uparrow$}
    & \textbf{JSD $\downarrow$}
    & \textbf{KS $\downarrow$}
    & \textbf{Succ. $\uparrow$}
    & \textbf{Edit $\downarrow$}
    & \textbf{Match $\uparrow$}
    & \textbf{JSD $\downarrow$}
    & \textbf{KS $\downarrow$}
    & \textbf{Succ. $\uparrow$} \\
    \midrule
    Evo-1-131k-Base
      & 0.5784 & 26.29 & 0.2155 & 0.2280 & 0.7273
      & 0.5593 & 25.88 & 0.1986 & 0.2266 & 0.3822
      & 0.5577 & 25.15 & 0.1675 & 0.2101 & 0.0436 \\
    Evo-1-8k-Base
      & 0.5950 & 25.62 & 0.2073 & 0.2251 & 0.6993
      & 0.5844 & 25.19 & 0.2000 & 0.2446 & 0.1820
      & 0.5717 & 24.87 & 0.1787 & 0.2469 & 0.0000 \\
    Evo-1.5-8k-Base
      & 0.5521 & 26.82 & 0.1563 & 0.1331 & 0.5315
      & 0.5326 & 26.42 & 0.1247 & 0.1139 & 0.2548
      & 0.5235 & 26.15 & 0.1049 & 0.1021 & 0.0109 \\
    Evo2 1B Base
      & 0.5508 & 26.92 & 0.1572 & 0.1358 & 0.7552
      & 0.5326 & 26.38 & 0.1285 & 0.1181 & 0.1274
      & 0.5248 & 26.04 & 0.1115 & 0.1103 & 0.0218 \\
    Evo2 40B
      & 0.5469 & 27.30 & 0.1525 & 0.1310 & 1.4270
      & 0.5293 & 26.69 & 0.1243 & 0.1151 & 0.6005
      & 0.5218 & 26.15 & 0.1076 & 0.1063 & 0.0545 \\
    Evo2 40B Base
      & 0.5478 & 27.23 & 0.1540 & 0.1313 & 1.4830
      & 0.5296 & 26.73 & 0.1256 & 0.1158 & 0.2912
      & 0.5225 & 26.12 & 0.1080 & 0.1064 & 0.1198 \\
\bottomrule
  \end{tabular*}
\end{table*}

\end{document}